\PassOptionsToPackage{dvipsnames,table}{xcolor}
\PassOptionsToPackage{shortlabels,inline}{enumitem}
\documentclass[]{fairmeta}

\usepackage[dvipsnames,table]{xcolor}
\usepackage{array}
\usepackage{amssymb}
\usepackage{caption}
\usepackage{subcaption}
\makeatletter
\@ifpackageloaded{enumitem}{}{\usepackage[shortlabels,inline]{enumitem}}
\makeatother
\setlist[itemize]{noitemsep,leftmargin=*,topsep=0em}
\setlist[enumerate]{noitemsep,leftmargin=*,topsep=0em}

\usepackage[utf8]{inputenc}
\usepackage[T1]{fontenc}

\usepackage{url}
\usepackage{booktabs}
\usepackage{amsfonts}
\usepackage{nicefrac}
\usepackage{microtype}
\usepackage{graphicx}
\usepackage{multirow}
\usepackage{amsmath}
\usepackage{algorithm}
\usepackage{algpseudocode}
\usepackage{tikz}
\usepackage{arydshln}
\usetikzlibrary{tikzmark,arrows.meta,calc,decorations.pathreplacing}
\usepackage{pifont}
\usepackage{xpatch}
\usepackage{xspace}
\usepackage{tcolorbox}
\tcbuselibrary{skins}

\makeatletter
\newcommand{\thickhline}{%
    \noalign {\ifnum 0=`}\fi \hrule height 1pt
    \futurelet \reserved@a \@xhline
}
\makeatother
\newcolumntype{I}{!{\vrule width 1pt}}

\newcommand{\ours}{RecWM}

\makeatletter
\DeclareRobustCommand\onedot{\futurelet\@let@token\@onedot}
\def\@onedot{\ifx\@let@token.\else.\null\fi\xspace}
\def\eg{\emph{e.g}\onedot}

\makeatother

\definecolor{recforward}{HTML}{65AC89}
\definecolor{recreverse}{HTML}{D86ABB}
\definecolor{recjoint}{HTML}{709CC2}
\definecolor{recjointbg}{HTML}{E9F2FA}

\newcommand{\rectrainmark}[3]{%
    \Statex
    \makebox[\linewidth][l]{%
        \textcolor{#2}{\textbf{#3}}\hfill
        \tikzmarknode{#1}{\strut}%
    }%
}

\newcommand{\recinfermark}[3]{%
    \Statex
    \tikzmarknode{#1}{\textcolor{#2}{\textbf{#3}}}%
}

\newcommand{\recflushalgorithmic}{%
    \xpatchcmd{\algorithmic}
        {\leftmargin\labelwidth \addtolength{\leftmargin}{\labelsep}}
        {\leftmargin=0pt}
        {}
        {\PackageWarning{recwm}{Failed to remove algorithm margin}}%
}

\newtcolorbox{recblock}[2][]{%
    colback=#2!5,
    colframe=#2!25,
    boxrule=0.3pt,
    arc=2pt,
    outer arc=2pt,
    boxsep=0pt,
    left=2pt,
    right=2pt,
    top=2pt,
    bottom=2pt,
    before skip=2pt,
    after skip=2pt,
    #1%
}

\makeatletter
\@ifpackageloaded{hyperref}{}{%
    \providecolor{cvprblue}{rgb}{0.21,0.49,0.74}%
    \usepackage[breaklinks,colorlinks,allcolors=cvprblue]{hyperref}%
}
\usepackage{orcidlink}
\AtEndPreamble{%
    \@ifpackageloaded{natbib}{}{%
        \usepackage[numbers,sort&compress]{natbib}%
    }%
    \@ifpackageloaded{cleveref}{}{%
        \usepackage[capitalize]{cleveref}%
        \crefname{section}{Sec.}{Secs.}%
        \Crefname{section}{Section}{Sections}%
        \Crefname{table}{Table}{Tables}%
        \crefname{table}{Tab.}{Tabs.}%
    }%
}
\makeatother

\definecolor{egoGaming}{HTML}{5C8FBD}
\definecolor{navigation}{HTML}{9274B5}

\newtcolorbox{promptbox}[2][teal]{
  enhanced,
  colback=#1!5,
  colframe=#1!50,
  title={#2},
  fonttitle=\bfseries,
  fontupper=\small,
  sharp corners,
  boxrule=0.8pt,
  width=\linewidth,
  left=3mm, right=3mm, top=2mm, bottom=2mm,
  before upper={\setlength{\parindent}{0pt}\setlength{\parskip}{3pt}}
}

\newcommand{\promptsection}[1]{%
  \begin{tcolorbox}[
    colback=gray!20, colframe=gray!20, boxrule=0pt, arc=1mm,
    left=1mm, right=1mm, top=0.6mm, bottom=0.6mm,
    before skip=2mm, after skip=2mm,
    halign=center, fontupper=\sffamily\bfseries]
    #1
  \end{tcolorbox}%
}

\newcommand{\prompttok}[1]{{\footnotesize\ttfamily\color{teal!60!black}#1}}
\newcommand{\promptfield}[2]{{\sffamily\bfseries #1}\enspace\prompttok{#2}}
\newcommand{\promptslot}[2]{\texttt{#1:}\enspace\textit{#2}}

\newcommand{\acttok}[1]{\texttt{\textless\textbar #1\textbar\textgreater}}

\newlength{\promptobsheight}
\newcommand{\promptobs}[3][1]{%
  \begin{tabular}[b]{@{}c@{}}
    {\footnotesize #2}\\[0.5mm]
    \if\relax\detokenize{#3}\relax
      \fboxsep=0pt\fboxrule=0.4pt
      \fcolorbox{gray!45}{white}{%
        \parbox[c][\promptobsheight][c]{#1\promptobsheight}{\centering\footnotesize\sffamily\color{gray} image}}%
    \else
      \includegraphics[height=\promptobsheight]{#3}%
    \fi
  \end{tabular}%
}

\DeclareRobustCommand{\afflogo}[2][1.3em]{%
  \ensuremath{\vcenter{\hbox{\includegraphics[height=#1]{#2}}}}}

\title{Reciprocal World Models: Mutual Refinement of Planning and Imagination}

\author[1,*]{Yifei~Dong}
\author[1,*]{Fengyi~Wu}
\author[1,*]{Yetong~Sha}
\author[1]{Lingdong~Kong}
\author[1]{Wei~Chow}
\author[2]{Qi~Dai}
\author[3]{Jiachen~Li}
\author[1]{Karen~Leung}
\author[1,\dagger]{Zhi-Qi~Cheng}

\affiliation[1]{\afflogo[1.15em]{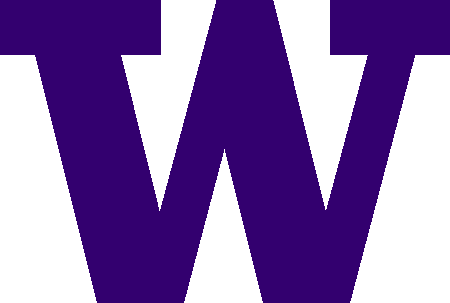}~University~of~Washington}
\affiliation[2]{\hspace{-0.25em}\smash{~\raisebox{0.1em}{\afflogo[1.3em]{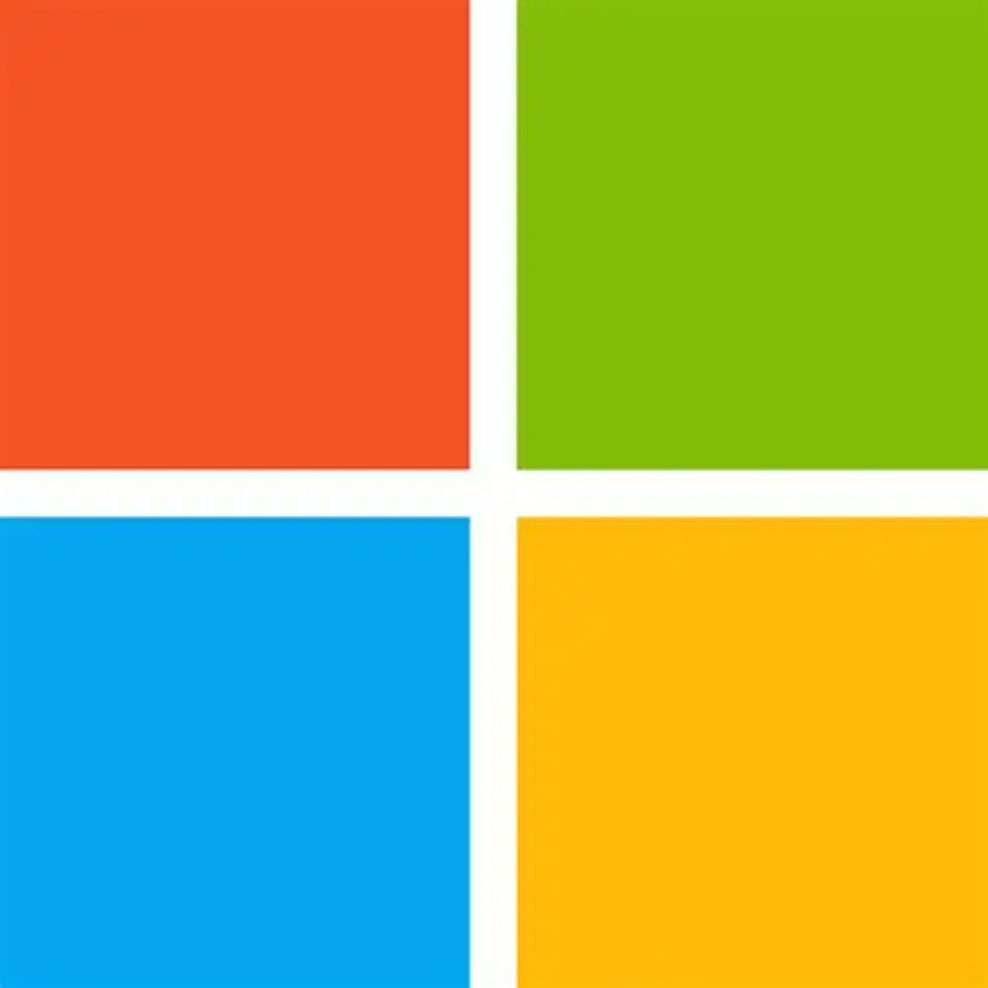}}}\hspace{-0.3em}~~Microsoft Research}
\affiliation[3]{\afflogo[1.4em]{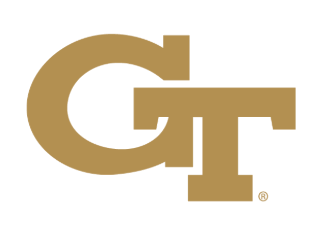}~Georgia~Institute~of~Technology}

\contribution[*]{Equal contribution}
\contribution[\dagger]{Corresponding author.}

\abstract{
World models support decision-making by predicting the consequences of actions. In sequential action-then-observation generation, however, the action remains fixed as its imagined outcome unfolds, preventing emerging visual evidence from revising that decision. To address this, we introduce the Reciprocal World Model (\ours{}), which enables planning and imagination to refine each other within a rollout step. \ours{} jointly generates reasoning, actions, and observations through masked discrete diffusion. To support reciprocal refinement, two-stage training first develops action-conditioned observation prediction and action inference from visual context, then learns joint generation. At inference, Reciprocal Decoding (RecD) brings these capabilities together to reassess committed actions and visual tokens under modified masks, selectively re-masking unsupported predictions for regeneration. Experiments on instruction-guided world modeling across grid-based gaming and egocentric navigation show improvements in planning, visual prediction, action--observation consistency, and reasoning grounding. Adding reciprocal checks to joint decoding raises average planning success from 70.9\% to 79.1\% and reduces final-frame DreamSim from 0.155 to 0.130.~A paired step-level analysis further shows that triggered revisions yield a net improvement in action--observation consistency.
}
  
\date{November 26, 2026}

\metadata[Code]{\url{https://github.com/F1y1113/RecWM}}

\begin{document}

\maketitle
   
\section{Introduction}
\label{sec:intro}

Building agents that can anticipate future states and act accordingly is a long-standing goal in artificial intelligence~\citep{ha2018world,hafner2019dream,hafner2022masteringataridiscreteworld,hafner2023mastering}. World models support this goal by predicting the consequences of actions, enabling agents to learn and plan through simulated experience~\citep{hafner2023mastering,bar2025navigation,survey_3d_4d_world_models,yang2026quovadis,chu2026agentic}. For example, when mentally rehearsing how to carry an object across a cluttered room, anticipating a collision may prompt a revised movement. This illustrates a reciprocal relationship: planning guides what to imagine, while imagination informs how to act.

\begin{figure}[!t]
    \centering
    \includegraphics[width=1\linewidth]{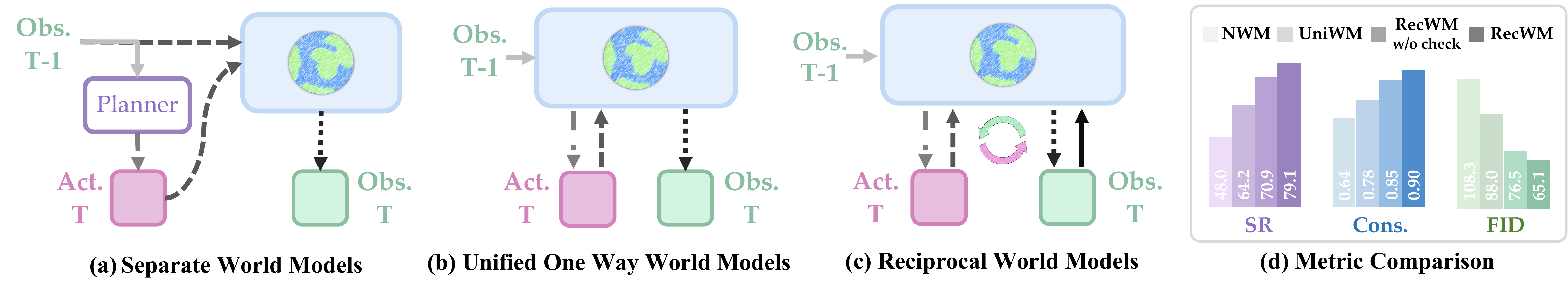}
    \caption{\textbf{Comparison of world model paradigms.} (a) A separate planner selects actions, and a world model predicts their outcomes~\citep{bar2025navigation}.~(b) A unified model alternates between action prediction and observation generation~\citep{dong2026towards}.~(c) Reciprocal World Model allows actions and imagined observations to refine each other.~(d) Scenario average results on planning success (SR$\uparrow$), action--observation consistency (Cons.$\uparrow$), and rollout visual quality (FID$\downarrow$).}
    \label{fig:paradigm}
\end{figure}

Existing approaches connect planning and imagination in different ways. Modular approaches use a world model to simulate actions proposed by a separate planner or policy~\citep{koh2021pathdreamer,bar2025navigation} (Fig.~\ref{fig:paradigm}a). To reduce discrepancies between separately learned planning and prediction components~\citep{dong2026towards}, recent work integrates action and observation prediction within a shared network~\citep{dong2026towards,cen2025worldvla}. Related multimodal reasoning methods ground intermediate reasoning in generated images~\citep{li2025imagine} or predicted visual latents~\citep{li2026toward}. However, shared representations or joint generation alone do not ensure that emerging visual predictions can revise committed actions within the current rollout step. In sequential action-then-observation generation, for example, the action remains fixed as its imagined outcome emerges: visual predictions can guide subsequent decisions but cannot revise the current action (Fig.~\ref{fig:paradigm}b). An incorrect action at a critical step can jeopardize the success of the entire task.

To enable this feedback during generation, we propose the \textbf{Reciprocal World Model} (\ours{}) (Fig.~\ref{fig:paradigm}c). Our key insight is that actions provide context for predicting visual outcomes, while those outcomes provide evidence for reconsidering the actions themselves. This interaction allows planning and imagination to refine each other before the current rollout step is finalized. \ours{} represents reasoning, actions, and observations in a shared sequence of discrete tokens and generates them through masked discrete diffusion. Two-stage training prepares the model for both joint generation and reciprocal revision. The first stage learns to predict visual outcomes from actions and infer actions from visual changes, including partially revealed next frames; the second stage learns to generate reasoning, actions, and observations jointly. At inference, Reciprocal Decoding (RecD) uses these capabilities to reassess committed predictions: \textcolor{recreverse}{\texttt{action checks}} evaluate actions against emerging visual evidence, while \textcolor{recforward}{\texttt{imagine checks}} evaluate visual tokens under the updated context. Predictions with insufficient support are re-masked and regenerated.

We evaluate \ours{} on instruction-guided world modeling, where models generate actions and corresponding future observations from initial observations and a language instruction. Across episodes spanning grid-based gaming and egocentric navigation, comparisons with multimodal models and world models show improvements in planning, visual prediction, and action--observation consistency (Fig.~\ref{fig:paradigm}d), together with stronger reasoning grounding. Training ablations assess the contributions of directional training and masking designs. At a fixed checkpoint, adding reciprocal checks to joint decoding improves planning success and visual prediction. Further analyses characterize check activation, while paired comparisons from the same decoding state show net consistency gains from triggered revisions. In summary, our contributions are threefold:

\noindent\textbf{(1)} We introduce the \textbf{Reciprocal World Model} (\ours{}), which enables reciprocal refinement of actions and imagined visual outcomes during generation.

\noindent\textbf{(2)} We propose a two-stage training scheme that learns forward and reverse prediction before joint generation, together with Reciprocal Decoding, which uses these capabilities for reciprocal checks with selective re-masking.

\noindent\textbf{(3)} Experiments across gaming and egocentric navigation show gains in planning, visual prediction, and action--observation consistency. Ablations support the benefits of directional training and reciprocal checks, while paired analyses quantify net consistency gains from triggered revisions.

\begin{figure*}
    \centering
    \includegraphics[width=1\linewidth]{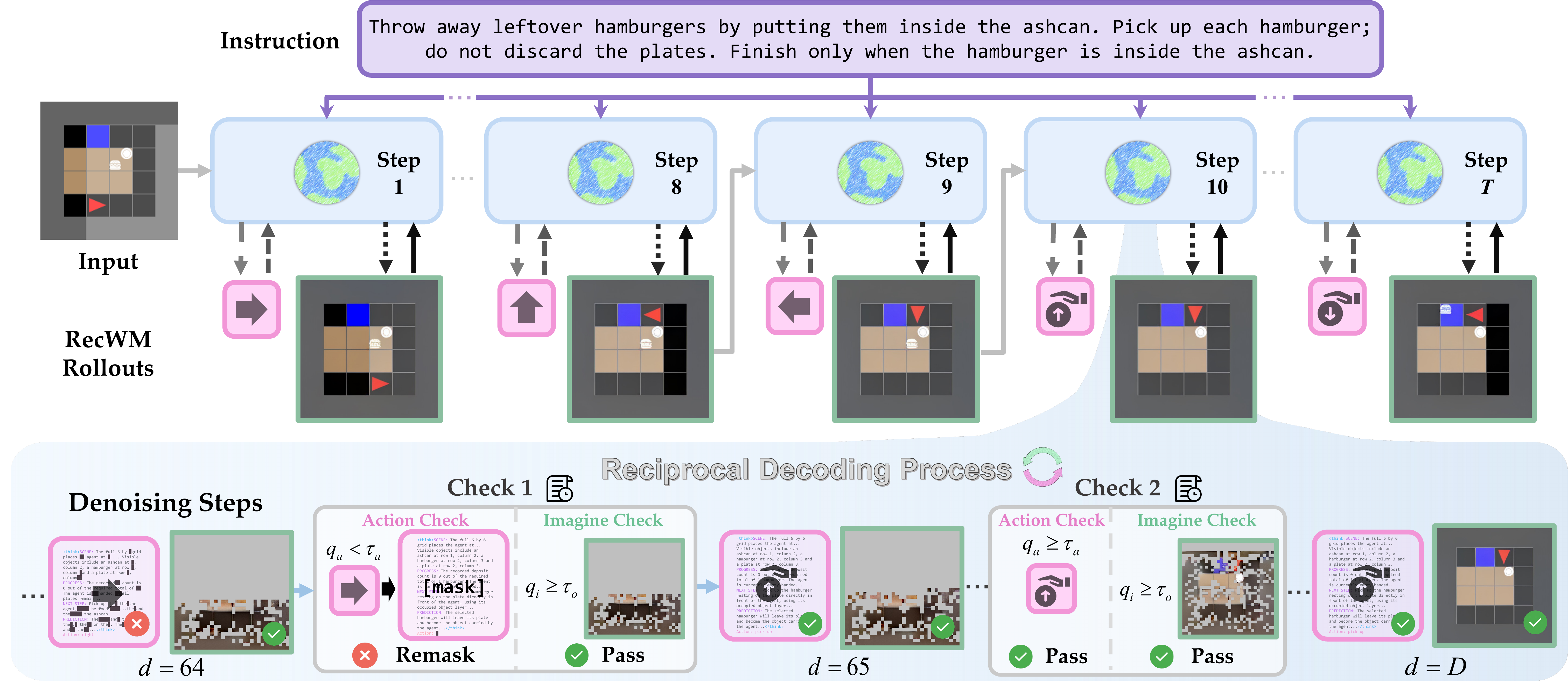}
    \caption{\textbf{Rollout and Reciprocal Decoding in \ours{}.} Top: instruction-guided world modeling in egocentric-grid gaming.~Given initial observations and a language instruction, \ours{} generates actions and future observations, using generated observations as context for later steps. Bottom: Reciprocal Decoding within one rollout step. Using the partially generated observation, \textcolor{recreverse}{\texttt{action check}} rejects and re-masks a committed move-right action. The action is regenerated as a pickup action, and subsequent denoising refines the observation.}
    \label{fig:task_diff}
\end{figure*}
\section{Related Work}
\label{sec:related}

\noindent \textbf{World Models} capture environment dynamics and predict future states through diverse representations and generative frameworks~\citep{xu2025U4D,survey_3d_4d_world_models,chi2026driver,chi2026pv}, including compact latent models~\citep{ha2018world,hafner2019dream,hafner2022masteringataridiscreteworld,hafner2023mastering,hafner2019learning,hafner2025training,gumbsch2023learning}, feature-space predictors~\citep{baldassarre2025back,bardes2024revisiting,assran2025v}, and large-scale generative systems~\citep{brooks2024video,agarwal2025cosmos,bruce2024genie,google2024genie2,google2025genie3,russell2025gaia,liao2025envisioner}. Action-conditioned generators further support interactive simulation and visual planning~\citep{alonso2024diffusion,valevski2024diffusion,bar2025navigation,yu2025gamefactory,xiao2025worldmem,zhang2025epona,yao2025navmorph,hu2026navthinker,wu2026tiresias}. Guidance during diffusion sampling also supports trajectory planning~\citep{janner2022planning,chen2024diffusion} and temporally consistent video prediction~\citep{song2025history}. Recent work integrates prediction and decision-making within a shared model~\citep{dong2026towards,liu2026palm,li2026lingbotva}. This integration takes different forms, including alternating action prediction and observation generation~\citep{dong2026towards}, interleaving verbal reasoning with generated visualizations~\citep{li2025imagine,wu2026govig}, and supporting forward and inverse dynamics through input masking~\citep{li2025unified}. Our work uses forward and reverse predictions to revise actions and imagined observations within the same rollout step.

\noindent \textbf{Unified Multimodal Models} support both understanding and generation of text and images within a single network, providing backbones for world models that reason in language while imagining future frames. Autoregressive models generate discrete visual tokens through next-token prediction~\citep{wu2025janus,liu2026lumina,wang2024emu3,chen2025januspro,wu2025vilau}, while hybrid models combine autoregressive text generation with diffusion-based image generation~\citep{wu2025omnigen2,cao2025hunyuanimage,deng2025emerging,zhou2025transfusion,xie2025showo2}. Standard autoregressive decoding does not revise committed tokens within a decoding pass, although generated images can guide subsequent text or additional generation rounds~\citep{wu2025omnigen2}. Masked diffusion provides an alternative by predicting tokens in parallel using bidirectional context~\citep{lou2023discrete,sahoo2024simple,austin2021d3pm,nie2025llada,chang2023muse,you2025llada}. MMaDA~\citep{yangmmada} and Lumina-DiMOO~\citep{xin2025lumina} extend this to unified text and image modeling. MMaDA-Parallel~\citep{tian2025mmada} enables concurrent text--image denoising and improves alignment through trajectory-level reinforcement learning.~\ours{} extends cross-modal interaction to explicit reassessment of committed actions and visual tokens, using learned forward and reverse capabilities to guide re-masking and regeneration.
\section{Task Formulation}
\label{sec:tasks}

\begin{figure*}[!t]
    \centering
    \includegraphics[width=\linewidth]{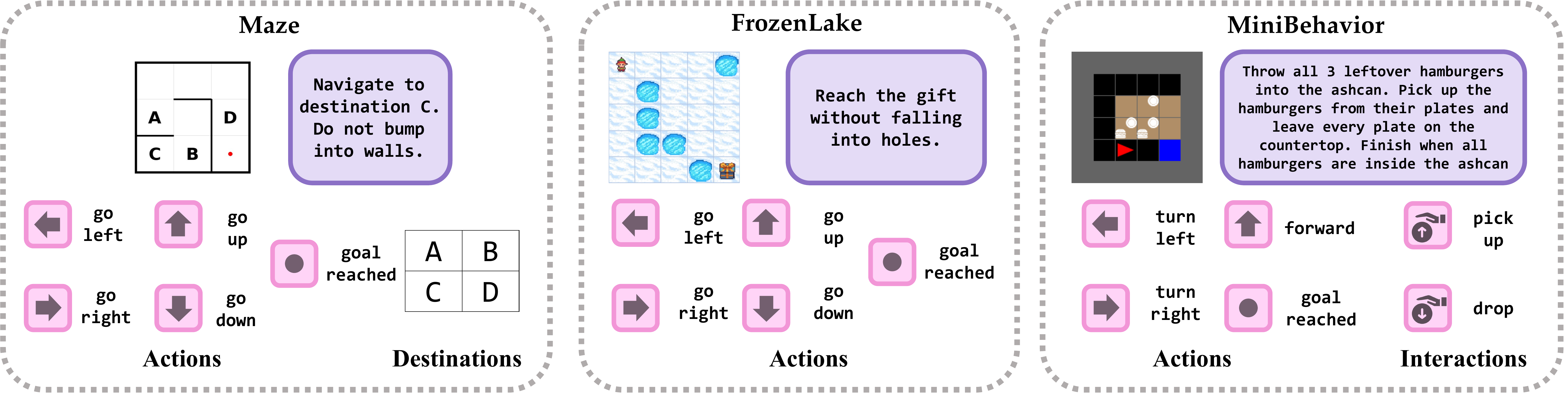}
    \caption{\textbf{Instructions and action types in global-grid gaming.} Each panel shows a global observation, an example instruction, and the action types for the illustrated task. The MiniBehavior example shows the hamburger disposal task.}
    \label{fig:scenarios}
\end{figure*}

Given an initial observation history $\mathbf{o}_0$ of length $H$ and a language instruction $l$, instruction-guided world modeling requires generating an action sequence $(\hat{a}_0,\ldots,\hat{a}_{T-1})$ to achieve the specified goal, together with the corresponding future observations $(\hat{o}_1,\ldots,\hat{o}_T)$, where $T$ denotes the rollout length. As illustrated in Fig.~\ref{fig:task_diff} (top), the task requires both goal-directed planning and consistency between actions and their predicted visual outcomes. We evaluate this task across three scenarios: global-grid gaming, egocentric-grid gaming, and egocentric navigation. Fig.~\ref{fig:scenarios} illustrates the observations, instructions, and action types in the three global-grid gaming environments: Maze, FrozenLake, and MiniBehavior.

Each observation $o\in\mathbb{R}^{h\times w\times3}$ is an RGB image of height $h$ and width $w$. Each gaming action $a$ is a discrete index identifying an available action in the corresponding environment, while navigation actions are represented as $a=(\mathbf{u},\phi)\in\mathbb{R}^3$, where $\mathbf{u}\in\mathbb{R}^2$ specifies planar displacement and $\phi\in\mathbb{R}$ specifies the yaw change. In \ours{}, we encode observations as $s=\operatorname{Enc}(o)$ and use a single model $f_\theta$ for both action planning and future-state prediction:
\begin{equation}
    \hat{a}_t \sim f_\theta\big(\cdot\mid\hat{\mathbf{s}}_t,l\big), \quad
    \hat{s}_{t+1} \sim f_\theta
    \big(\cdot\mid\hat{\mathbf{s}}_t,l,\hat{a}_t\big),
    \label{eq:recwm}
\end{equation}
where $t=0,\ldots,T-1$ indexes the environment step. Context $\hat{\mathbf{s}}_t$ retains the latest $K$ state representations from initial and generated observations and is updated after each step.

Unlike modular approaches that separate planning from visual prediction~\citep{bar2025navigation}, this formulation learns both functions within a shared model. Separate training can introduce discrepancies between the planner's state assumptions and the world model's predictions~\citep{dong2026towards}, motivating this unified formulation.  We further describe how two-stage training and reciprocal checks enable planning and imagination to refine each other during generation in the next section.


\section{Method}
\label{sec:method}

We present the Reciprocal World Model (\ours{}), which enables reciprocal refinement between planning and imagination within a single discrete diffusion model. We describe its architecture and token representation (Sec.~\ref{sec:architecture}), two-stage training scheme (Sec.~\ref{sec:training}), and Reciprocal Decoding (RecD) procedure (Sec.~\ref{sec:decoding}). Fig.~\ref{fig:training_inference} summarizes how RecD uses the capabilities learned through two-stage training for joint generation and reciprocal checks at inference.

\begin{figure*}[!t]
\centering
\begin{promptbox}{Gaming -- Maze}

\promptsection{Input (context prefix)}

\promptfield{System prompt}{[SYS]}

You are an agent navigating a grid maze to the destination named in the instruction. Each observation shows the complete maze from above: black lines are walls, letters (A, B, C, D) mark the destinations, the small red dot marks your starting cell, the red line with arrows traces the route you have taken so far, and the red cross marker at the end of the line is your current position.

Cells are indexed as [row, col] with 0-based indexing. [0, 0] is the top-left cell; row increases downward and col increases rightward. Refer to positions with these coordinates.

Choose one action per step. \acttok{go\_up} moves to [row-1, col]; \acttok{go\_down} moves to [row+1, col]; \acttok{go\_left} moves to [row, col-1]; \acttok{go\_right} moves to [row, col+1]. \acttok{goal\_reached} is the terminal action and is used only when the red marker is on the destination cell.

Locate your cell and the requested destination letter. Move to a neighboring cell only if it is inside the maze and no wall blocks the passage. Use the red trail to recognize visited cells and avoid re-entering dead ends.

After a move, the red line extends by one arrow into the neighboring cell and the red marker moves to its new end, while the red dot, walls, and letters stay unchanged.

Write the reasoning in four slots. In SCENE, describe visible objects and their layout. In PROGRESS, summarize what is complete and what remains. In NEXT STEP, state the action and its expected effect. In PREDICTION, describe the next observation. Preserve any supplied answer content and complete the rest consistently.

\medskip
\begin{minipage}[t]{0.35\linewidth}
    \vspace{0pt}
    \raggedright
    \promptfield{Task header and instruction}{[INSTR]}\par
    Complete the reasoning, action, and next observation for the current step.\par
    \textbf{Instruction:} Navigate to destination C. Do not bump into walls.
\end{minipage}\hfill
\begin{minipage}[t]{0.63\linewidth}
    \vspace{0pt}
    \centering
    \promptfield{Context observations}{[BOI] $\cdots$ [EOI]}\par
    \smallskip
    \promptobs{$o_{t-3}$}{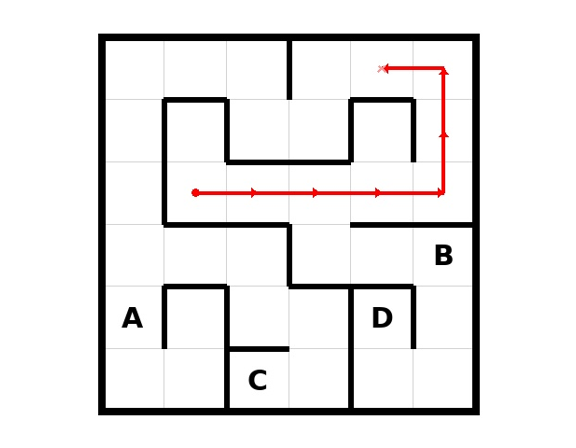}\hfill
    \promptobs{$o_{t-2}$}{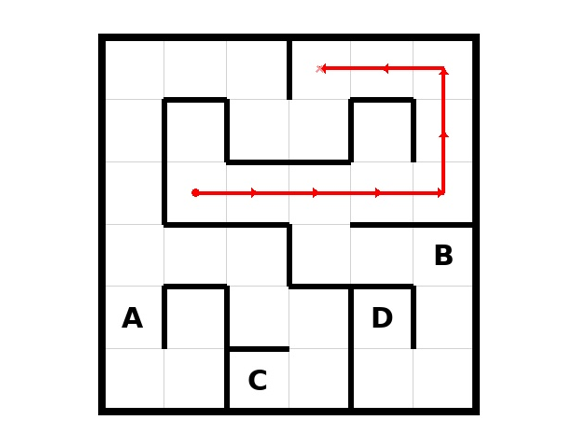}\hfill
    \promptobs{$o_{t-1}$}{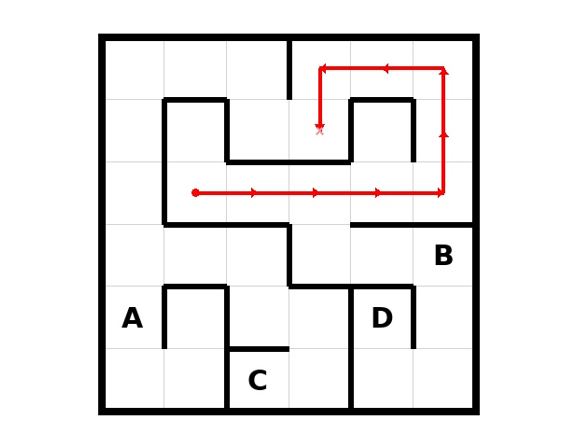}\hfill
    \promptobs{$o_t$}{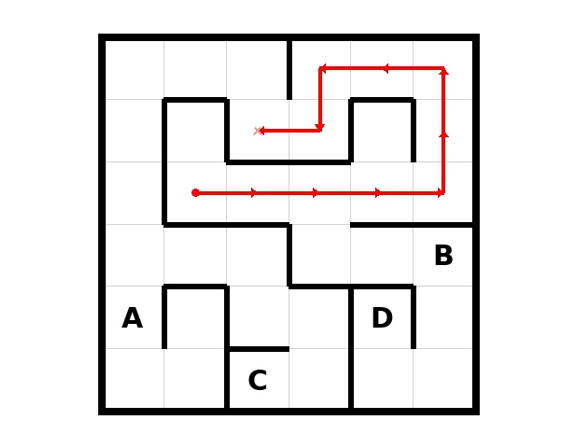}
\end{minipage}\par

\promptsection{Output (answer region)}

\begin{minipage}[t]{0.60\linewidth}
    \vspace{0pt}
    \raggedright
    \promptfield{Reasoning}{\textless think\textgreater\ $\cdots$ \textless/think\textgreater}\par
    \promptslot{SCENE}{Looking at my current view, I am at [1, 2] on a 6x6 maze grid. I can see destinations A at [4, 0], B at [3, 5], C at [5, 2], D at [4, 4].}\par
    \promptslot{PROGRESS}{I moved left from [0, 4] to [0, 3], then down to [1, 3] and left again to [1, 2]. I still need to reach destination C at [5, 2].}\par
    \promptslot{NEXT STEP}{Go up to move from [1, 2] to [0, 2].}\par
    \promptslot{PREDICTION}{After Go up, the red marker will be at [0, 2] with the path line extended by one segment from [1, 2] to [0, 2].}\par
    \medskip
    \promptfield{Action:}{}\acttok{go\_up}
\end{minipage}\hfill
\begin{minipage}[t]{0.37\linewidth}
    \vspace{0pt}
    \centering
    \promptfield{Next observation}{[BOI] img$_{t+1}$ [EOI]}\par
    \smallskip
    \promptobs{$o_{t+1}$}{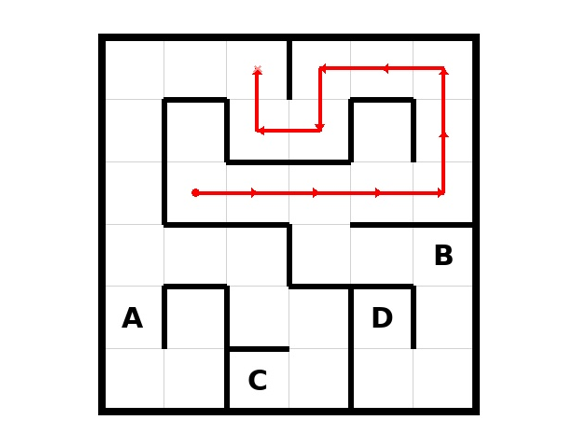}
\end{minipage}\par

\end{promptbox}
\caption{\textbf{Prompt and answer example for global-grid gaming (Maze).} The context prefix contains the system prompt, task instruction, and observations ordered from oldest to current. The answer contains four reasoning slots, one action token, and the next observation. The training mode determines which answer tokens are masked and supervised.}
\label{fig:maze_prompt}
\end{figure*}

\subsection{Architecture and Unified Token Representation}
\label{sec:architecture}

We focus on multimodal masked diffusion models that predict both text and image tokens using bidirectional context~\citep{yangmmada,xin2025lumina,tian2025mmada}. We choose Lumina-DiMOO~\citep{xin2025lumina} as our backbone for its multimodal understanding and generation capabilities. We denote the model by $f_\theta$ and its token probabilities by $p_\theta$. We structure the inputs and outputs (reasoning, actions, and observations) of $f_\theta$ through a unified vocabulary and sequence layout, as detailed below.

\noindent \textbf{Unified vocabulary.}~Following prior work~\citep{team2024chameleon,xieshow,yangmmada}, we represent text, actions, and images using a shared discrete vocabulary $\mathcal{V} = \mathcal{V}_T \cup \mathcal{V}_A \cup \mathcal{V}_I$. Here, $\mathcal{V}_I$ contains $C$ visual codebook entries from a pretrained aMUSEd-VQ tokenizer~\citep{patil2024amused}, and $\mathcal{V}_T$ contains $V_T$ text tokens from a BPE tokenizer~\citep{xin2025lumina}. Gaming actions are represented by dedicated tokens in $\mathcal{V}_A$, while navigation actions are expressed as text and encoded with the same BPE tokenizer.

\noindent \textbf{Sequence layout.}~At environment step $t$, the model receives $K_t=\min(K,H+t)$ context frames, comprising the most recent frames from the initial observations and previously generated observations. The full sequence consists of a context prefix followed by an answer region:
\begin{equation}
\underbrace{
\texttt{[SYS]\,[INSTR]}\;
\bigl[\texttt{[BOI]\,img}_j\texttt{\,[EOI]}\bigr]_{j=1}^{K_t}
}_{\text{context prefix}}
\;
\underbrace{
\texttt{[BOA]}\,\fbox{\texttt{answer}}\,\texttt{[EOA]}
}_{\text{answer region}}.
\label{eq:seq_layout}
\end{equation}
Here, $j$ indexes the context frames in chronological order. \texttt{[SYS]} denotes the system-prompt span, and \texttt{[INSTR]} contains the task header and instruction $l$. The tokens \texttt{[BOI]/[EOI]} delimit images, while \texttt{[BOA]/[EOA]} delimit the answer region. The prefix remains unmasked, and masking and supervision are restricted to the answer region. The answer follows the template
\begin{equation}
\fbox{\texttt{answer}} =
\texttt{<think>\,reasoning\,</think>}\;
\texttt{action\,[BOI]\,img}_{t+1}\texttt{\,[EOI]}.
\label{eq:answer}
\end{equation}
The \texttt{reasoning} span contains four slots: SCENE describes the visible scene and spatial layout; PROGRESS summarizes task-relevant state changes; NEXT STEP specifies the intended action and its expected effect; and PREDICTION describes the anticipated next observation. The \texttt{action} span contains a single gaming action token from $\mathcal{V}_A$ or navigation action text encoded with $\mathcal{V}_T$, and \texttt{img}$_{t+1}$ contains visual tokens of the next observation. Fig.~\ref{fig:maze_prompt} illustrates this sequence layout with a Maze example in the global-grid gaming setting, showing the system prompt, task instruction, context observations, and an example answer. Additional prompt examples for the other environments and observation settings are provided in Appendix~\ref{sec:app-instruction-scenarios}.

\begin{figure}[!t]
    \centering
    \includegraphics[width=0.65\linewidth]{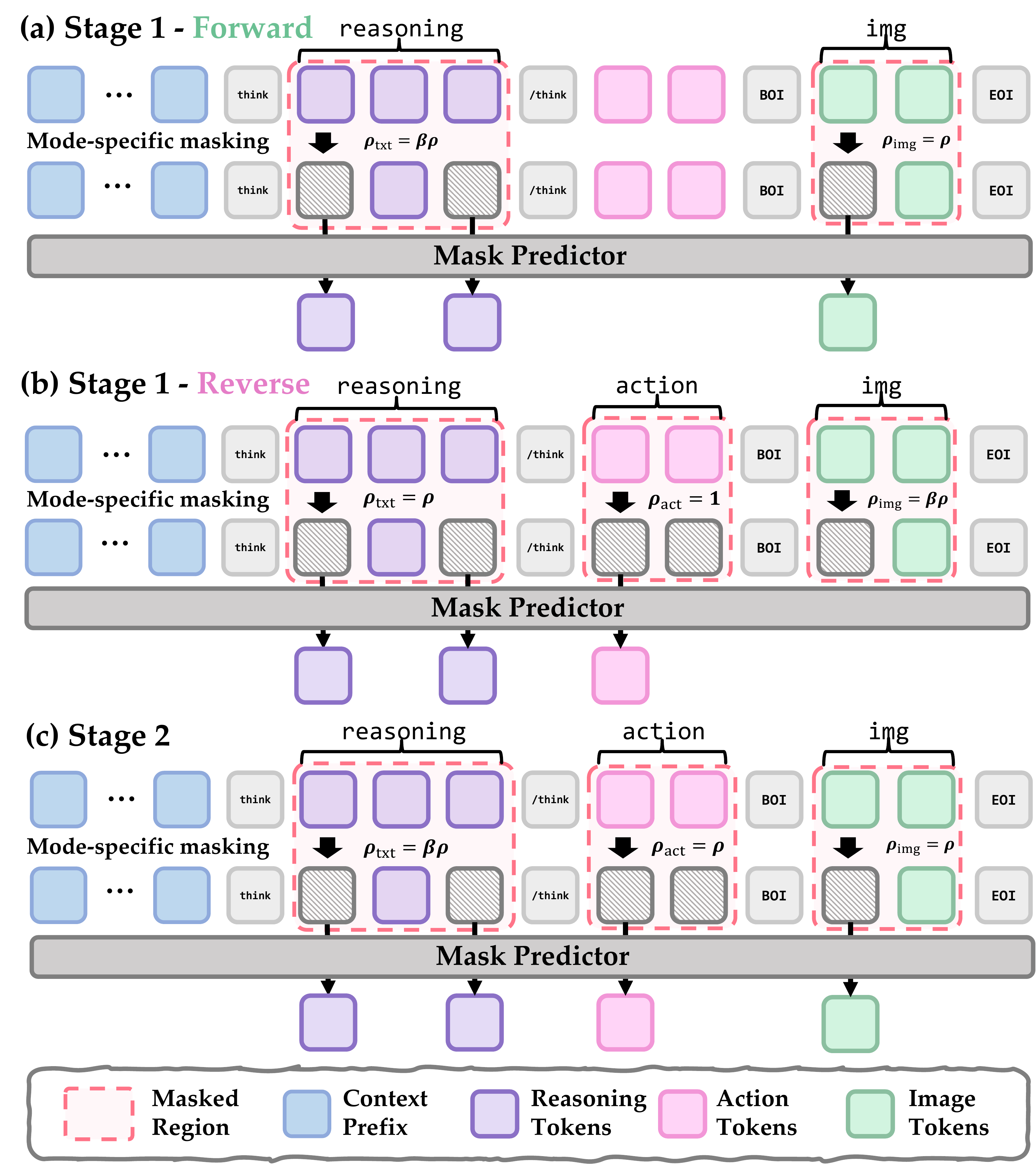}
    \caption{\textbf{Two-stage training of \ours{}.} Stage~1 reconstructs masked tokens in two modes: (a) forward mode predicts reasoning and next-frame tokens conditioned on the action; (b) reverse mode predicts reasoning and action tokens conditioned on the next frame. (c) Stage~2 jointly reconstructs all three regions.}
    \label{fig:training}
\end{figure}

\subsection{Two-Stage Training}
\label{sec:training}

In this section, we define three training modes with different prediction regions. Stage~1 learns forward observation prediction and reverse action inference; Stage~2 learns joint generation of reasoning, actions and observations as shown in Figs.~\ref{fig:training} and \ref{fig:training_inference}a.

\noindent \textbf{Training objective.}~We train the model to reconstruct masked answer tokens conditioned on the visible context. For each clean sequence $\mathbf{x}\in\mathcal{V}^{L}$, where $L$ denotes the sequence length, we sample $r\sim\mathcal{U}(0,1)$ and compute the base masking ratio $\rho=\cos(\pi r/2)$. We obtain $\tilde{\mathbf{x}}$ by replacing selected answer tokens with \texttt{[MASK]} according to $\rho$ and mode-specific rules. To balance the contributions of regions with different lengths, we define the total training objective $\mathcal{L}(\theta)$ as the sum of regional losses $\mathcal{L}_p$, where each $\mathcal{L}_p$ is the expected mean masked-token cross-entropy for region $p$:
\begin{equation}
\mathcal{L}_p
= -\mathbb{E}_{\mathbf{x},m,\tilde{\mathbf{x}}}
\left[
    \frac{\sum_{i \in \mathcal{M}_p}
    \log p_\theta(x_i \mid \tilde{\mathbf{x}})}
    {\max(1,|\mathcal{M}_p|)}
\right],
\qquad
\mathcal{L}(\theta)
= \sum_{p \in \{\text{txt},\,\text{act},\,\text{img}\}} \mathcal{L}_p,
\label{eq:loss}
\end{equation}
where $m$ denotes the training mode, set $\mathcal{M}_p$ contains supervised masked positions in $p$, and $p_\theta(x_i\mid\tilde{\mathbf{x}})$ is the probability assigned to the original token $x_i$ at position $i$. Regions with no supervised masked positions contribute zero loss.

\begin{figure}[t]
    \centering
    \small
    \recflushalgorithmic
    \setlength{\tabcolsep}{0pt}
    \renewcommand{\arraystretch}{1}

    \begin{tabular}{
        @{}p{0.45\columnwidth}
        @{\hspace{0.10\columnwidth}}
        p{0.45\columnwidth}@{}
    }

    \rule{0.45\columnwidth}{1pt}
&
\rule{0.45\columnwidth}{1pt}
\\
    \begin{minipage}[t]{\linewidth}
        \raggedright
        \textbf{(a) Two-stage training (Sec.~\ref{sec:training})}
        \par\vspace{1pt}\hrule\vspace{2pt}

        \noindent
        \parbox[t]{\linewidth}{%
            \raggedright
            \setlength{\parindent}{0pt}
            \textbf{Input:} Sequences $\mathbf{x}$ containing
            reasoning, action, and image regions
            $p_{\text{txt}},p_{\text{act}},p_{\text{img}}$;
            pretrained model $f_\theta$;
            masking ratios
            $\rho_{\text{txt}},\rho_{\text{act}},\rho_{\text{img}}$;
            base ratio $\rho$ and asymmetry factor $\beta$.
        }
        \par\vspace{4pt}

        \begin{algorithmic}
            \Statex \textbf{Stage 1}
        \end{algorithmic}

        \begin{recblock}{recforward}
            \begin{algorithmic}
                \rectrainmark{rec-train-f}{recforward}
                    {Forward mode}

                \Statex
                    $\rho_{\text{act}}\!=\!\!0,\!\;
                     \rho_{\text{txt}}\!=\!\beta\rho,\!\!\;
                     \rho_{\text{img}}\!=\!\!\rho$~(Eq.~\ref{eq:causal_corruption})

                \State $\mathcal{L}\gets
                    \mathcal{L}_{\text{txt}}
                    +\mathcal{L}_{\text{img}}$

                \Statex \textcolor{black!55}{%
                    $\triangleright$ Mean masked CE on
                    $p_{\text{txt}},p_{\text{img}}$}

                \State Update $\theta$ to minimize $\mathcal{L}$
                    \tikzmarknode{rec-train-f-end}{\strut}
            \end{algorithmic}
        \end{recblock}
    \end{minipage}
    &
    \begin{minipage}[t]{\linewidth}
        \raggedright
        \textbf{(b) Reciprocal Decoding (Sec.~\ref{sec:decoding})}
        \par\vspace{1pt}\hrule\vspace{2pt}

        \noindent
        \parbox[t]{\linewidth}{%
            \raggedright
            \setlength{\parindent}{0pt}
            \textbf{Input:} Sequence $\tilde{\mathbf{x}}$
            containing masked regions
            $p_{\text{txt}},p_{\text{act}},p_{\text{img}}$;
            trained model~$f_\theta$;
            denoising steps~$D$; warm-up ratio $\alpha$;
            thresholds $\tau_c,\tau_a,\tau_o$;
            scores $q_a$, $q_i$.
        }
        \par\vspace{4pt}

        \begin{algorithmic}
            \Statex \textbf{for} $d=1,\ldots,D$ \textbf{do}
        \end{algorithmic}

        \noindent\hspace*{0.6em}%
        \begin{minipage}[t]{\dimexpr\linewidth-0.6em\relax}
            \begin{recblock}[colframe=recjoint!25]{recjoint}
                \begin{algorithmic}
                    \Statex
                        \tikzmarknode{rec-infer-j}
                            {Denoise $p_{\text{txt}},p_{\text{act}},p_{\text{img}}$}

                    \State Reveal $p_{\text{txt}}$ on a linear schedule

                    \State Commit $p_{\text{act}}$ at confidence
                        $\tau_c$

                    \State Reveal $p_{\text{img}}$
                        from $d=\lfloor\alpha D\rfloor$
                        \tikzmarknode{rec-infer-j-end}{\strut}
                \end{algorithmic}
            \end{recblock}
        \end{minipage}
        \par

        \begin{algorithmic}
            \Statex \hspace*{0.6em}\textbf{if}
                $d\in\{\lfloor D/2\rfloor,\lfloor3D/4\rfloor\}$
                \textbf{then}
        \end{algorithmic}
    \end{minipage}
    \\[3pt]

    \begin{minipage}[t]{\linewidth}
        \raggedright

        \begin{recblock}{recreverse}
            \begin{algorithmic}
                \rectrainmark{rec-train-r}{recreverse}
                    {Reverse mode}

                \Statex
                    $\rho_{\text{act}}\!=\!1,\;
                     \rho_{\text{txt}}\!=\!\rho,\;
                     \rho_{\text{img}}\!=\!\beta\rho$

                \State $\mathcal{L}\gets
                    \mathcal{L}_{\text{txt}}
                    +\mathcal{L}_{\text{act}}$ 

                \State Update $\theta$ to minimize $\mathcal{L}$
                    \tikzmarknode{rec-train-r-end}{\strut}
            \end{algorithmic}
        \end{recblock}

        \begin{algorithmic}
            \Statex \textbf{Stage 2}
            \State Continue from Stage~1 weights
        \end{algorithmic}

        \begin{recblock}[colframe=recjoint!25]{recjoint}
            \begin{algorithmic}
                \rectrainmark{rec-train-j}{recjoint}
                    {Joint mode}

                \Statex
                    $\rho_{\text{act}}\!=\!\rho,\;
                     \rho_{\text{txt}}\!=\!\beta\rho,\;
                     \rho_{\text{img}}\!=\!\rho$

                \State $\mathcal{L}\gets
                    \mathcal{L}_{\text{txt}}
                    +\mathcal{L}_{\text{act}}
                    +\mathcal{L}_{\text{img}}$~(Eq.~\ref{eq:loss})

                \State Update $\theta$ to minimize $\mathcal{L}$
                    \tikzmarknode{rec-train-j-end}{\strut}
            \end{algorithmic}
        \end{recblock}

        \begin{algorithmic}
            \Statex \textbf{Output:} Trained model $f_\theta$
        \end{algorithmic}
    \end{minipage}
    &
    \begin{minipage}[t]{\linewidth}
        \raggedright

        \noindent\hspace*{1.2em}%
        \begin{minipage}[t]{\dimexpr\linewidth-1.2em\relax}

            \begin{recblock}{recreverse}
                \begin{algorithmic}
                    \recinfermark{rec-infer-r}{recreverse}
                        {Action check}

                    \State Mask $p_{\text{act}}$ and compute $q_a$

                    \Statex \textbf{if} $q_a<\tau_a$
                        \textbf{then} keep masked

                    \Statex \textbf{else} restore the original action
                        \tikzmarknode{rec-infer-r-end}{\strut}
                \end{algorithmic}
            \end{recblock}

            \begin{recblock}{recforward}
                \begin{algorithmic}
                    \recinfermark{rec-infer-f}{recforward}
                        {Imagine check}

                    \State Mask $p_{\text{img}}$ and compute $q_i$

                    \State Restore tokens with $q_i\geq\tau_o$
                        \tikzmarknode{rec-infer-f-end}{\strut}
                \end{algorithmic}
            \end{recblock}

        \end{minipage}
        \par

        \begin{algorithmic}
            \Statex \hspace*{0.6em}\textbf{end if}
            \Statex \textbf{end for}
            \Statex \textbf{Output:} Reasoning, action,
                and next observation from $\tilde{\mathbf{x}}$
        \end{algorithmic}
    \end{minipage}
    \\
    \rule{0.44\columnwidth}{1pt}
    &\rule{0.46\columnwidth}{1pt}
    \end{tabular}

    \begin{tikzpicture}[
        remember picture,
        overlay,
        rec brace/.style={
            decorate,
            decoration={brace,amplitude=2pt},
            line width=1pt
        },
        rec link/.style={
            dashed,
            dash pattern=on 2pt off 2pt,
            -{Stealth[length=1.1mm]},
            line width=1pt,
            rounded corners=1pt,
            preaction={
                draw=white,
                solid,
                line width=1pt,
                arrows=-
            }
        }
    ]

        \foreach \id/\col in {f/recforward,r/recreverse,j/recjoint} {

            \coordinate (T-\id-top) at
                ([xshift=3pt]rec-train-\id.north east);

            \coordinate (T-\id-bottom) at
                (T-\id-top |- rec-train-\id-end.south);

            \draw[rec brace,draw=\col]
                (T-\id-top) -- (T-\id-bottom);

            \coordinate (T-\id) at
                ($(T-\id-top)!0.5!(T-\id-bottom)+(2pt,0)$);

            \coordinate (I-\id-top) at
                ([xshift=-3pt]rec-infer-\id.north west);

            \coordinate (I-\id-bottom) at
                (I-\id-top |- rec-infer-\id-end.south);

            \draw[rec brace,draw=\col]
                (I-\id-bottom) -- (I-\id-top);

            \coordinate (I-\id) at
                ($(I-\id-top)!0.5!(I-\id-bottom)+(-2pt,0)$);
        }

        \coordinate (laneF) at ($(T-r)!0.25!(I-r)$);
        \coordinate (laneJ) at ($(T-r)!0.75!(I-r)$);

        \draw[rec link,draw=recforward]
            (T-f)
            -- (laneF |- T-f)
            -- (laneF |- I-f)
            -- (I-f);

        \draw[rec link,draw=recjoint]
            (T-j)
            -- (laneJ |- T-j)
            -- (laneJ |- I-j)
            -- (I-j);

        \draw[rec link,draw=recreverse]
            (T-r) -- (I-r);

    \end{tikzpicture}

    \caption{\textbf{Pseudocode for two-stage training and RecD in
    \ours{}.} Training develops next-frame prediction and action
    inference through forward and reverse modes, followed by joint
    generation. During inference, the model jointly generates actions
    and observations, using the capabilities learned during two-stage
    training to reassess them through \textcolor{recreverse}{\texttt{action}} and \textcolor{recforward}{\texttt{imagine}} checks, re-masking unsupported predictions for regeneration.}
    \label{fig:training_inference}
\end{figure}

\noindent \textbf{Stage 1: directional modes.}~
Let $\rho_{\text{txt}}$ and $\rho_{\text{img}}$ denote token masking ratios for reasoning and images, respectively, and let $\rho_{\text{act}}$ denote probability of masking the entire action region.

\begin{itemize}
    \item The \emph{forward mode} (Fig.~\ref{fig:training}a) keeps the action visible ($\rho_{\text{act}}=0$) and reconstructs masked reasoning and next-frame tokens. We use asymmetric masking to preserve more reasoning context for visual reconstruction:
\begin{equation}
    \rho_{\text{txt}}=\beta\rho,\qquad
    \rho_{\text{img}}=\rho,\qquad
    \beta\sim\mathcal{U}(\beta_{\min},1),
    \label{eq:causal_corruption}
\end{equation}
where $\beta_{\min}\!\!\in\!(0,\!1)\!$ controls maximum masking asymmetry.

    \item The \emph{reverse mode} (Fig.~\ref{fig:training}b) reconstructs masked reasoning and action tokens using visual evidence from the current and next observations. All action tokens are masked ($\rho_{\text{act}}=1$), together with the action-revealing reasoning slots NEXT STEP and PREDICTION. The remaining reasoning tokens are masked at ratio $\rho$, while next-frame tokens are masked at ratio $\beta\rho$, with $\beta$ sampled as in forward mode. This provides visual context for action inference while exposing the model to incomplete next frames. Only masked reasoning and action tokens are supervised, and both directional modes use the same annotated samples.
\end{itemize}

\noindent \textbf{Stage 2: joint mode.}~Starting from Stage~1 weights, we train the model in \emph{joint mode} (Fig.~\ref{fig:training}c) to reconstruct reasoning, actions, and next frames jointly. Reasoning and image tokens follow Eq.~\ref{eq:causal_corruption}. We set $\rho_{\text{act}}=\rho$, masking entire action region with probability $\rho$ and otherwise keeping it fully visible.

\begin{figure}[!t]
    \centering
    \includegraphics[width=1\linewidth]{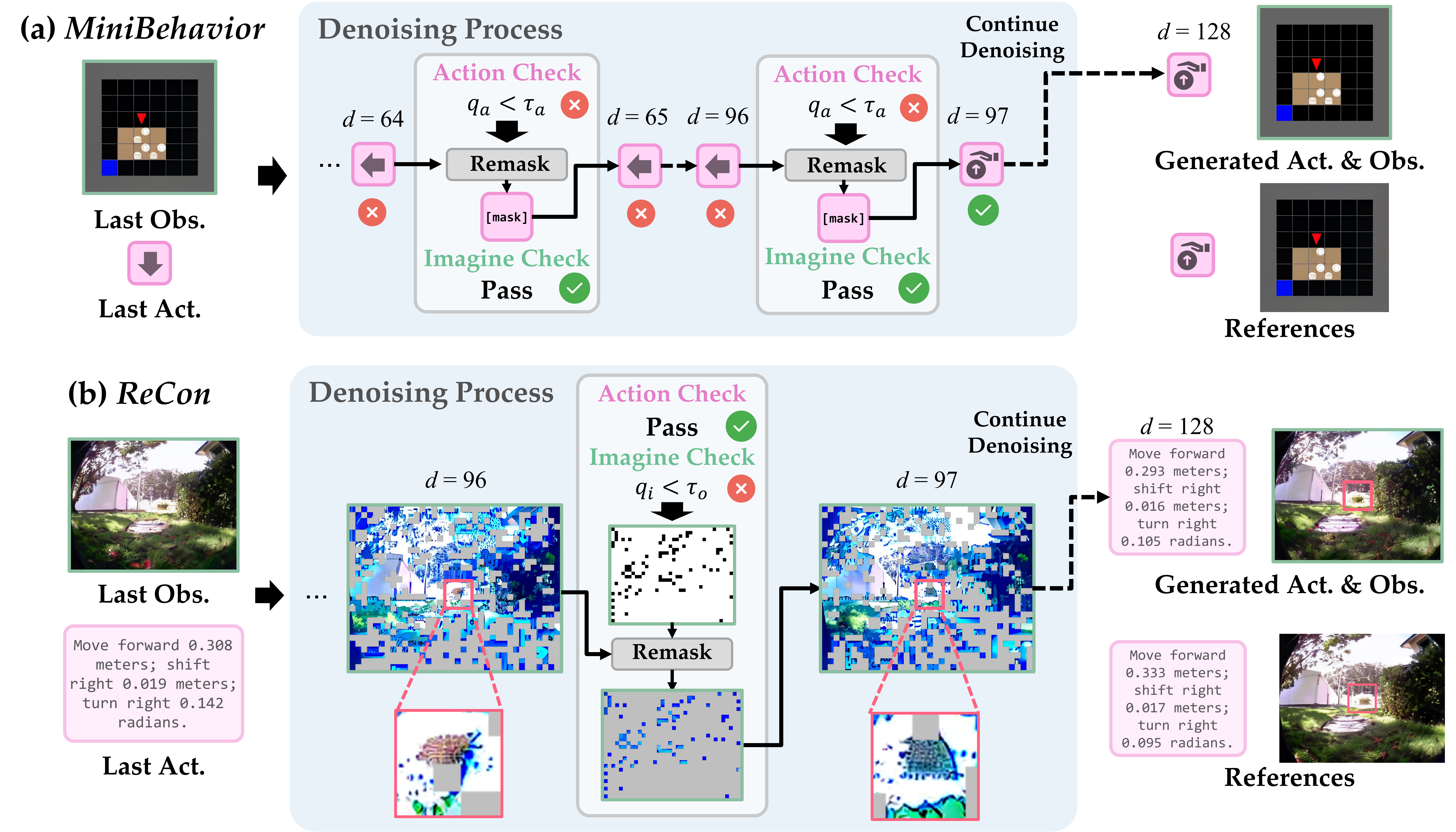}
    \caption{\textbf{Reciprocal checks during denoising.} (a) Both \textcolor{recreverse}{\texttt{action checks}} reject the move-left action, while \textcolor{recforward}{\texttt{imagine checks}} pass. Regeneration yields move-left after the first rejection and the correct pickup action after the second. (b) The action is retained, while low-support visual tokens are re-masked for refinement.}
    \label{fig:check}
\end{figure}

\subsection{Inference with Reciprocal Decoding}
\label{sec:decoding}

For inference, we propose Reciprocal Decoding (RecD), which supplements base denoising schedules with reciprocal checks using the forward and reverse prediction capabilities learned during two-stage training.

\noindent \textbf{Denoising schedules.}~We initialize the reasoning, action, and next-frame regions with masked tokens and generate them jointly over $D$ denoising steps ($d=1,\ldots,D$). Reasoning tokens follow a linear reveal schedule, with higher-confidence predictions committed first. Before action commitment, each denoising step selects the highest-probability valid token at each action position and computes candidate confidence as the geometric mean of their decoding probabilities. The entire action is committed when its confidence reaches the threshold $\tau_c$, or immediately before image generation begins at $\lfloor\alpha D\rfloor$, where the warm-up ratio $\alpha$ defaults to $1/4$. Image tokens then follow a cosine masking schedule with confidence-based re-masking~\citep{chang2022maskgit}.

\noindent \textbf{Reciprocal~checks.}~After the check positions $d \in \{\lfloor D/2\rfloor, \lfloor 3D/4\rfloor\}$, RecD applies an \textcolor{recreverse}{\texttt{action check}} followed by an \textcolor{recforward}{\texttt{imagine check}} performed by the same model $f_\theta$, as detailed in Fig.~\ref{fig:training_inference}b and illustrated in Fig.~\ref{fig:check}.

\begin{itemize}
\item The \textcolor{recreverse}{\texttt{action check}} masks all action tokens and the NEXT STEP and PREDICTION slots, leaving the remaining context unchanged. We compute the support score $q_a$ as the geometric mean of the original action-token probabilities under this context. If $q_a$ falls below the action-check threshold $\tau_a$, these regions remain masked~(\eg Fig.~\ref{fig:check}a); otherwise, their pre-check contents are restored.

\item The \textcolor{recforward}{\texttt{imagine check}} masks all revealed next-frame tokens and scores their original values using the context left by the \textcolor{recreverse}{\texttt{action check}}. Each token is restored if its probability $q_i$ reaches the imagine-check threshold $\tau_o$; otherwise, it remains masked~(\eg Fig.~\ref{fig:check}b). This check runs whether the action is retained or rejected.
\end{itemize}

Re-masked positions re-enter subsequent denoising. A rejected action is regenerated and committed as a whole in the next denoising step without applying $\tau_c$; reasoning and image tokens follow their respective schedules. The completed observation updates the rolling context, which retains at most $K$ frames.

\section{Experiments}
\label{sec:experiments}

We describe the experimental settings, compare \ours{} with existing methods, and analyze our design choices through ablation studies, paired analyses, and qualitative examples.

\subsection{Experimental Settings}
\label{sec:experimental_settings}

\begin{table*}[t]
    \centering
    \caption{\textbf{Comparison on instruction-guided world modeling} across gaming and navigation. Gaming pools the global-grid and egocentric-grid settings. Venue refers to the original method. SR and SPL are reported as percentages. DS denotes DreamSim. Best values are in bold.}
    \label{tab:main_langwm}
    \small
    \setlength{\tabcolsep}{2pt}
    \renewcommand\arraystretch{1.4}
    \resizebox{\linewidth}{!}{
    \begin{tabular}{l|cIIc*{7}{c}Ic*{7}{c}}
        \hline\thickhline
        \rowcolor{gray!20}
        &
        & \multicolumn{8}{cI}{\textbf{Gaming}}
        & \multicolumn{8}{c}{\textbf{Navigation}} \\
        \rowcolor{gray!20}& &
        &
        & \multicolumn{2}{c}{Rollout}
        & \multicolumn{2}{c}{Final frame}
        &
        &
        &
        &
        & \multicolumn{2}{c}{Rollout}
        & \multicolumn{2}{c}{Final frame}
        &
        &   \\
        \cline{5-8}\cline{13-16}
        \rowcolor{gray!20}\multirow{-3}{*}{\textbf{Method}}&\multirow{-3}{*}{\textbf{Venue}} &\multirow{-2}{*}{SR$\uparrow$} & \multirow{-2}{*}{SPL$\uparrow$}& FID$\downarrow$ & DS$\downarrow$
        & FID$\downarrow$ & DS$\downarrow$ &\multirow{-2}{*}{Cons.$\uparrow$} &
        \multirow{-2}{*}{RG$\uparrow$}&\multirow{-2}{*}{SR$\uparrow$} & \multirow{-2}{*}{SPL$\uparrow$}& FID$\downarrow$ & DS$\downarrow$
        & FID$\downarrow$ & DS$\downarrow$ &\multirow{-2}{*}{Cons.$\uparrow$} &
        \multirow{-2}{*}{RG$\uparrow$} \\
        \hline\hline

Anole~\citep{chern2024anole}
& arXiv'24
& 2.4 & 1.4 & 369.4 & 0.67 & 385.8 & 0.72 & 0.01 & 0.03
& 2.3 & 1.6 & 309.5 & 0.76 & 357.4 & 0.77 & 0.03 & 0.04 \\
\rowcolor{gray!10}
\emph{\hspace{1.5em} + fine-tuning}
&
& 41.3 & 30.9 & 97.6 & 0.23 & 146.0 & 0.30 & 0.53 & 0.57
& 28.1 & 21.5 & 193.5 & 0.38 & 285.6 & 0.47 & 0.48 & 0.52 \\

BAGEL~\citep{deng2025emerging}
& arXiv'25
& 5.6 & 3.1 & 142.2 & 0.22 & 183.9 & 0.30 & 0.06 & 0.17
& 4.4 & 2.7 & 200.9 & 0.46 & 308.0 & 0.58 & 0.17 & 0.14 \\
\rowcolor{gray!10}
\emph{\hspace{1.5em} + fine-tuning}
&
& 59.3 & 51.6 & 74.2 & 0.14 & 98.0 & 0.18 & 0.69 & 0.66
& 34.6 & 29.7 & 161.4 & 0.28 & 239.4 & 0.35 & 0.66 & 0.65 \\

MMaDA~\citep{yangmmada}
& NeurIPS'25
& 2.7 & 1.6 & 232.7 & 0.41 & 273.9 & 0.48 & 0.05 & 0.11
& 4.3 & 2.6 & 278.4 & 0.52 & 336.0 & 0.64 & 0.04 & 0.07 \\
\rowcolor{gray!10}
\emph{\hspace{1.5em} + fine-tuning}
&
& 54.8 & 46.1 & 78.9 & 0.16 & 112.3 & 0.22 & 0.63 & 0.68
& 35.6 & 28.9 & 168.6 & 0.31 & 258.9 & 0.39 & 0.59 & 0.66 \\

Lumina-DiMOO~\citep{xin2025lumina}
& arXiv'25
& 6.4 & 3.9 & 102.8 & 0.18 & 154.0 & 0.25 & 0.12 & 0.19
& 5.6 & 4.7 & 201.5 & 0.44 & 306.1 & 0.58 & 0.08 & 0.09 \\
\rowcolor{gray!10}
\emph{\hspace{1.5em} + fine-tuning}
&
& 76.8 & 69.5 & 54.7 & 0.09 & 70.2 & 0.13 & 0.78 & 0.79
& 39.2 & 33.7 & 139.2 & 0.23 & 226.8 & 0.30 & 0.67 & 0.69 \\

MMaDA-Parallel-M~\citep{tian2025mmada}
& ICLR'26
& 3.7 & 2.4 & 168.6 & 0.36 & 191.5 & 0.44 & 0.03 & 0.12
& 4.1 & 2.7 & 243.8 & 0.52 & 322.6 & 0.61 & 0.08 & 0.11 \\
\rowcolor{gray!10}
\emph{\hspace{1.5em} + fine-tuning}
&
& 60.4 & 51.9 & 71.5 & 0.13 & 108.5 & 0.19 & 0.67 & 0.74
& 34.8 & 27.5 & 155.8 & 0.27 & 241.0 & 0.36 & 0.64 & 0.70 \\

MMaDA-Parallel-A~\citep{tian2025mmada}
& ICLR'26
& 7.9 & 5.0 & 122.3 & 0.29 & 153.7 & 0.38 & 0.16 & 0.23
& 6.8 & 5.2 & 186.3 & 0.41 & 264.8 & 0.46 & 0.11 & 0.13 \\
\rowcolor{gray!10}
\emph{\hspace{1.5em} + fine-tuning}
&
& 78.6 & 70.4 & 52.3 & 0.09 & 73.8 & 0.12 & 0.79 & 0.81
& 38.2 & 33.1 & 132.6 & 0.21 & 218.4 & 0.28 & 0.69 & 0.71 \\

MVoT~\citep{li2025imagine}
& ICML'25
& 52.6 & 42.1 & 85.0 & 0.19 & 119.7 & 0.25 & 0.61 & 0.65
& 31.4 & 26.5 & 177.2 & 0.33 & 269.4 & 0.41 & 0.56 & 0.64 \\

\hline

NWM~\citep{bar2025navigation}
& CVPR'25
& 56.4 & 50.2 & 89.6 & 0.20 & 124.5 & 0.25 & 0.58 & --
& 39.5 & 33.8 & 126.9 & 0.24 & 216.4 & 0.31 & 0.69 & -- \\

NWM (lang)~\citep{bar2025navigation}
& CVPR'25
& 58.0 & 52.1 & 86.2 & 0.21 & 118.4 & 0.27 & 0.60 & --
& 41.2 & 35.4 & 129.0 & 0.23 & 213.8 & 0.30 & 0.72 & -- \\

LCVN-Uni~\citep{dong2026lcvn}
& NeurIPS'26
& 71.2 & 63.6 & 58.3 & 0.11 & 79.2 & 0.15 & 0.75 & --
& 51.4 & 42.9 & 122.4 & 0.20 & 210.5 & 0.27 & 0.76 & -- \\

UniWM~\citep{dong2026towards}
& ECCV'26
& 74.5 & 67.4 & 56.2 & 0.10 & 75.6 & 0.14 & 0.77 & --
& 53.8 & 46.0 & 119.8 & 0.19 & 208.2 & 0.26 & 0.79 & -- \\

\hline

\rowcolor{recjointbg!80}\textbf{\ours{}} w/o checks
& --
& 85.2 & 81.5 & 36.5 & 0.05 & 51.7 & 0.06 & 0.88 & 0.91
& 56.6 & 52.4 & 116.5 & 0.17 & 206.4 & 0.25 & 0.81 & 0.86 \\

\rowcolor{recjointbg!80}\textbf{\ours{}}
& --
& \textbf{94.9} & \textbf{92.8}
& \textbf{21.4} & \textbf{0.03}
& \textbf{32.1} & \textbf{0.03}
& \textbf{0.92} & \textbf{0.95}
& \textbf{63.2} & \textbf{58.5}
& \textbf{108.8} & \textbf{0.16}
& \textbf{194.9} & \textbf{0.23}
& \textbf{0.87} & \textbf{0.93}
\\
        \hline
    \end{tabular}}
    \vspace{-0.1cm}
\end{table*}

\noindent \textbf{Datasets.}~We evaluate on three scenarios: global-grid gaming using Maze~\citep{ivanitskiy2023configurable}, MiniBehavior~\citep{jin2023mini}, and FrozenLake~\citep{brockman2016openai,wu2024vsp}; egocentric-grid gaming using MiniBehavior with local, agent-centered observations; and egocentric navigation using four real-world robot datasets: Go Stanford~\citep{hirose2018gonet}, ReCon~\citep{shah2021rapid}, SCAND~\citep{karnan2022socially}, and HuRoN~\citep{hirose2023sacson}. Gaming is split by map configuration, with the global and egocentric views of each MiniBehavior configuration assigned to the same split; navigation is split by source trajectory. The training/test splits contain 13{,}684/3{,}421 episodes for global-grid gaming, 7{,}857/1{,}964 for egocentric-grid gaming, and 14{,}098/3{,}524 for egocentric navigation.~We annotate each episode with a language instruction and each step with structured reasoning text. All methods share same splits and are evaluated on test sets.

\noindent \textbf{Evaluation metrics.}~\textbf{(1) Planning:} we report success rate (SR) and success weighted by path length (SPL)~\citep{anderson2018evaluation}. Gaming success is evaluated by executing generated actions in the simulator; SPL uses optimal task-completion lengths, including object interactions. Navigation success requires endpoint error below the reference trajectory's average step length~\citep{dong2026lcvn}, with SPL weighted by reference-path length. \textbf{(2) Imagination:} we report rollout and final-frame FID~\citep{heusel2017fid} and DreamSim~\citep{fu2023dreamsim} against reference observations. Since successful alternative routes may differ visually from the references, we interpret these metrics alongside planning and consistency. \textbf{(3) Consistency (Cons.) and (4) Reasoning Grounding (RG):} both are evaluated using GPT-5.6 Sol~\citep{openai2026gpt56} in an LLM-as-a-judge framework~\citep{tian2025mmada,zheng2023llmjudge,fu2025learning}. Cons. measures action--observation agreement, scoring inconsistent or visually uncertain steps as zero. RG averages binary evidence-support scores over the four reasoning slots. Both metrics are averaged over steps within each episode and then over episodes. All methods use the same judge prompts and inference settings, with method identities withheld. We validate the judge against three blinded human annotators on stratified subsets of test steps for both Cons. and RG.

\noindent \textbf{Baselines.}~We compare against multimodal and world models.~Multimodal baselines include Anole-7B~\citep{chern2024anole}, BAGEL-7B-MoT (w/ Think)~\citep{deng2025emerging}, MMaDA-8B~\citep{yangmmada}, Lumina-DiMOO-8B~\citep{xin2025lumina}, and two MMaDA-Parallel variants~\citep{tian2025mmada}: MMaDA-Parallel-M based on MMaDA-8B and MMaDA-Parallel-A based on Lumina-DiMOO-8B. We evaluate each model in zero-shot and fine-tuned (FT) settings. All FT variants and MVoT~\citep{li2025imagine} are trained on our task data using our answer format to jointly predict reasoning, actions, and observations.~World-model baselines include UniWM~\citep{dong2026towards}, LCVN-Uni~\citep{dong2026lcvn}, and original and language-conditioned variants of NWM~\citep{bar2025navigation}, both paired with LCVN-AC following~\citep{dong2026lcvn}. All world models are adapted to our task and action spaces and trained on same task data. \ours{} w/o checks disables reciprocal checks with other settings unchanged.

\noindent \textbf{Implementation details.}~We initialize \ours{} from Lumina-DiMOO-8B~\citep{xin2025lumina} with fixed pretrained tokenizers. Stage~1 mixes forward and reverse modes at a 1:1 ratio. We set $\beta_{\min}=0.3$ for asymmetric masking in both stages and retain up to $K=4$ context frames, using $H=1$ initial observation for gaming and $H=4$ for navigation. Inference uses $D=128$ denoising steps, a warm-up ratio $\alpha=1/4$, and checks after denoising steps 64 and 96. The commitment and check thresholds are $\tau_c=0.4$, $\tau_a=0.5$, and $\tau_o=0.3$. We train on four NVIDIA A100 GPUs (80\,GB each). Additional dataset, implementation, and experimental details are provided in the appendix.

\begin{table}[t]
    \centering
    \caption{\textbf{Comparison on both val seen and unseen splits of LCVN benchmark~\citep{dong2026lcvn}.} NWM, NWM (lang), and LCVN-WM
    are paired with LCVN-AC for navigation. Best values are in bold.}
    \label{tab:lcvn}
    \footnotesize
    \setlength{\tabcolsep}{6pt}
    \renewcommand\arraystretch{1.2}
    \resizebox{1\columnwidth}{!}{
    \begin{tabular}{lII*{3}{p{1.15cm}<{\centering}}I*{3}{p{1.15cm}<{\centering}}}
        \hline\thickhline
        \rowcolor{gray!20}
        & \multicolumn{3}{cI}{\textbf{Val Seen}}
        & \multicolumn{3}{c}{\textbf{Val Unseen}} \\
        \cline{2-4}\cline{5-7}
        \rowcolor{gray!20}
        \multirow{-2}{*}{\textbf{Method}}
        & ATE$\downarrow$
        & RPE$\downarrow$
        & SR$\uparrow$
        & ATE$\downarrow$
        & RPE$\downarrow$
        & SR$\uparrow$ \\
        \hline\hline

        NWM~\citep{bar2025navigation}
        & 0.75 & 0.24 & 0.28
        & 2.17 & 0.73 & 0.12 \\

        NWM (lang)~\citep{bar2025navigation,dong2026lcvn}
        & 0.72 & 0.26 & 0.29
        & 2.21 & 0.70 & 0.13 \\

        LCVN-WM~\citep{dong2026lcvn}
        & 0.34 & 0.12 & 0.43
        & 1.51 & 0.56 & 0.19 \\

        LCVN-Uni~\citep{dong2026lcvn}
        & 0.36 & \textbf{0.11} & 0.42
        & 1.21 & 0.47 & 0.27 \\

        UniWM~\citep{dong2026towards}
        & 0.35 & \textbf{0.11} & 0.43
        & 1.14 & 0.43 & 0.29 \\ 

        \hline

        \rowcolor{recjointbg!80}\textbf{\ours{}} w/o checks
        & 0.34 & 0.15 & 0.44
        & 1.29 & 0.52 & 0.30 \\

        \rowcolor{recjointbg!80}\textbf{\ours{}}
        & \textbf{0.30} & 0.12 & \textbf{0.51}
        & \textbf{1.05} & \textbf{0.41}  & \textbf{0.34} \\

        \hline
    \end{tabular}}
    \vspace{-0.1cm}
\end{table}

\begin{figure*}[!t]
    \centering
    \includegraphics[width=\linewidth]{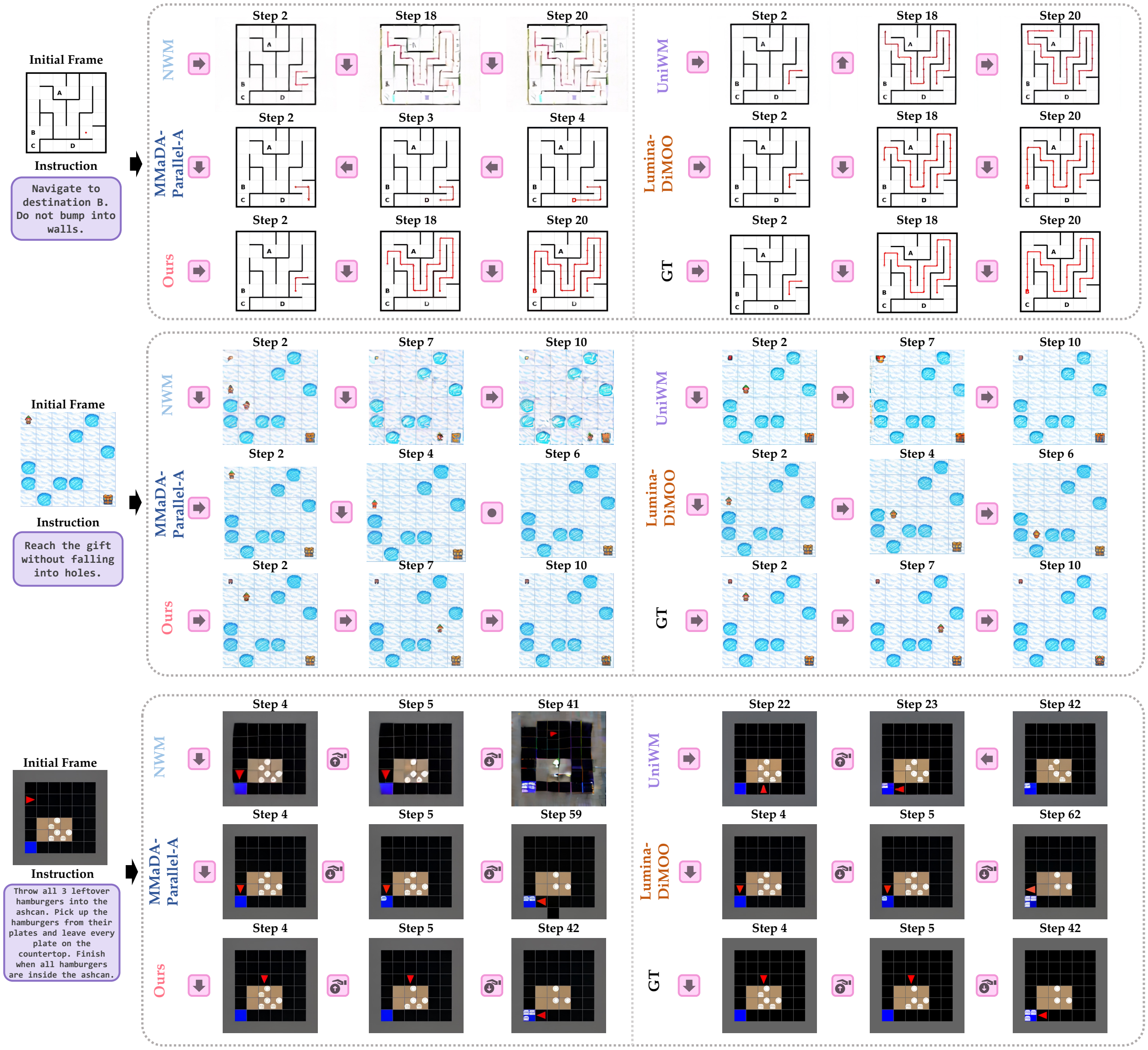}
    \caption{\textbf{Qualitative comparisons in global-grid gaming.}~We compare \ours{}, NWM, UniWM, MMaDA-Parallel-A (FT), and
    Lumina-DiMOO (FT) on Maze, FrozenLake, and MiniBehavior.
    Each task panel shows initial observation and instruction,
    followed by observations and corresponding actions at selected
    rollout steps, with ground truth for reference.}
    \label{fig:sota}
\end{figure*}

\subsection{Comparison with Baselines}
\label{sec:main_comparison}

\begin{figure}[!t]
    \centering
    \includegraphics[width=0.9\linewidth]{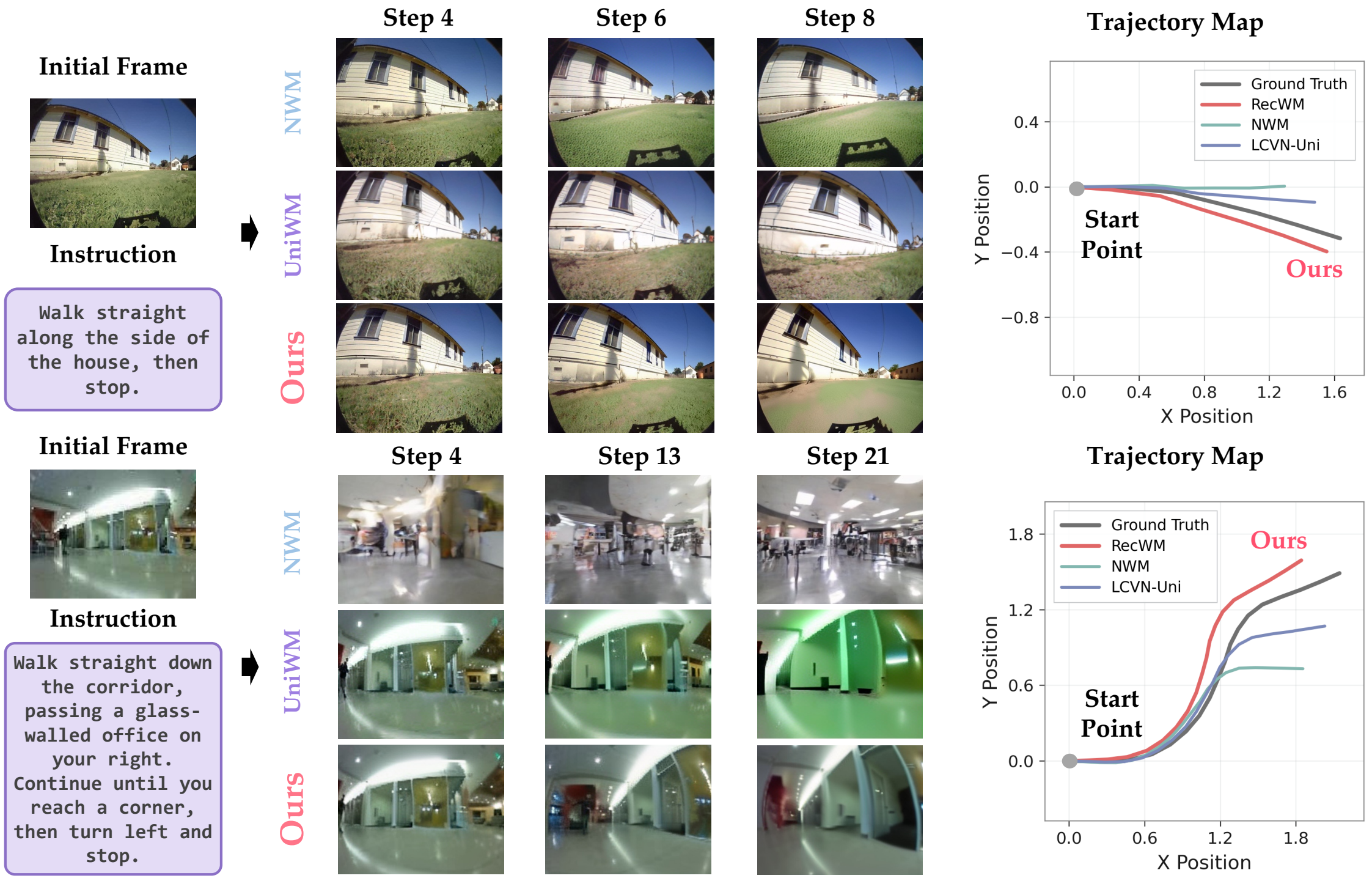}
    \caption{\textbf{Qualitative results} of our RecWM, NWM, and UniWM on instruction-guided world modeling in Navigation.}
    \label{fig:visual_res2}
\end{figure}

Tab.~\ref{tab:main_langwm} shows that \ours{} performs competitively across all reported metrics, including against Lumina-DiMOO (FT) and MMaDA-Parallel-A (FT), which share our backbone. In gaming, the small gap between SR and SPL for \ours{} indicates high average path efficiency among its successful episodes. The MiniBehavior example in Fig.~\ref{fig:sota} illustrates a difference in task progression: \ours{} picks up a hamburger at step 5, whereas UniWM walks around to the lower side of the countertop and performs its first pickup only at step 23. In navigation, the gains extend to both planning and visual prediction. Fig.~\ref{fig:visual_res2} shows that \ours{} better follows the instructed motion while preserving scene structure across changing viewpoints throughout the rollout.

We additionally evaluate \ours{} on LCVN~\citep{dong2026lcvn}, a public benchmark for language-conditioned visual navigation. For this experiment, we initialize a separate model directly from pretrained Lumina-DiMOO-8B and apply our two-stage training using only the LCVN training split, augmented with four-slot reasoning annotations derived from its observations, actions, and instructions. Tab.~\ref{tab:lcvn} reports competitive performance on both seen and unseen validation splits. Comparing the same checkpoint with and without reciprocal checks, SR improves from 0.44 to 0.51 on the seen split and from 0.30 to 0.34 on the unseen split, with lower ATE and RPE in both settings.

\subsection{Ablation and Analysis}
\label{sec:ablation}

Results in this section are averaged without weighting over the gaming and navigation settings in Tab.~\ref{tab:main_langwm}.

\noindent \textbf{Does two-stage training prepare \ours{} for Reciprocal Decoding?}~
Tab.~\ref{tab:abl_training} compares three training schemes: joint-only training, forward-mode training followed by joint training, and our full scheme, which mixes forward and reverse modes at a 1:1 ratio in Stage~1 before joint training in Stage~2. All variants start from the same pretrained initialization, use the same total number of optimization updates, and are evaluated with identical RecD settings. Forward-mode training raises SR from 67.8\% to 71.2\% and reduces final-frame DreamSim from 0.172 to 0.144 relative to joint-only training. The full scheme further raises SR to 79.1\% and consistency from 0.815 to 0.895. Thus, learning action-conditioned visual prediction provides a useful starting point, while including reverse action inference brings additional gains.

We further examine how the masking rules prepare the model for the inputs encountered during decoding. Fixing $\beta=1$ removes the masking asymmetry while retaining each mode's action-masking and supervision rules; in reverse mode, NEXT STEP and PREDICTION remain fully masked. This reduces SR from 79.1\% to 75.4\% and consistency from 0.895 to 0.858. Separately, leaving the complete next frame visible in every reverse-mode example, by setting $\rho_{\mathrm{img}}=0$, reduces SR to 77.1\% and consistency to 0.874. This result supports training reverse inference with incomplete next frames, matching the partial visual evidence available during action checks. 

\noindent \textbf{How do reciprocal checks improve predictions?}~
All decoding variants in Tab.~\ref{tab:abl_decoding} use the same trained checkpoint. Serial decoding generates reasoning and actions during steps 1--31, fixes them, and generates the next frame during steps 32--128. Joint decoding instead uses the base schedules of RecD, including the same image-generation warm-up, but disables both reciprocal checks. Moving from serial to joint decoding raises SR from 67.6\% to 70.9\% and reduces final-frame DreamSim from 0.162 to 0.155. Adding both checks further raises SR to 79.1\% and reduces DreamSim to 0.130, showing that joint generation alone does not account for the full improvement. The two checks contribute differently. Relative to joint decoding without checks, action-only checks raise SR by 4.3 percentage points, compared with 1.6 points for imagine-only checks. Conversely, imagine-only checks reduce final-frame DreamSim from 0.155 to 0.140, compared with 0.147 for action-only checks. Applying both achieves the best planning, visual prediction, and consistency scores. These results support complementary roles: action checks contribute more to selecting effective actions, while imagine checks contribute more to refining their visual outcomes.

\begin{table}[!t]
 \renewcommand\arraystretch{1.2}
    \centering
    \caption{\textbf{Training ablations.} All variants use matched total training updates and identical RecD settings. Partial frame denotes partially masked next-frame conditioning in reverse mode.}
    \label{tab:abl_training}
    \footnotesize
    \setlength{\tabcolsep}{2pt}
    \resizebox{\columnwidth}{!}{
    \begin{tabular}{*{5}{p{1.45cm}<{\centering}}IIp{1.4cm}<{\centering}p{1.4cm}<{\centering}p{1.9cm}<{\centering}p{1.7cm}<{\centering}}
        \hline\thickhline
        \rowcolor{gray!20}
        & 
        & 
        & Partial
        & Asym.
        & 
        & 
        & 
        &  \\
        \rowcolor{gray!20}\multirow{-2}{*}{Forward}& \multirow{-2}{*}{Reverse}& \multirow{-2}{*}{Joint}& Frame & Masking &\multirow{-2}{*}{SR$\uparrow$} &\multirow{-2}{*}{SPL$\uparrow$}& \multirow{-2}{*}{Final DS$\downarrow$} &\multirow{-2}{*}{Cons.$\uparrow$} \\
        \hline\hline
        \textcolor[rgb]{1,0,0}{\texttimes}
        & \textcolor[rgb]{1,0,0}{\texttimes}
        & \textcolor[rgb]{0,0,1}{\ding{51}}
        & --
        & \textcolor[rgb]{0,0,1}{\ding{51}}
        & 67.8 & 62.9 & 0.172 & 0.786 \\
        \textcolor[rgb]{0,0,1}{\ding{51}}
        & \textcolor[rgb]{1,0,0}{\texttimes}
        & \textcolor[rgb]{0,0,1}{\ding{51}}
        & --
        & \textcolor[rgb]{0,0,1}{\ding{51}}
        & 71.2 & 67.0 & 0.144 & 0.815 \\
        \textcolor[rgb]{0,0,1}{\ding{51}}
        & \textcolor[rgb]{0,0,1}{\ding{51}}
        & \textcolor[rgb]{0,0,1}{\ding{51}}
        & \textcolor[rgb]{0,0,1}{\ding{51}}
        & \textcolor[rgb]{1,0,0}{\texttimes}
        & 75.4 & 70.8 & 0.139 & 0.858 \\
        \textcolor[rgb]{0,0,1}{\ding{51}}
        & \textcolor[rgb]{0,0,1}{\ding{51}}
        & \textcolor[rgb]{0,0,1}{\ding{51}}
        & \textcolor[rgb]{1,0,0}{\texttimes}
        & \textcolor[rgb]{0,0,1}{\ding{51}}
        & 77.1 & 73.9 & 0.135 & 0.874 \\
        \rowcolor{recjointbg!80}\textcolor[rgb]{0,0,1}{\ding{51}}
        & \textcolor[rgb]{0,0,1}{\ding{51}}
        & \textcolor[rgb]{0,0,1}{\ding{51}}
        & \textcolor[rgb]{0,0,1}{\ding{51}}
        & \textcolor[rgb]{0,0,1}{\ding{51}}
        & \textbf{79.1} & \textbf{75.7}
        & \textbf{0.130} & \textbf{0.895} \\
        \hline
    \end{tabular}}
    \vspace{-0.2cm}
\end{table}

We next test whether these gains can be explained by additional computation or delayed action commitment. Full RecD uses 128 denoising steps and four additional model evaluations for the two pairs of checks. Allocating all 132 evaluations to denoising without checks changes SR only from 70.9\% to 71.0\%, leaving final-frame DreamSim at 0.155. The Late-64 and Late-96 variants instead keep the action and the NEXT STEP and PREDICTION slots masked until step 64 or 96, while image generation begins at step 32. They therefore allow visual evidence to develop before the action is first committed, but do not subsequently check or revise that commitment. Their SR values of 71.7\% and 72.9\% remain below the 75.2\% achieved by action-only checks. Under these schedules, delayed commitment does not recover the gains from action reassessment.

\begin{figure*}[!t]
    \setlength{\abovecaptionskip}{-3pt}
    \setlength{\belowcaptionskip}{0pt}
    \centering

    \begin{minipage}[t]{0.49\textwidth}
        \vspace{0pt}
        \centering
        \includegraphics[width=1\linewidth]{Figures/check_abl.png}
        \par
    \end{minipage}
    \hfill
    \begin{minipage}[t]{0.49\textwidth}
        \vspace{0pt}
        \centering
        \includegraphics[width=1\linewidth]{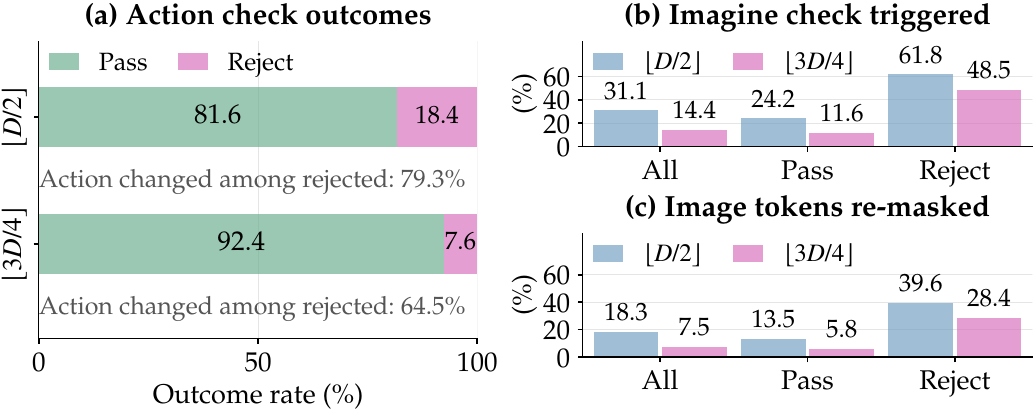}
        \par
    \end{minipage}

    \par\nointerlineskip
    \vspace{-14pt} 
    \begin{minipage}[b]{0.49\textwidth}
        \caption{\textbf{Qualitative results of reciprocal checks}
        on generated rollouts. Disabling checks leads to the wrong destination.}
        \label{fig:check_abl}
    \end{minipage}
    \hfill
    \begin{minipage}[b]{0.49\textwidth}
        \caption{\textbf{Check activation statistics.}
        (a) Action check outcomes per check position and fraction of
        rejected actions changed after regeneration.
        (b) Fraction of samples with re-masked image tokens and
        (c) fraction of re-masked tokens, by action check outcome.}
        \label{fig:check_activation}
    \end{minipage}
\end{figure*}

To assess whether the emerging next frame contributes to action revision, NoVis retains both checks but hides next-frame tokens during action checking and rejected-action regeneration. The imagine check and all thresholds remain unchanged. SR drops from 79.1\% to 73.1\%, and consistency from 0.895 to 0.860, bringing performance close to imagine-only checks. This supports the contribution of predicted visual evidence to the action-checking and regeneration process. Fig.~\ref{fig:check_activation} reports action rejection rates and the fraction of rejected actions that change after regeneration, and shows more frequent and extensive visual re-masking following action rejection. Fig.~\ref{fig:check_abl} illustrates how reciprocal checks affect subsequent task progression.

\noindent \textbf{Do triggered revisions improve action--observation consistency?}~
We copy the same intermediate decoding state into two branches immediately before a selected check position. The intervention branch applies the action check, the imagine check, or both in sequence; the control branch skips the selected checks. Both branches use the same base denoising settings and random seed, disable all subsequent checks, and finish the current environment step. At the second check position, their shared starting state already includes the default first checks and the intervening denoising steps.

Tab.~\ref{tab:check_effects} includes pairs for which the intervention triggers re-masking: action rejection for action-only checks, visual re-masking for imagine-only checks, or either event for the combined intervention. Fix is the percentage of these pairs that change from inconsistent to consistent, and Harm is the percentage that change in the opposite direction; both use all triggered pairs as their denominator. Each intervention defines its own triggered subset, so the rows do not compare intervention effects on a common set of samples. Fix exceeds Harm for every intervention at both positions. For the combined intervention, $\mathrm{Fix}-\mathrm{Harm}$ gives net consistency gains of 20.3 percentage points at $D/2$ and 15.1 points at $3D/4$ within the corresponding triggered subsets.

\begin{table}[!t]
    \renewcommand{\arraystretch}{1.1}
    \centering
    \caption{
    \textbf{Decoding ablations.}
    All variants share the same training checkpoint.
    Serial fixes reasoning and actions before image denoising.
    Late-$k$ delays action commitment and generation of NEXT STEP and
    PREDICTION until step $k$.
    NoVis masks next-frame tokens during action checking and rejected-action regeneration.
    }
    \label{tab:abl_decoding}
    \footnotesize
    \setlength{\tabcolsep}{3.5pt}
    \resizebox{\columnwidth}{!}{
    \begin{tabular}{p{1.6cm}p{1.9cm}<{\centering}|p{1.1cm}<{\centering}|p{1.3cm}<{\centering}p{1.3cm}<{\centering}IIp{1.4cm}<{\centering}p{1.4cm}<{\centering}p{1.9cm}<{\centering}p{1.7cm}<{\centering}}
        \hline\thickhline

        \rowcolor{gray!20}
        & & & \multicolumn{2}{cII}{Checks}
        & & & & \\

        \rowcolor{gray!20}
        \multirow{-2}{*}{Gen.}
        & \multirow{-2}{*}{Variant}
        & \multirow{-2}{*}{$D$}
        & Act.
        & Img.
        & \multirow{-2}{*}{SR$\uparrow$}
        & \multirow{-2}{*}{SPL$\uparrow$}
        & \multirow{-2}{*}{Final DS$\downarrow$}
        & \multirow{-2}{*}{Cons.$\uparrow$} \\
        \hline\hline

        Serial
        & --
        & 128
        & \textcolor[rgb]{1,0,0}{\texttimes}
        & \textcolor[rgb]{1,0,0}{\texttimes}
        & 67.6 & 63.4 & 0.162 & 0.818 \\

        Joint
        & --
        & 128
        & \textcolor[rgb]{1,0,0}{\texttimes}
        & \textcolor[rgb]{1,0,0}{\texttimes}
        & 70.9 & 67.0 & 0.155 & 0.845 \\

        Joint
        & --
        & 132
        & \textcolor[rgb]{1,0,0}{\texttimes}
        & \textcolor[rgb]{1,0,0}{\texttimes}
        & 71.0 & 67.2 & 0.155 & 0.847 \\

        Joint
        & Late-64
        & 128
        & \textcolor[rgb]{1,0,0}{\texttimes}
        & \textcolor[rgb]{1,0,0}{\texttimes}
        & 71.7 & 67.9 & 0.150 & 0.862 \\

        Joint
        & Late-96
        & 128
        & \textcolor[rgb]{1,0,0}{\texttimes}
        & \textcolor[rgb]{1,0,0}{\texttimes}
        & 72.9 & 69.5 & 0.148 & 0.870 \\

        \hline

        Joint
        & --
        & 128
        & \textcolor[rgb]{0,0,1}{\ding{51}}
        & \textcolor[rgb]{1,0,0}{\texttimes}
        & 75.2 & 71.2 & 0.147 & 0.874\\

        Joint
        & --
        & 128
        & \textcolor[rgb]{1,0,0}{\texttimes}
        & \textcolor[rgb]{0,0,1}{\ding{51}}
        & 72.5 & 68.8 & 0.140 & 0.857 \\

        Joint
        & NoVis
        & 128
        & \textcolor[rgb]{0,0,1}{\ding{51}}
        & \textcolor[rgb]{0,0,1}{\ding{51}}
        & 73.1 & 69.6 & 0.142 & 0.860 \\

        \rowcolor{recjointbg!80}
        Joint
        & Full
        & 128
        & \textcolor[rgb]{0,0,1}{\ding{51}}
        & \textcolor[rgb]{0,0,1}{\ding{51}}
        & 79.1 & 75.7 & 0.130 & 0.895 \\

        \hline
    \end{tabular}}
\end{table}

\begin{figure}[!t]
    \centering
    \includegraphics[width=0.8\linewidth]{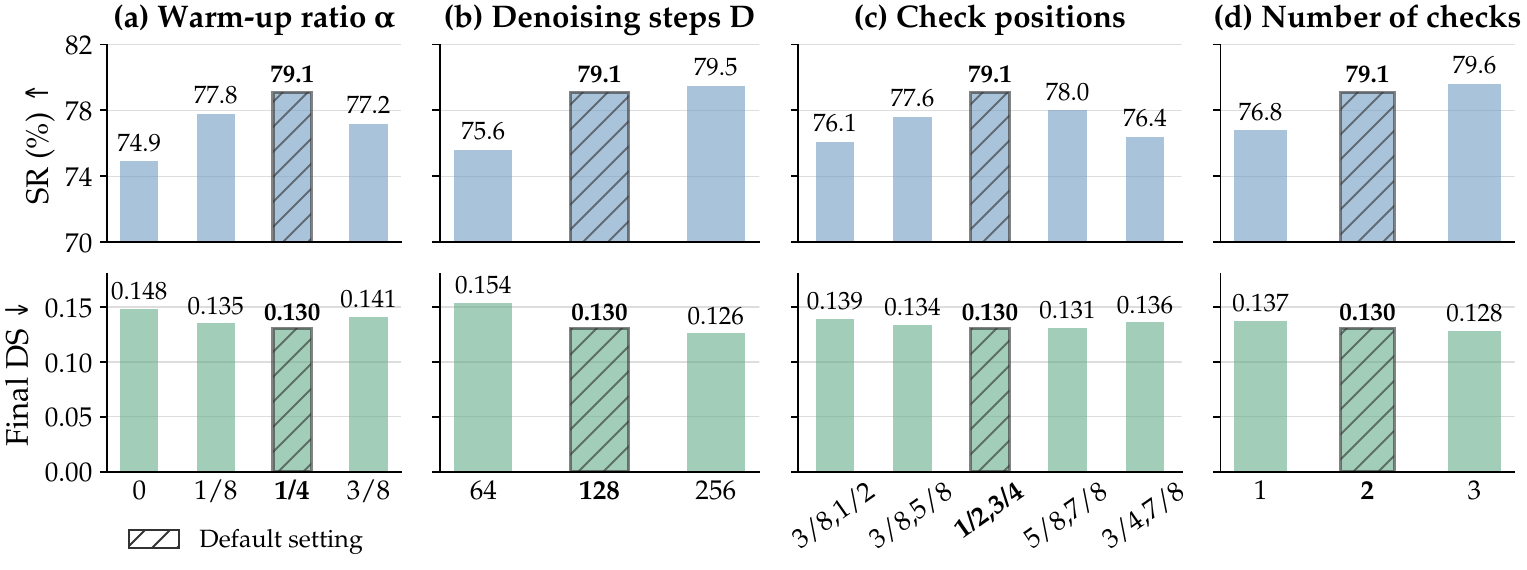}
    \caption{\textbf{Decoding ablations} of warm-up ratio, denoising steps, and check schedules.~Bold values for default settings.}
    \label{fig:abl_schedule}
    \vspace{-0.2cm}
\end{figure}

\noindent \textbf{How should Reciprocal Decoding be scheduled?}~
Fig.~\ref{fig:abl_schedule} examines the warm-up ratio, denoising budget, and check schedule. With $D=128$ and checks fixed at steps 64 and 96, we vary the warm-up ratio $\alpha$ over $\{0,1/8,1/4,3/8\}$. Starting image generation immediately gives an SR of 74.9\%, compared with 79.1\% when $\alpha=1/4$, while a longer warm-up of $3/8$ reduces SR to 77.2\%. A moderate warm-up allows reasoning and action context to develop while leaving enough steps for visual refinement.

Holding the denoising budget and thresholds fixed, we compare five pairs of check positions, applying an action check followed by an imagine check at each position. The default pair $\{D/2,3D/4\}$ performs best among the tested pairs. Earlier checks have less generated visual evidence available, while later checks leave fewer denoising steps for regenerating rejected predictions. 

With $\alpha=1/4$ and checks at $D/2$ and $3D/4$, increasing the budget from 64 to 128 denoising steps raises SR from 75.6\% to 79.1\%, whereas doubling it again to 256 yields 79.5\%. To vary the number of check positions, we keep $D=128$ and all thresholds fixed. For one position, we average the results of checking at $D/2$ alone and at $3D/4$ alone. Two positions use the default pair $\{D/2,3D/4\}$. For three positions, we add one of $\{3D/8,5D/8,7D/8\}$ to the default pair and average the results of the three resulting schedules. Each position includes both an action check and an imagine check, and schedules receive equal weight within each count. SR rises from 76.8\% with one position to 79.1\% with two and 79.6\% with three. The diminishing gains support our default of 128 denoising steps with checks at steps 64 and 96.

\begin{table}[!t]
 \renewcommand\arraystretch{0.95}
    \centering
    \caption{\textbf{Effects of triggered revisions on action--observation consistency.} Fix and Harm report the percentages of triggered pairs whose completed step changes from inconsistent to consistent, or vice versa, relative to the control branch that skips the selected checks. Both use all triggered pairs for the corresponding intervention and check position as the denominator. Each intervention defines its own triggered subset.}
    \label{tab:check_effects}
    \small
    \setlength{\tabcolsep}{3pt}
    \resizebox{\columnwidth}{!}{
    \begin{tabular}{p{3.0cm}|p{3.0cm}<{\centering}p{3.0cm}<{\centering}IIp{2.9cm}<{\centering}p{2.9cm}<{\centering}}
        \hline\thickhline
         \rowcolor{gray!20}Check position
        & Action Check
        &Imagine Check
        & Fix (\%)$\uparrow$
        & Harm (\%)$\downarrow$ \\
        \hline\hline
        \multirow{3}{*}{$\lfloor D/2\rfloor$}
        & \textcolor[rgb]{0,0,1}{\ding{51}}
        & \textcolor[rgb]{1,0,0}{\texttimes}
        & 15.6 & 3.2 \\
        & \textcolor[rgb]{1,0,0}{\texttimes}
        & \textcolor[rgb]{0,0,1}{\ding{51}}
        & 17.8 & 3.5 \\
        & \textcolor[rgb]{0,0,1}{\ding{51}}
        & \textcolor[rgb]{0,0,1}{\ding{51}}
        & 23.4 & 3.1 \\
        \hline
        \multirow{3}{*}{$\lfloor3D/4\rfloor$}
        & \textcolor[rgb]{0,0,1}{\ding{51}}
        & \textcolor[rgb]{1,0,0}{\texttimes}
        & 12.1 & 4.3 \\
        & \textcolor[rgb]{1,0,0}{\texttimes}
        & \textcolor[rgb]{0,0,1}{\ding{51}}
        & 14.9 & 4.6 \\
        & \textcolor[rgb]{0,0,1}{\ding{51}}
        & \textcolor[rgb]{0,0,1}{\ding{51}}
        & 19.2 & 4.1 \\
        \hline
    \end{tabular}}

\end{table}

\noindent \textbf{Do visual history and reasoning supervision help?}~
Tab.~\ref{tab:abl_context} first varies the inference-time context window $K$ at a fixed trained checkpoint. All variants start with the same available initial history, with $H=1$ for gaming and $H=4$ for navigation, but receive only the most recent $\min(K,H+t)$ observations at rollout step $t$. Increasing $K$ from 1 to 2 and 4 raises SR from 71.8\% to 76.5\% and 79.1\%, respectively. SPL, final-frame DreamSim, and consistency also improve. These gains show the benefit of temporal context, including previously generated observations as the rollout proceeds.

For reasoning supervision, we train three variants separately with matched total optimization updates: one omits the reasoning region, one retains only the four slot headers, and one uses full reasoning annotations. The action and next-frame regions retain their mode-specific masking and supervision rules. Headers alone raise SR from 70.6\% to 73.8\%, while full annotations further raise it to 79.1\%. Full annotations also reduce final-frame DreamSim from 0.136 to 0.130 and improve consistency from 0.869 to 0.895 relative to headers alone. Full reasoning annotations improve planning, visual prediction, and consistency beyond headers alone.

\begin{table}[!t]
    \renewcommand\arraystretch{0.9}
    \centering
    \caption{\textbf{Ablations of visual context size and reasoning supervision.} Results are averaged without weighting over gaming and navigation. SR and SPL are reported as percentages. Bold values indicate default settings.}
    \label{tab:abl_context}
    \small
    \setlength{\tabcolsep}{4.5pt}
    \resizebox{\columnwidth}{!}{
    \begin{tabular}{p{4.6cm}|p{2.8cm}<{\centering}IIp{1.65cm}<{\centering}p{1.65cm}<{\centering}p{2.0cm}<{\centering}p{1.8cm}<{\centering}}
        \hline\thickhline
        \rowcolor{gray!20}Setting & Value
        & SR$\uparrow$ & SPL$\uparrow$
        & Final DS$\downarrow$ & Cons.$\uparrow$ \\
        \hline\hline

        & 1
        & 71.8 & 66.4 & 0.141 & 0.862 \\
        & 2
        & 76.5 & 72.3 & 0.134 & 0.881 \\
        \rowcolor{recjointbg!80}\cellcolor{white}\multirow{-3}{*}{Context size $K$}& \textbf{4}
        & \textbf{79.1} & \textbf{75.7} & \textbf{0.130} & \textbf{0.895} \\
        \hline

        & None
        & 70.6 & 64.2 & 0.143 & 0.848 \\
        & Headers only
        & 73.8 & 69.5 & 0.136 & 0.869 \\
        \rowcolor{recjointbg!80}\cellcolor{white}\multirow{-3}{*}{Reasoning}& \textbf{Full}
        & \textbf{79.1} & \textbf{75.7} & \textbf{0.130} & \textbf{0.895} \\
        \hline
    \end{tabular}}
\end{table}

\subsection{Qualitative Results and Analysis}
\label{sec:qualitative}

Fig.~\ref{fig:sota} compares generated rollouts of \ours{}, NWM, UniWM, MMaDA-Parallel-A (FT), and Lumina-DiMOO (FT) on Maze, FrozenLake, and MiniBehavior in the global-grid gaming setting. We examine the selected steps to assess visual quality, task progression, and action--observation consistency.

In \emph{Maze}, NWM produces a forked trail and degraded walls and destination labels. UniWM preserves visual clarity but moves away from destination B, while MMaDA-Parallel-A (FT) reaches D instead. Both Lumina-DiMOO (FT) and \ours{} follow the reference route to B, illustrating that visual clarity and goal-directed planning are distinct aspects of performance.

In \emph{FrozenLake}, NWM duplicates the elf at step 2 and later distorts the scene, while UniWM and MMaDA-Parallel-A (FT) omit the elf in later selected frames. Lumina-DiMOO (FT) preserves the scene but depicts movements that disagree with the rightward actions at steps 4 and 6. \ours{} matches the reference positions at steps 2 and 7 and reaches the gift at step 10.

In \emph{MiniBehavior}, NWM progressively loses the room layout. UniWM performs its first pickup at step 23, compared with step 5 for \ours{}, and leaves two hamburgers on the countertop at step 42. Both MMaDA-Parallel-A (FT) and Lumina-DiMOO (FT) place a hamburger in the ashcan at step 5 without removing any of the three hamburgers from the countertop. \ours{} removes a hamburger after pickup and ends with three hamburgers in the ashcan and three empty plates on the countertop at step 42.

These examples show three kinds of baseline errors: degraded frames, wrong or roundabout routes, and observations that disagree with the chosen action. In the egocentric navigation examples (Fig.~\ref{fig:visual_res2}), \ours{} follows the instructions and produces trajectories closer to the references. Fig.~\ref{fig:check_abl} further illustrates how reciprocal checks affect subsequent rollout steps: an early action difference leads the agent to destination C with checks but to a wrong destination without them.

\noindent \textbf{Failure analysis.}~Rejecting a prediction does not guarantee correction. In Fig.~\ref{fig:check}a, regeneration repeats the rejected movement action after the first check and produces the correct pickup action only after the second.

\section{Conclusion}
\label{sec:conclusion}

We presented the Reciprocal World Model (\ours{}), which connects planning and imagination through mutual refinement during generation. Two-stage training develops forward and reverse prediction capabilities before joint generation, and Reciprocal Decoding uses these capabilities to reassess and selectively regenerate committed actions and visual tokens. Experiments across grid-based gaming and egocentric navigation demonstrate improvements in planning, visual prediction, action--observation consistency, and reasoning grounding. Controlled ablations support the complementary roles of directional training and reciprocal checks, while paired analyses show that triggered revisions improve consistency. These findings highlight the value of using a world model's predictive capabilities to revise its own decisions and imagined outcomes.

\section*{Acknowledgments}
This work was supported in part by faculty startup funding from the University of Washington and by the U.S. Department of Transportation through the Pacific Northwest Transportation Consortium (PacTrans), including PacTrans seed funding and the Small Research Projects program. We also gratefully acknowledge graduate student awards from the Vicky L. Carwein and William B. Andrews Endowment for Graduate Programs and the University of Washington Graduate School Top Scholar program. Zhi-Qi Cheng acknowledges faculty development support through the Undergraduate Research Faculty Fellows program, supported by the Henry Luce Foundation and the campus Office of Academic Affairs, and through Engineering Unleashed and the Kern Entrepreneurial Engineering Network (KEEN), generously supported by The Kern Family Foundation.

\appendix
\section*{Appendix}
\addcontentsline{toc}{section}{Appendix}  

\setcounter{table}{0}  
\setcounter{figure}{0}

\setcounter{equation}{0}
\renewcommand{\thetable}{A\arabic{table}}
\renewcommand{\thefigure}{A\arabic{figure}}

\renewcommand{\theequation}{A\arabic{equation}}

\noindent This supplementary material provides additional details on the datasets, method, and experiments to support the main paper. It is organized into three sections. First, Sec.~\ref{sec:app-scenarios} describes the evaluation scenarios and datasets, covering trajectory preprocessing, dataset splits, instruction and reasoning design, and the annotation and review process. Second, Sec.~\ref{sec:app-method} details \ours{}, including the sequence and action representation illustrated by representative prompt templates, the masking and supervision rules of two-stage training, and the implementation of Reciprocal Decoding with its action and imagine checks. Finally, Sec.~\ref{sec:app-experiments} presents more implementation details, baseline configurations, evaluation protocols, ablation configurations, and the statistics used in the Reciprocal Decoding analysis, together with additional ablation and qualitative results.

\section{Additional Scenario and Dataset Details}
\label{sec:app-scenarios}

This section expands on the scenarios introduced in Sec.~\ref{sec:tasks} and the datasets described in Sec.~\ref{sec:experimental_settings}. We detail trajectory preprocessing, followed by instruction and reasoning design for the three scenarios. Fig.~\ref{fig:scenarios} in the main text provides an overview of instructions and action types in three global-grid gaming environments.

\subsection{Trajectory Preprocessing}
\label{sec:app-scenario-details}

We construct trajectories from three gaming environments and four real-world navigation datasets. Gaming provides controlled layouts and object interactions, while navigation provides egocentric observations of real-world motion.

\noindent \textbf{Maze.}~Maze~\citep{ivanitskiy2023configurable} provides configurable grid layouts containing walls, traversable passages, an agent starting position, and labeled destinations. Its abstract visual representation makes spatial connectivity and route planning central to the task. We use the global observation setting, where the full maze is visible.

\noindent \textbf{MiniBehavior.}~MiniBehavior~\citep{jin2023mini} is a procedurally generated simulator for tasks involving navigation and object interactions. Objects have spatial relationships and states that change through actions such as picking up, placing, opening, and toggling. We use three MiniBehavior tasks: Installing a Printer, Throwing Away Leftovers, and Opening Packages, which we refer to as the printer, hamburger, and package tasks. Each is represented in both global and egocentric observation settings.

\noindent \textbf{FrozenLake.}~FrozenLake is a grid-based navigation environment implemented in Gym~\citep{brockman2016openai}, with visual task constructions studied in prior spatial-planning work~\citep{wu2024vsp}. Its layouts contain a starting position, a goal, safe frozen tiles, and holes. Compared with Maze, its rendered scenes contain richer visual patterns while introducing spatial hazards.

\noindent \textbf{Gaming preprocessing.}~We construct episodes from simulator trajectories and record the complete observation sequence, corresponding actions, and step-aligned agent coordinates. For egocentric MiniBehavior, we additionally retain agent orientations and render local observations according to the agent's pose. We use the MiniBehavior simulator to construct additional map configurations for three selected tasks. The resulting collection contains 4{,}500 Maze and 2{,}784 FrozenLake episodes. MiniBehavior contributes 3{,}647 printer episodes, 3{,}128 hamburger episodes, and 3{,}046 package episodes in each setting. Thus, gaming collection comprises 17{,}105 global-grid and 9{,}821 egocentric-grid episodes.

\noindent \textbf{Go Stanford.}~Go Stanford~\citep{hirose2018gonet} provides approximately 24 hours of indoor video from more than 25 environments, collected using a TurtleBot2 with fisheye cameras. The observations cover corridors, doorways, and narrow indoor passages, providing navigation trajectories through structurally repetitive scenes.

\noindent \textbf{ReCon.}~ReCon~\citep{shah2021rapid} contains over 5{,}000 self-supervised trajectories collected across nine environments over 18 months. Data collection uses time-correlated random walks and captures substantial variation in lighting and seasonal appearance. The trajectories provide diverse layouts and visual landmarks for open-world navigation.

\noindent \textbf{SCAND.}~SCAND~\citep{karnan2022socially} contains 8.7 hours of navigation demonstrations spanning approximately 40\,km. Human operators controlled a wheeled Clearpath Jackal and a legged Boston Dynamics Spot in indoor and outdoor environments.

\noindent \textbf{HuRoN.}~HuRoN~\citep{hirose2023sacson} contains approximately 75 hours of autonomous navigation across five environments, with over 4{,}000 human--robot interactions.

\noindent \textbf{Navigation preprocessing.}~We derive navigation actions from changes in recorded robot position and orientation. Following prior navigation pipelines~\citep{bar2025navigation,sridhar2024nomad}, we normalize per-frame planar displacement by each dataset's average step size (meters) and remove backward-motion segments. We discard trajectories shorter than three steps and use Qwen3.6-27B~\citep{qwen36_27b} to segment the remaining trajectories into semantically coherent sub-scenes. Each resulting episode retains aligned observations, robot poses, and actions specifying planar displacement and yaw change. After preprocessing and selection, our navigation collection contains 5{,}040 episodes from Go Stanford, 5{,}196 episodes from ReCon, 2{,}713 episodes from SCAND, and 4{,}673 episodes from HuRoN, totaling 17{,}622 episodes.

\noindent \textbf{Dataset splits.}~We use an approximately 80/20 train/test split within each scenario. For gaming, we split by map configuration, defined by the wall and object layout, the agent's starting position, and the goal. Each configuration appears in only one split. For MiniBehavior, both the global and egocentric views of a given map are assigned to that split. For navigation, all episodes derived from the same source trajectory are assigned to the same split. The resulting training/test splits contain 13{,}684/3{,}421 episodes for global-grid gaming, 7{,}857/1{,}964 for egocentric-grid gaming, and 14{,}098/3{,}524 for egocentric navigation. All methods use the same splits and are evaluated on the same 8{,}909 test episodes.

\subsection{Instruction and Reasoning Design}
\label{sec:app-data}
\label{sec:app-instruction-scenarios}

\begin{figure*}[!t]
\centering
\begin{promptbox}{Gaming -- FrozenLake}

\promptsection{Input (context prefix)}

\promptfield{System prompt}{[SYS]}

You are an agent crossing a frozen lake grid to reach the gift without falling into a hole. Each observation shows the whole grid from above: the elf (a small figure with a red/green hat) is you, light blue or white tiles are frozen and safe, dark blue circles are holes, and the orange wrapped present is the gift.

Cells are indexed as [row, col] with 0-based indexing. [0, 0] is the top-left cell; row increases downward and col increases rightward. Refer to positions with these coordinates.

Choose one action per step, and each move shifts the elf by exactly one tile. \acttok{go\_up} moves to [row-1, col]; \acttok{go\_down} moves to [row+1, col]; \acttok{go\_left} moves to [row, col-1]; \acttok{go\_right} moves to [row, col+1]. \acttok{goal\_reached} is the terminal action and is used only when the elf is on the gift tile.

Entering a hole fails the task, so before each move check the tile you would step onto. Plan a route over frozen tiles from your current tile to the gift, and take a detour whenever the direct direction leads into a hole or off the grid. Among safe routes, prefer the shortest.

After a move, the elf appears on the neighboring tile in the chosen direction, while the holes and the gift stay where they are.

Write the reasoning in four slots. In SCENE, describe visible objects and their layout. In PROGRESS, summarize what is complete and what remains. In NEXT STEP, state the action and its expected effect. In PREDICTION, describe the next observation. Preserve any supplied answer content and complete the rest consistently.

\medskip
\begin{minipage}[t]{0.35\linewidth}
    \vspace{0pt}
    \raggedright
    \promptfield{Task header and instruction}{[INSTR]}\par
    Complete the reasoning, action, and next observation for the current step.\par
    \textbf{Instruction:} Reach the gift without falling into holes.
\end{minipage}\hfill
\begin{minipage}[t]{0.63\linewidth}
    \vspace{0pt}
    \centering
    \promptfield{Context observations}{[BOI] $\cdots$ [EOI]}\par
    \smallskip
    \promptobs{$o_{t-3}$}{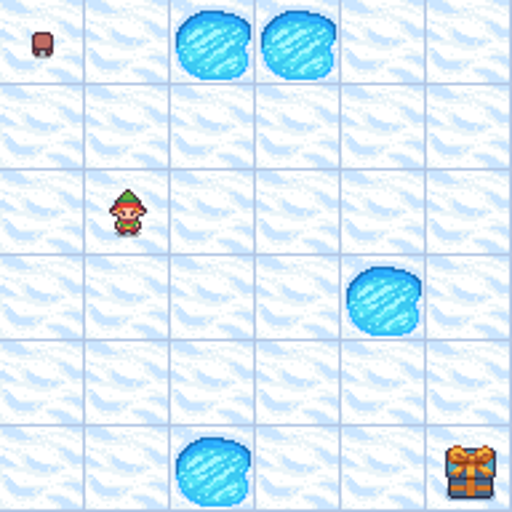}\hfill
    \promptobs{$o_{t-2}$}{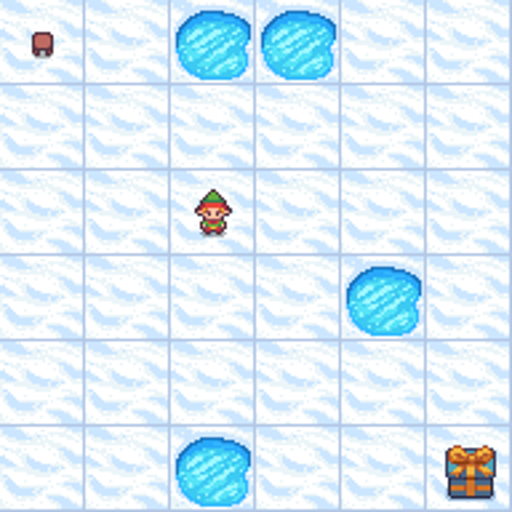}\hfill
    \promptobs{$o_{t-1}$}{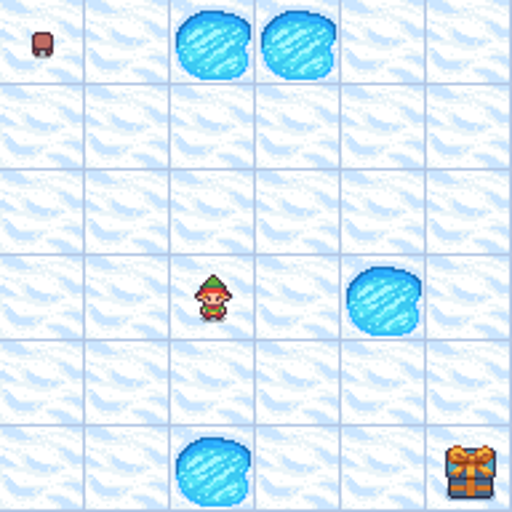}\hfill
    \promptobs{$o_t$}{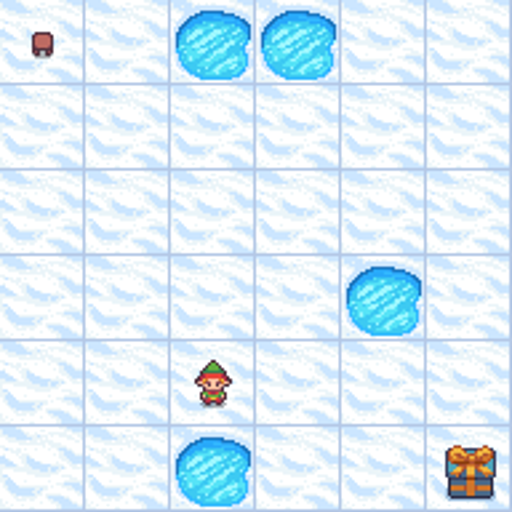}
\end{minipage}\par

\promptsection{Output (answer region)}

\begin{minipage}[t]{0.60\linewidth}
    \vspace{0pt}
    \raggedright
    \promptfield{Reasoning}{\textless think\textgreater\ $\cdots$ \textless/think\textgreater}\par
    \promptslot{SCENE}{Looking at my current view, I am at [4, 2] on a 6x6 frozen lake grid. I can see holes at [0, 2], [0, 3], [3, 4], [5, 2] and the gift at [5, 5].}\par
    \promptslot{PROGRESS}{I crossed from [2, 1] to [2, 2] and moved down two tiles to reach [4, 2]. The gift at [5, 5] is still ahead.}\par
    \promptslot{NEXT STEP}{The tile directly below me at [5, 2] is a hole. Go right from [4, 2] to the safe tile at [4, 3].}\par
    \promptslot{PREDICTION}{After Go right, the elf will appear at [4, 3], on safe frozen ice.}\par
    \medskip
    \promptfield{Action:}{}\acttok{go\_right}
\end{minipage}\hfill
\begin{minipage}[t]{0.37\linewidth}
    \vspace{0pt}
    \centering
    \promptfield{Next observation}{[BOI] img$_{t+1}$ [EOI]}\par
    \smallskip
    \promptobs{$o_{t+1}$}{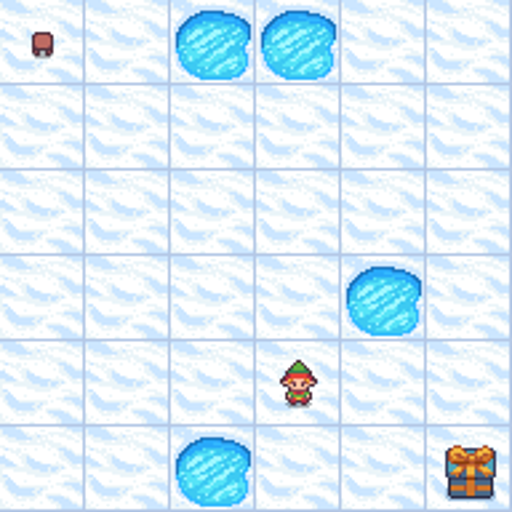}
\end{minipage}\par

\end{promptbox}
\caption{\textbf{Prompt template for global-grid gaming (FrozenLake).} The answer contains four reasoning slots, one action token, and the next frame. The training mode determines which answer tokens are masked and supervised.}
\label{fig:frozenlake_prompt}
\vspace{-0.3cm}
\end{figure*}

\begin{figure*}[!p]
\centering
\begin{promptbox}{Gaming -- MiniBehavior Hamburger (Global View)}

\promptsection{Input (context prefix)}

\promptfield{System prompt}{[SYS]}

You are an agent completing a hamburger disposal task in a MiniBehavior grid room: the leftover hamburgers must end up inside the ashcan, and any other requirement in the instruction must be respected. Each observation shows the entire map from above. The red triangle is you and points in the direction you are facing. Locate the hamburgers, the plates, the ashcan, and the furniture before acting.

Actions are relative to your facing direction. \acttok{turn\_left} and \acttok{turn\_right} rotate you by 90 degrees without changing your cell. \acttok{forward} moves you one cell in the facing direction and is possible only if that cell is free. Object actions apply to the cell directly in front of you, which can hold objects at three levels indexed 0, 1, and 2. \acttok{pickup\_0}, \acttok{pickup\_1}, and \acttok{pickup\_2} pick up the object at that level, and \acttok{drop\_0}, \acttok{drop\_1}, and \acttok{drop\_2} put the carried object down at that level. \acttok{drop\_in} puts the carried object inside the container in front of you and is the action that disposes of a hamburger in the ashcan. You can carry only one object at a time. \acttok{goal\_reached} is the terminal action and is used only when every hamburger is inside the ashcan.

Handle one hamburger at a time. Face it and pick it up, then move next to the ashcan, face it, and use \acttok{drop\_in}. Leave plates and other objects in place. Track how many hamburgers have been deposited and how many remain.

After a turn, the triangle keeps its cell and points in the new direction. After a forward move, it appears in the next cell. After a pickup or drop, the affected object disappears from or appears in the cell in front of you.

Write the reasoning in four slots. In SCENE, describe visible objects and their layout. In PROGRESS, summarize what is complete and what remains. In NEXT STEP, state the action and its expected effect. In PREDICTION, describe the next observation. Preserve any supplied answer content and complete the rest consistently.

\medskip
\begin{minipage}[t]{0.35\linewidth}
    \vspace{0pt}
    \raggedright
    \promptfield{Task header and instruction}{[INSTR]}\par
    Complete the reasoning, action, and next observation for the current step.\par
    \textbf{Instruction:} Throw all 3 leftover hamburgers into the ashcan. Pick up the hamburgers from their plates and leave every plate on the countertop. Finish when all hamburgers are inside the ashcan.
\end{minipage}\hfill
\begin{minipage}[t]{0.63\linewidth}
    \vspace{0pt}
    \centering
    \promptfield{Context observations}{[BOI] $\cdots$ [EOI]}\par
    \smallskip
    \promptobs{$o_{t-3}$}{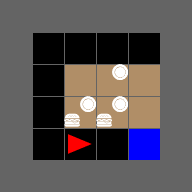}\hfill
    \promptobs{$o_{t-2}$}{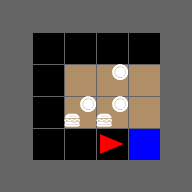}\hfill
    \promptobs{$o_{t-1}$}{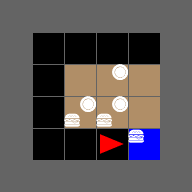}\hfill
    \promptobs{$o_t$}{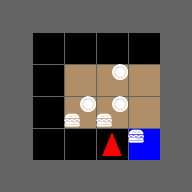}
\end{minipage}\par

\promptsection{Output (answer region)}

\begin{minipage}[t]{0.60\linewidth}
    \vspace{0pt}
    \raggedright
    \promptfield{Reasoning}{\textless think\textgreater\ $\cdots$ \textless/think\textgreater}\par
    \promptslot{SCENE}{The full 6 by 6 grid places the agent at row 4, column 3, facing up on the map. Visible objects include an ashcan at row 4, column 4, a hamburger at row 3, column 2, a hamburger at row 3, column 3, a plate at row 3, column 3, a plate at row 3, column 2 and a plate at row 2, column 3.}\par
    \promptslot{PROGRESS}{The agent has put one hamburger in the ashcan and turned toward the countertop. Two hamburgers remain on their plates.}\par
    \promptslot{NEXT STEP}{Pick up the hamburger resting on the plate directly in front of the agent, using its occupied object layer. Take the food rather than the plate, so the plate remains on the countertop and the hamburger can be carried to the ashcan.}\par
    \promptslot{PREDICTION}{The selected hamburger will leave its plate and become the object carried by the agent, while the plate stays on the countertop. The agent remains in the same cell and keeps the same heading, so the surrounding layout should not shift or rotate.}\par
    \medskip
    \promptfield{Action:}{}\acttok{pickup\_2}
\end{minipage}\hfill
\begin{minipage}[t]{0.37\linewidth}
    \vspace{0pt}
    \centering
    \promptfield{Next observation}{[BOI] img$_{t+1}$ [EOI]}\par
    \smallskip
    \promptobs{$o_{t+1}$}{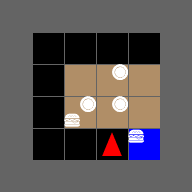}
\end{minipage}\par

\end{promptbox}
\caption{\textbf{Prompt template for global-grid gaming (MiniBehavior Hamburger).} The answer contains four reasoning slots, one action token, and the next frame. The training mode determines which answer tokens are masked and supervised.}
\label{fig:hamburger_prompt}
\end{figure*}

\begin{figure*}[!p]
\centering
\begin{promptbox}[egoGaming]{Gaming -- MiniBehavior Printer (Egocentric View)}
\promptsection{Input (context prefix)}

\promptfield{System prompt}{[SYS]}

You are an agent completing a printer installation task in a MiniBehavior grid room. Pick up the printer from the floor, place it on the table, and switch it on. The red triangle marks your position and points in the direction you are facing. The small white icon marks the printer, the brown block is the table, and the highlighted area shows your current field of view.

Choose one action token per step. Movement is relative to your heading. \acttok{turn\_left} and \acttok{turn\_right} rotate you in place by 90 degrees counterclockwise and clockwise, respectively. \acttok{forward} moves you one cell in the direction you are facing if that cell is free. After a turn, the triangle stays in the same cell, and the highlighted field of view rotates with it. The table and other stationary objects remain in their cells after every action.

Object actions apply to the cell directly in front of you, which can hold objects at three levels indexed 0, 1, and 2. \acttok{pickup\_0}, \acttok{pickup\_1}, and \acttok{pickup\_2} pick up the object at that level, and \acttok{drop\_0}, \acttok{drop\_1}, and \acttok{drop\_2} put the carried object down at that level. \acttok{toggle} flips the power state of the printer in the cell in front of you. \acttok{goal\_reached} is the terminal action and is used only when the printer is on the table and switched on.

Use the recent observations to track the printer through the task: whether it is still on the floor, being carried, or already on the table. The printer starts on the floor, which is level 0: face it and use \acttok{pickup\_0}. Its white icon then disappears from that cell. While carrying the printer, move next to the table, face one of its cells, and use \acttok{drop\_2} to place it on the tabletop. Once the printer is on the table in front of you, use \acttok{toggle} once to switch it on. Switching it on produces no visible change. After a successful toggle, output \acttok{goal\_reached}.

Write the reasoning in four slots. In SCENE, describe visible objects and their layout. In PROGRESS, summarize what is complete and what remains. In NEXT STEP, state the action and its expected effect. In PREDICTION, describe the next observation. Preserve any supplied answer content and complete the rest consistently.

\medskip
\begin{minipage}[t]{0.35\linewidth}
    \vspace{0pt}
    \raggedright
    \promptfield{Task header and instruction}{[INSTR]}\par
    Complete the reasoning, action, and next observation for the current step.\par
    \textbf{Instruction:} Pick up the printer from the floor, place it on the table, and turn it on.
\end{minipage}\hfill
\begin{minipage}[t]{0.63\linewidth}
    \vspace{0pt}
    \centering
    \promptfield{Context observations}{[BOI] $\cdots$ [EOI]}\par
    \smallskip
    \promptobs{$o_{t-3}$}{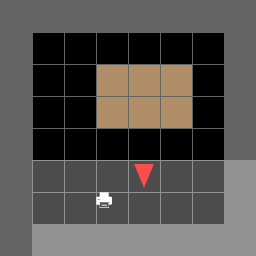}\hfill
    \promptobs{$o_{t-2}$}{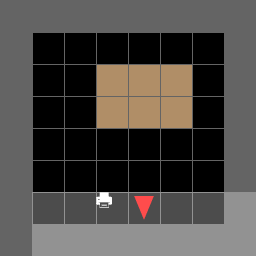}\hfill
    \promptobs{$o_{t-1}$}{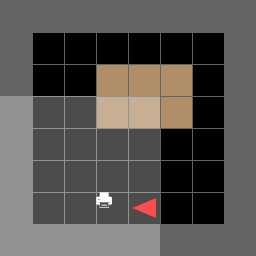}\hfill
    \promptobs{$o_t$}{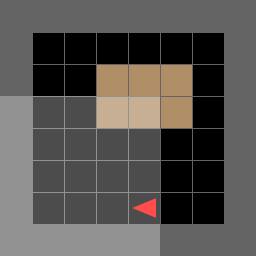}
\end{minipage}\par

\promptsection{Output (answer region)}

\begin{minipage}[t]{0.60\linewidth}
    \vspace{0pt}
    \raggedright
    \promptfield{Reasoning}{\textless think\textgreater\ $\cdots$ \textless/think\textgreater}\par
    \promptslot{SCENE}{The red triangle faces left near the bottom of the grid. The highlight extends to its left and covers only part of the table above the agent.}\par
    \promptslot{PROGRESS}{The agent has picked up the printer. The white printer icon has disappeared from its floor cell, and the printer still needs to be placed on the table.}\par
    \promptslot{NEXT STEP}{Turn right 90 degrees in place to face up the board toward the table.}\par
    \promptslot{PREDICTION}{The red triangle stays in the same cell and points up. The highlight shifts from the area to the agent's left to the area above it, bringing the whole table into the highlight. The table remains in the same cells.}\par
    \medskip
    \promptfield{Action:}{}\acttok{turn\_right}
\end{minipage}\hfill
\begin{minipage}[t]{0.37\linewidth}
    \vspace{0pt}
    \centering
    \promptfield{Next observation}{[BOI] img$_{t+1}$ [EOI]}\par
    \smallskip
    \promptobs{$o_{t+1}$}{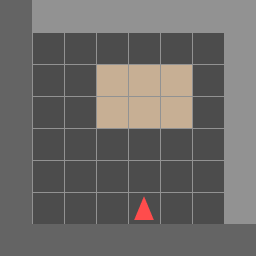}
\end{minipage}\par

\end{promptbox}
\caption{\textbf{Prompt template for egocentric-grid gaming (MiniBehavior Printer).} Only the highlighted light gray region, which marks the agent's current field of view, is visible to the agent. The training mode determines which answer tokens are masked and supervised.}
\label{fig:printer_prompt}
\vspace{-0.2cm}
\end{figure*}

\begin{figure*}[!p]
\centering
\begin{promptbox}[navigation]{Navigation}
\promptsection{Input (context prefix)}

\promptfield{System prompt}{[SYS]}

You are an agent navigating a real-world environment from an egocentric camera. The instruction describes the intended movement using the surroundings and visible landmarks. The observations are the most recent camera frames in chronological order, and the last one is the current view.

For each step, output either a movement action or the terminal response \texttt{goal reached}. A movement action specifies one local pose increment relative to the current camera heading: a forward or backward distance in meters, a left or right shift in meters, and a left or right turn in radians. Write all three components in full, for example: Move forward 0.115 meters; shift left 0.023 meters; turn right 0.150 radians. For a component that is zero at the stated precision, write Do not move forward or backward, do not shift sideways, or do not turn. When the observations indicate that the instructed goal has been reached, write only \texttt{goal reached} in the action region. A small or zero pose increment alone does not indicate arrival.

Use visible landmarks and changes across recent frames to assess progress. Report arrival only when the observations support it. Choose a motion increment consistent with the instruction and recent movement, staying on traversable ground and clear of people and obstacles.

Predict the next view from the action. A right turn shifts stationary landmarks to the left in the image, and a left turn shifts them to the right. Forward motion enlarges what lies ahead, and translation shifts nearby objects more than distant ones.

Write the reasoning in four slots. In SCENE, describe visible objects and their layout. In PROGRESS, summarize what is complete and what remains. In NEXT STEP, state the action and its expected effect. In PREDICTION, describe the next observation. Preserve any supplied answer content and complete the rest consistently.

\medskip
\begin{minipage}[t]{0.35\linewidth}
    \vspace{0pt}
    \raggedright
    \promptfield{Task header and instruction}{[INSTR]}\par
    Complete the reasoning, action, and next observation for the current step.\par
    \textbf{Instruction:} Walk straight past the glass-walled office on the left. Continue until you reach the lounge area with the yellow armchair and blue sofa, then turn right and stop near the blue sofa.\par
    
\end{minipage}\hfill
\begin{minipage}[t]{0.63\linewidth}
    \vspace{0pt}
    \centering
    \promptfield{Context observations}{[BOI] $\cdots$ [EOI]}\par
    \smallskip
    \promptobs{$o_{t-3}$}{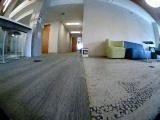}\hfill
    \promptobs{$o_{t-2}$}{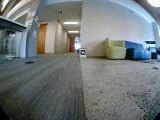}\hfill
    \promptobs{$o_{t-1}$}{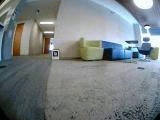}\hfill
    \promptobs{$o_t$}{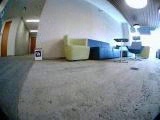}
\end{minipage}\par

\promptsection{Output (answer region)}

\begin{minipage}[t]{0.60\linewidth}
    \vspace{0pt}
    \raggedright
    \promptfield{Reasoning}{\textless think\textgreater\ $\cdots$ \textless/think\textgreater}\par
    \promptslot{SCENE}{The view faces a carpeted lounge. A yellow armchair and a blue sofa are ahead, with a corridor opening to the left.}\par
    \promptslot{PROGRESS}{Over the recent views I have moved forward and turned right toward the seats. The sofa is closer, and I still need to approach it.}\par
    \promptslot{NEXT STEP}{Move forward 0.191 meters, shift left 0.002 meters, and turn left 0.085 radians to move closer to the blue sofa while keeping the seating area ahead.}\par
    \promptslot{PREDICTION}{The seats should remain ahead and appear slightly closer. The corridor opening should still be visible on the left after the left turn.}\par
    \medskip
    \promptfield{Action:}{}Move forward 0.191 meters; shift left 0.002 meters; turn left 0.085 radians.
\end{minipage}\hfill
\begin{minipage}[t]{0.37\linewidth}
    \vspace{0pt}
    \centering
    \promptfield{Next observation}{[BOI] img$_{t+1}$ [EOI]}\par
    \smallskip
    \promptobs{$o_{t+1}$}{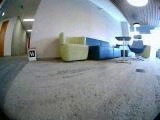}
\end{minipage}\par

\end{promptbox}
\caption{\textbf{Prompt template for egocentric navigation (HuRoN).} The answer contains four reasoning slots, a movement action in text, and the next frame. The training mode determines which answer tokens are masked and supervised.}
\label{fig:nav_prompt}
\end{figure*}

Each episode is paired with a language instruction, and each transition is annotated with structured reasoning. Prompt construction and context organization can influence how language models use the supplied information~\citep{liu2025grl,wang2025we}. Gaming instructions follow task-specific templates that specify the goal and relevant constraints. Each task has several paraphrased templates, and task parameters such as the destination label or the number of objects are filled in per episode. Navigation instructions use simple directional cues and descriptions of key landmarks along the route. Across both domains, instructions exclude trajectory metadata such as grid coordinates, metric displacements, and numerical turn angles. Fig.~\ref{fig:maze_prompt} in the main text provides a Maze example, while Figs.~\ref{fig:frozenlake_prompt}--\ref{fig:nav_prompt} provide additional examples of how episode-level instructions are paired with step-level reasoning, actions, and observations across the three scenarios.

\noindent \textbf{Global-grid gaming instructions.}~The full grid is visible in each environment. Fig.~\ref{fig:scenarios} illustrates the instructions and action types for Maze, FrozenLake, and MiniBehavior. In Maze, an instruction such as \textit{``Navigate to destination C. Do not bump into walls.''} identifies the target by its label, states the wall constraint, and leaves the route to the agent. In FrozenLake, \textit{``Reach the gift without falling into holes.''} specifies both the destination and the hazard to avoid. MiniBehavior instructions describe the required object interactions and their completion conditions. For example, \textit{``Pick up the printer from the floor, place it on the table, and turn it on.''} specifies the printer's final location and state. The instruction \textit{``Throw all 3 leftover hamburgers into the ashcan. Pick up the hamburgers from their plates and leave every plate on the countertop. Finish when all hamburgers are inside the ashcan.''} identifies the objects to move, their destination, and the objects to leave in place. In the package task, \textit{``Open both packages. Complete the task only when both packages are open.''} specifies the packages to open and the completion condition.

\noindent \textbf{Egocentric-grid gaming instructions.}~MiniBehavior uses the same task-specific instructions in the egocentric and global-view settings. In the egocentric setting, the agent's position and orientation determine its local field of view. Turning in place can bring an object into view or move it out of view even when the object remains in the same cell. As shown in Fig.~\ref{fig:printer_prompt}, for example, a right turn brings the entire table into the highlighted field of view.

\noindent \textbf{Egocentric navigation instructions.}~Navigation instructions describe a route through visible landmarks and simple directional cues. In the HuRoN example shown in Fig.~\ref{fig:nav_prompt}, the instruction directs the agent past a glass-walled office toward a lounge identified by a yellow armchair and a blue sofa, then asks it to turn right and stop near the sofa. The instruction specifies the route and stopping location, while the step-level movement actions provide the corresponding displacements and yaw changes.

\noindent \textbf{Structured reasoning.}~Each transition is annotated using the following four slots:
\begin{itemize}
    \item \textbf{SCENE} describes the relevant objects or landmarks visible in the current observation and their spatial relationships.
    \item \textbf{PROGRESS} summarizes what has been accomplished toward the instructed goal and what remains to be done.
    \item \textbf{NEXT STEP} states the intended action and its expected effect on the task.
    \item \textbf{PREDICTION} describes what the agent expects to see in the next observation after taking that action.
\end{itemize}

\noindent \textbf{Annotation and review.}~We use Qwen3.6-27B~\citep{qwen36_27b} to generate episode instructions and step-level reasoning annotations from recorded observations, actions, and task context. For each gaming source, we provide the task-specific requirements and representative examples of the prescribed instruction format. Navigation allows more variation in wording and landmark choice, so we provide a larger set of annotated examples illustrating how to express route segments using simple directional cues and key landmarks. We then manually check and refine a sampled subset of the generated annotations. Because gaming instructions follow relatively fixed templates, we sample approximately 10\% of the gaming episodes. Navigation annotations involve more scene elements and rely more on landmarks and other visual information, so we sample around 20\% of navigation episodes.

After this annotation process, we assess the quality of the final annotations on 500 gaming episodes and 500 navigation episodes. Three human annotators rate the instructions and associated reasoning on a six-point scale, judging whether each instruction agrees with the recorded trajectory, whether references to objects and landmarks are grounded in the observations, and whether the reasoning is consistent with the recorded actions and state transitions. The mean ratings are 5.9/6 for gaming and 5.5/6 for navigation.

\section{Additional Training and Decoding Details}
\label{sec:app-method}

\subsection{Sequence and Action Representation}
\label{sec:app-representation}

\noindent \textbf{Sequence layout.}~At environment step $t$, the model $f_\theta$ receives the language instruction $l$ and the available observation history. Following Eqs.~(2)--(3) of the main paper, each sequence consists of a context prefix followed by an answer region. The prefix contains the system prompt, the task header and instruction, and the context frames in chronological order. The answer is enclosed by \texttt{[BOA]} and \texttt{[EOA]} and contains three prediction regions in order: the reasoning text, the action $a_t$, and the visual tokens of the next observation $o_{t+1}$. The reasoning text is enclosed by \texttt{<think>} and \texttt{</think>} and contains the four slots SCENE, PROGRESS, NEXT STEP, and PREDICTION. Each image is delimited by \texttt{[BOI]} and \texttt{[EOI]}. The context prefix remains visible throughout training and decoding. 

\noindent \textbf{Token and action representations.}~Let $\mathbf{x}=(x_1,\ldots,x_L)\in\mathcal{V}^{L}$ denote a clean training sequence of length $L$. The shared vocabulary is $\mathcal{V}=\mathcal{V}_T\cup\mathcal{V}_A\cup\mathcal{V}_I$, where $\mathcal{V}_T$ contains BPE text tokens~\citep{xin2025lumina}, $\mathcal{V}_A$ contains dedicated gaming action tokens, and $\mathcal{V}_I$ contains visual codes from the pretrained aMUSEd-VQ tokenizer~\citep{patil2024amused}. A gaming action is represented by a single token from $\mathcal{V}_A$. A navigation movement action $a_t=(\mathbf{u}_t,\phi_t)$ specifies a planar displacement $\mathbf{u}_t\in\mathbb{R}^2$ in the robot's local frame and a yaw change $\phi_t$. The terminal action is the text \texttt{goal reached}. Both movement and terminal actions are encoded using the same BPE tokenizer as the reasoning. These tokens belong to $\mathcal{V}_T$, and their positions in the answer constitute the action region. Consequently, an operation on the entire action region covers one token for gaming and the full action text for navigation.

\noindent \textbf{Rolling visual context.}~Let $H$ denote the number of initial observations and $K$ the maximum number of context frames. At rollout step $t$, starting from $t=0$, the prefix contains the most recent $K_t=\min(K,H+t)$ frames from the initial history and previously generated observations. We use $H=1$ for gaming, $H=4$ for navigation, and $K=4$ by default. After each environment step, the generated observation is appended to the history, and up to $K$ of the most recent frames are retained.

\subsection{Two-Stage Training Configuration}
\label{sec:app-training}

Stage~1 mixes forward and reverse modes at a 1:1 ratio using the same annotated transitions. Stage~2 continues from the Stage~1 weights using joint mode. Each mode specifies which parts of the answer are masked and which masked positions contribute to the reconstruction objective.

\noindent \textbf{Masking ratios.}~For each clean sequence $\mathbf{x}$, we sample $r$ to determine the base masking ratio $\rho$ and draw an asymmetry factor $\beta$:
\begin{equation}
    r\sim\mathcal{U}(0,1),\quad
    \rho=\cos(\pi r/2),\quad
    \beta\sim\mathcal{U}(\beta_{\min},1).
\end{equation}
Each training mode uses $\rho$ and $\beta$ to set its regional masking ratios. The quantities $\rho_{\mathrm{txt}}$ and $\rho_{\mathrm{img}}$ denote the token masking ratios for reasoning and next-frame tokens, respectively. The quantity $\rho_{\mathrm{act}}$ denotes the probability of masking the entire action region, so all tokens of a navigation action share a single masking decision. Replacing the selected answer tokens with \texttt{[MASK]} yields the masked sequence $\tilde{\mathbf{x}}$.

\noindent \textbf{Training objective.}~Following Eq.~(4) of the main paper, let $m\in\{\mathrm{forward},\mathrm{reverse},\mathrm{joint}\}$ denote the training mode and let $p\in\{\mathrm{txt},\mathrm{act},\mathrm{img}\}$ index an answer region. For a given example and mode, $\mathcal{M}_p$ is the set of positions in region $p$ that are both masked and supervised. At position $i$, $x_i$ is the original token and $p_\theta(x_i\mid\tilde{\mathbf{x}})$ is the probability assigned to that token at that position given the corrupted sequence. The total training objective is $\mathcal{L}(\theta)=\mathcal{L}_{\mathrm{txt}} +\mathcal{L}_{\mathrm{act}}+\mathcal{L}_{\mathrm{img}}$. For each region $p\in\{\mathrm{txt},\mathrm{act},\mathrm{img}\}$, the loss is
\begin{equation}
    \mathcal{L}_p =
    -\mathbb{E}_{\mathbf{x},m,\tilde{\mathbf{x}}}
    \left[
        \frac{\sum_{i\in\mathcal{M}_p}
              \log p_\theta(x_i\mid\tilde{\mathbf{x}})}
             {\max(1,|\mathcal{M}_p|)}
    \right].
\end{equation}
For each example, the loss is averaged over the supervised masked positions within each region. A region contributes zero for that example when $\mathcal{M}_p=\varnothing$. The expectation is taken over training examples, the mode used in the current stage, and the sampled corruption. Normalizing within each region prevents a longer region from receiving greater weight solely because it contains more tokens.

\noindent \textbf{Stage~1, forward mode.}~The complete action remains visible, and the model reconstructs masked reasoning and next-frame tokens conditioned on that action. We set
\begin{equation}
    \rho_{\mathrm{act}}=0,\qquad
    \rho_{\mathrm{txt}}=\beta\rho,\qquad
    \rho_{\mathrm{img}}=\rho.
\end{equation}
The lower reasoning masking ratio preserves reasoning context for predicting the visual outcome. Masked reasoning and next-frame positions are supervised, and $\mathcal{M}_{\mathrm{act}}=\varnothing$.

\noindent \textbf{Stage~1, reverse mode.}~The model reconstructs the action and masked reasoning from the observation history and the visible portion of the next frame. We always mask the entire action together with the NEXT STEP and PREDICTION slots, since these slots can disclose the action or its expected effect. We set
\begin{equation}
    \rho_{\mathrm{act}}=1,\qquad
    \rho_{\mathrm{txt}}=\rho,\qquad
    \rho_{\mathrm{img}}=\beta\rho.
\end{equation}
Here $\rho_{\mathrm{txt}}$ applies to the remaining reasoning tokens in SCENE and PROGRESS. The fully masked NEXT STEP and PREDICTION slots also contribute to reasoning supervision. The reduced next-frame masking ratio preserves visual evidence for action inference while exposing the model to partially revealed outcomes. Masked reasoning and action positions are supervised. The next frame serves as conditioning, so $\mathcal{M}_{\mathrm{img}}=\varnothing$ even when visual tokens are masked.

\noindent \textbf{Stage~2, joint mode.}~Reasoning and next-frame tokens follow the forward-mode asymmetric masking rule, and the action is masked as a whole with probability $\rho$:
\begin{equation}
    \rho_{\mathrm{act}}=\rho,\qquad
    \rho_{\mathrm{txt}}=\beta\rho,\qquad
    \rho_{\mathrm{img}}=\rho.
\end{equation}
The reasoning ratio applies across all four slots. When the action is masked, visible reasoning and next-frame tokens provide context for inferring it. When the action remains visible, it provides context for reconstructing reasoning and the next frame. Masked prediction positions in all three regions are supervised in joint mode. On examples with a visible action, $\mathcal{M}_{\mathrm{act}}=\varnothing$.

\subsection{Reciprocal Decoding Implementation}
\label{sec:app-denoising}
\label{sec:app-recd}

RecD generates one answer per environment step $t$. Within that step, $d=1,\ldots,D$ indexes denoising steps. A revealed token is a prediction currently visible to the model. Committing an action makes its entire span visible, and a subsequent action check can reassess that commitment. The model $f_\theta$ performs denoising and both reciprocal checks.

\noindent \textbf{Initialization and reveal schedules.}~At the start of each environment step, the reasoning, action, and next-frame prediction regions are initialized with \texttt{[MASK]}. Reasoning tokens follow a linear reveal schedule, with higher-confidence predictions committed first. For a positive warm-up ratio $\alpha$, image generation begins at step $\lfloor\alpha D\rfloor$. Image tokens then follow a cosine masking schedule, with higher-confidence predictions retained and lower-confidence predictions re-masked for refinement, following MaskGIT~\citep{chang2022maskgit}. The default settings $D=128$ and $\alpha=1/4$ place the start of image generation at step 32.

\noindent \textbf{Initial action commitment.}~Let $\mathcal{A}$ denote the token positions in the action span and let $\tilde{\mathbf{x}}_d$ denote the input to denoising step $d$. Before initial commitment, the model selects the highest-probability valid token $v_i^{(d)}$ at each action position $i\in\mathcal{A}$. The confidence of this candidate action is the geometric mean of selected tokens' decoding probabilities:
\begin{equation}
    c_d =
    \exp\!\left(
        \frac{1}{|\mathcal{A}|}
        \sum_{i\in\mathcal{A}}
        \log p_\theta(v_i^{(d)}\mid\tilde{\mathbf{x}}_d)
    \right).
\end{equation}
The model commits the full candidate when $c_d\geq\tau_c$, with $\tau_c=0.4$. For $\alpha>0$, an action that has yet to reach this threshold is committed immediately before image generation begins. In the $\alpha=0$ ablation, image generation begins at step 1, and the action is committed when its confidence reaches $\tau_c$ or, at the latest, immediately before the first scheduled check.

\noindent \textbf{Check order and timing.}~RecD applies checks after denoising steps $\lfloor D/2\rfloor$ and $\lfloor 3D/4\rfloor$. At each check position, the action check runs first. The imagine check follows immediately under the context left by the action check, including when the action has been rejected. Each check uses one additional model evaluation. Both checks occur before the next denoising step.

\noindent \textbf{Action check.}~For each $i\in\mathcal{A}$, let $\bar{x}_i$ denote the value at position $i$ of the committed action as it stands before the check. We save the current contents of the action span and the NEXT STEP and PREDICTION slots, then temporarily mask all three spans. Let $\tilde{\mathbf{x}}^{a}$ denote this check input. The remaining reasoning and the partially revealed next frame retain their pre-check visibility. The action support score is
\begin{equation}
    q_a =
    \exp\!\left(
        \frac{1}{|\mathcal{A}|}
        \sum_{i\in\mathcal{A}}
        \log p_\theta(\bar{x}_i\mid\tilde{\mathbf{x}}^{a})
    \right).
\end{equation}
Every term is the probability assigned at position $i$ to its previously committed token $\bar{x}_i$. All terms are read from the same model evaluation with the entire action masked. For gaming, $q_a$ reduces to the probability of the single committed action token. For navigation, the geometric mean gives one score for the full span. If $q_a\geq\tau_a$, we restore the action and the two reasoning slots to their pre-check contents, including their original visibility pattern. If $q_a<\tau_a$, the action is rejected and all three spans remain masked.

\noindent \textbf{Imagine check.}~Let $\mathcal{R}$ denote the set of next-frame positions revealed immediately before the imagine check, and let $\bar{x}_i$ be the stored visual token at each $i\in\mathcal{R}$. We temporarily mask all positions in $\mathcal{R}$ together. Positions that were already masked stay masked, so the entire next frame is hidden during scoring. Let $\tilde{\mathbf{x}}^{o}$ denote the resulting input. If the action check retained the action, the restored action is visible in $\tilde{\mathbf{x}}^{o}$. If it rejected the action, the action and both associated reasoning slots are still masked.

A single model evaluation gives the visual support score for each previously revealed token:
\begin{equation}
    q_i = p_\theta(\bar{x}_i\mid\tilde{\mathbf{x}}^{o}),
    \qquad i\in\mathcal{R}.
\end{equation}
Thus, $q_i$ is the probability assigned at position $i$ to its stored visual token under the common check input. Each position is assessed separately. We restore $\bar{x}_i$ if $q_i\geq\tau_o$ and leave it masked otherwise, with $\tau_o=0.3$. Positions outside $\mathcal{R}$ in the next-frame region remain masked.

\noindent \textbf{Regeneration and rollout update.}~After both checks, denoising resumes from the resulting sequence. A rejected action is regenerated and committed as a whole in the next denoising step without reapplying $\tau_c$. Regeneration may produce the same action again. Re-masked reasoning and visual tokens follow their respective denoising schedules. With $D=128$, checks follow denoising steps 64 and 96, and denoising resumes at steps 65 and 97, respectively. At the end of denoising step $D$, the completed next observation is appended to the rolling context for environment step $t+1$.

\section{Additional Experimental Details}
\label{sec:app-experiments}

\subsection{Implementation Details}
\label{sec:app-implementation}

\noindent \textbf{Model and optimization.}~We initialize \ours{} from Lumina-DiMOO-8B~\citep{xin2025lumina} and keep the pretrained text and image tokenizers fixed. We optimize the model with AdamW~\citep{loshchilov2017decoupled}, a learning rate of $2\times10^{-6}$, and an effective batch size of 4. Stage~1 and Stage~2 use approximately \(300\mathrm{K}\) and \(150\mathrm{K}\) optimization updates, respectively. Training uses four NVIDIA A100 GPUs with 80\,GB memory each.

\noindent \textbf{Inputs and rollout configuration.}~Observations are processed at resolution $576\times448$ for Maze, $512\times512$ for FrozenLake and both MiniBehavior views, and $448\times576$ for navigation (height $\times$ width). For our instruction-guided world modeling task, we use one initial observation for gaming and four for navigation, retaining at most four recent observations in the context.

\noindent \textbf{Default masking and decoding settings.}~Stage~1 mixes forward and reverse modes at a 1:1 ratio, and both stages use $\beta_{\min}=0.3$. Inference uses $D=128$ denoising steps with $\alpha=1/4$, so image generation begins at step 32. Action and imagine checks follow steps 64 and 96. We set $\tau_c=0.4$, $\tau_a=0.5$, and $\tau_o=0.3$. Secs.~\ref{sec:app-training} and~\ref{sec:app-recd} describe the training masks, denoising schedules, and reciprocal checks.

\noindent \textbf{Run configuration.}~For \ours{}, we disable classifier-free guidance (CFG) for action generation, and use text- and image-conditioning guidance coefficients of \(2.5\) and \(1.0\), respectively, for observation generation in both gaming and navigation. Within each evaluation protocol, methods and ablation variants use the same evaluation episodes.

\subsection{Baseline Configurations}
\label{sec:app-baselines}

We compare multimodal models and world models under the evaluation protocols described in Sec.~5.1 of the main paper. For the instruction-guided world modeling comparison in Tab.~1, all task-trained baselines use the same training split as \ours{}. The separate LCVN benchmark comparison in Tab.~2 uses the LCVN training and validation splits.

\noindent \textbf{Multimodal models.}~We evaluate Anole-7B~\citep{chern2024anole}, BAGEL-7B-MoT (w/ Think)~\citep{deng2025emerging}, MMaDA-8B~\citep{yangmmada}, Lumina-DiMOO-8B~\citep{xin2025lumina}, and two MMaDA-Parallel variants~\citep{tian2025mmada}. MMaDA-Parallel-M uses the MMaDA-8B backbone, and MMaDA-Parallel-A uses Lumina-DiMOO-8B. We evaluate each model both zero-shot and after fine-tuning (FT) on our task data. We also evaluate MVoT~\citep{li2025imagine} after fine-tuning on the same data. All FT variants and MVoT use our answer format to jointly predict the reasoning, action, and next observation at each environment step. Training follows each model's architecture. Autoregressive components use next-token prediction, and diffusion components use their respective denoising objectives.

\noindent \textbf{World models.}~For Tab.~1, we adapt UniWM~\citep{dong2026towards}, LCVN-Uni~\citep{dong2026lcvn}, and the original and language-conditioned variants of NWM~\citep{bar2025navigation,dong2026lcvn} to our task inputs and scenario-specific action spaces. Both NWM variants use LCVN-AC for action prediction, following~\citep{dong2026lcvn}. For the LCVN benchmark in Tab.~2, the results of NWM, NWM (lang), LCVN-WM, and LCVN-Uni are taken from~\citep{dong2026lcvn}, and we train UniWM on the LCVN training split with the same configuration as for Tab.~1. Unless otherwise stated, the configurations below describe the models trained for Tab.~1.

\noindent \textbf{NWM~\citep{bar2025navigation}.}~We use a CDiT-XL backbone with approximately 1B parameters and a context length of 4, following the reported NWM configuration~\citep{bar2025navigation}. NWM generates each observation using 250 denoising steps. For Tab.~1, we train NWM and LCVN-AC from scratch on our task training split.

\noindent \textbf{LCVN-WM~\citep{dong2026lcvn}.}~LCVN-WM is a diffusion-based world model that generates observations in the latent space of an ImageVAE and is paired with LCVN-AC for action prediction. It appears only in Tab.~2, where we report the results of its $K=4$ configuration from~\citep{dong2026lcvn}.

\noindent \textbf{LCVN-Uni~\citep{dong2026lcvn}.}~We fine-tune LCVN-Uni from GAIR Anole-7B~\citep{chern2024anole} with a 4096-token context window and keep the bin, text, and image tokenizers fixed. Images are resized to $448\times448$. We update only rank-16 LoRA adapters~\citep{hu2022lora} in the Transformer's query, key, and value projections. We train for 20 epochs using AdamW with a learning rate of $2\times10^{-4}$. With four NVIDIA A100 GPUs (80\,GB each), a per-GPU batch size of 1, and two gradient accumulation steps, the effective batch size is 8.

\noindent \textbf{UniWM~\citep{dong2026towards}.}~We fine-tune UniWM~\citep{dong2026towards} from GAIR Anole-7B~\citep{chern2024anole} with a 4096-token context window and keep the tokenizers fixed. Each input image is resized to $448\times448$ and encoded as 784 visual tokens. We update only rank-16 LoRA adapters~\citep{hu2022lora} in the Transformer's query, key, and value projections. We train for 20 epochs using AdamW with a learning rate of $2\times10^{-4}$ and an effective batch size of 8 on four NVIDIA A100 GPUs (80\,GB each). At inference, the boundary tokens \texttt{<boss>} and \texttt{<eoss>} trigger the storage of key and value states from decoder layers 0, 7, 15, 23, and 31 in the intra-step memory bank.

\subsection{Evaluation Protocols}
\label{sec:app-evaluation}

We evaluate instruction-guided world modeling on 8{,}909 held-out episodes, comprising 3{,}421 global-grid gaming episodes, 1{,}964 egocentric-grid gaming episodes, and 3{,}524 egocentric navigation episodes. The separate LCVN experiment uses its own training and validation splits.

\noindent \textbf{Aggregation.}~We report results at two levels.
\emph{Domain level} (Tab.~1 of the main paper): Gaming pools the
global-grid and egocentric-grid test episodes (5{,}385 in total), and
Navigation pools the test episodes of the four source datasets (3{,}524
in total). SR, SPL, DreamSim, Cons., and RG are computed per episode and
averaged over all test episodes of the domain. FID is a set-level statistic: we compute it separately for each
environment--observation setting in gaming and each source dataset in
navigation, and report the unweighted mean of these values.
\emph{Overall level}: Fig.~\ref{fig:paradigm}(d), Tabs.~\ref{tab:abl_training}, \ref{tab:abl_decoding}, and~\ref{tab:abl_context}, and Fig.~\ref{fig:abl_schedule} in the main text, together with Tab.~\ref{tab:abl_schedule} in this appendix, report the unweighted mean of the Gaming and Navigation scores.

\noindent \textbf{Planning in gaming.}~We execute the generated actions in the simulator and record whether the resulting state satisfies the instruction. Success rate (SR) is the fraction of successful episodes~\citep{dong2026havln}. For success weighted by path length (SPL)~\citep{anderson2018evaluation}, the reference length is the minimum number of actions required to complete the task. This count includes both movement and any required object interactions, such as picking up, placing, or activating an object.

\noindent \textbf{Planning in navigation.}~We parse each generated action into a planar displacement and a yaw change in the robot's local frame. Starting from the initial reference pose, we rotate each displacement into the world frame using the current heading, add it to the current position, and then update the heading by the predicted yaw change. An episode succeeds if the distance between the predicted and reference endpoints is strictly below the mean step length of that episode's reference trajectory~\citep{dong2026lcvn}. For SPL, we use the length of the recorded reference path. An episode with a missing or unparsable action remains in the evaluation set and receives zero for both SR and SPL.

\noindent \textbf{Visual prediction.}~We evaluate generated observations against the recorded reference trajectories using FID~\citep{heusel2017fid} and DreamSim~\citep{fu2023dreamsim}. Rollout FID compares the set of generated rollout frames with the corresponding set of reference frames. Rollout DreamSim averages distances between generated and reference frames aligned by episode and step. For final-frame FID, each episode contributes only its final frame to each set. Final-frame DreamSim compares the generated and reference final frames of each episode. A successful alternative route can differ visually from the recorded route, so we interpret these metrics alongside planning success and action--observation consistency.

\noindent \textbf{Action--observation consistency.}~We use GPT-5.6 Sol~\citep{openai2026gpt56} to assess whether the observable transition between two consecutive rollout observations agrees with the generated action for that step. The judge assigns a score of 1 when the transition is consistent with the action and 0 when it is inconsistent or the visual evidence is insufficient to decide. For actions whose correct outcome leaves the observation unchanged, such as \acttok{toggle}, a blocked move, or the terminal action, the step is consistent when the two observations show no task-relevant change. A step is scored as having insufficient evidence only when the generated
observations are too blurry or ambiguous for the judge to tell what changed. We average these binary judgments over steps within each episode and then over episodes to obtain Cons.

\noindent \textbf{Reasoning grounding.}~The same judge assigns a binary score to each of the four reasoning slots. For SCENE, the evidence is the current context observations. For PROGRESS, it is the observation history and task context. For NEXT STEP, the judge checks whether the stated next step agrees with the generated action and the relevant context. For PREDICTION, it checks whether the anticipated outcome agrees with the generated next observation and action. A slot receives 1 if the evidence supports its essential content without a clear contradiction, and 0 otherwise. Missing or uninformative slots receive 0. We average over the four slots, then over steps within each episode, and finally over episodes. We report RG only for methods that generate reasoning.

\noindent \textbf{Judge configuration.}~We use two fixed prompt templates for Cons. and RG. The judge receives no information identifying the generating method. All methods and ablation variants are evaluated using the same prompts, scoring criteria, and inference settings. We validate the automated judgments on stratified subsets of generated test steps from gaming and navigation. Three annotators independently apply the same binary scoring criteria while blinded to the generating method and the automated judgments. For Cons., each annotator rates the action--observation transition at each of 400 sampled steps. For RG, each annotator rates the four reasoning slots separately at each of 200 sampled steps. We take the majority vote of the three annotators as the human reference for each transition or slot. Fleiss' $\kappa$ among the three annotators is 0.92 for Cons. and 0.95 for RG. Against the human majority, the automated judge achieves 97\% agreement for Cons. (Cohen's $\kappa=0.94$) and 96\% for RG (Cohen's $\kappa=0.90$). For RG, these agreement statistics are computed on the individual binary slot judgments before averaging them into an RG score.

\noindent \textbf{LCVN benchmark.}~For Tab.~2 of the main paper, we train a separate \ours{} model initialized directly from the pretrained Lumina-DiMOO-8B checkpoint~\citep{xin2025lumina}. All task-specific training uses the LCVN training split~\citep{dong2026lcvn}, for which we generate step-level reasoning annotations in the same four-slot format: SCENE, PROGRESS, NEXT STEP, and PREDICTION. These annotations are based on the training episodes' observations, actions, and language instructions. We then apply the same two-stage training scheme, mixing forward and reverse modes in Stage~1 and continuing with joint mode in Stage~2. At inference, we apply RecD to the resulting checkpoint. The \ours{} w/o checks variant uses the same checkpoint with both reciprocal checks disabled. We report absolute trajectory error (ATE), relative pose error (RPE), and SR on the LCVN \textit{val seen} and \textit{val unseen} splits, following the evaluation protocol of~\citep{dong2026lcvn}. Tab.~2 reports SR as a fraction, while Tab.~1 reports it as a percentage. For LCVN-Uni, we report the $K=2$ configuration from~\citep{dong2026lcvn}; all other methods use $K=4$.

\subsection{Ablation Configurations}
\label{sec:app-ablations}

We describe the training variants in Tab.~3, the decoding variants in Tab.~4, and the schedule ablations in Fig.~\ref{fig:abl_schedule} of the main paper, together with additional ablations of visual context size and reasoning supervision. Unless otherwise specified, each variant uses the default settings in Sec.~\ref{sec:app-implementation}.

\noindent \textbf{Training schemes.}~All variants in Tab.~3 start from the same pretrained initialization, use the same total number of optimization updates, and are evaluated with identical RecD settings. Joint-only training uses joint mode for all updates. The forward-only Stage~1 variant uses forward mode for every Stage~1 update and then switches to joint mode in Stage~2. The full training scheme mixes forward and reverse modes at a 1:1 ratio in Stage~1 and uses joint mode in Stage~2.

\noindent \textbf{Asymmetric masking.}~To ablate asymmetric masking, we fix $\beta=1$ in all three training modes. The reasoning and next-frame regions then use the same base masking ratio $\rho$ wherever token-level masking applies. In reverse mode, NEXT STEP and PREDICTION are still masked in full, and $\rho$ applies to the remaining reasoning tokens. Each mode retains its original action-masking rule and supervised regions.

\noindent \textbf{Partial-frame conditioning.}~To test the role of partial visual evidence in reverse mode, we set $\rho_{\mathrm{img}}=0$ for every reverse-mode example, leaving the complete next frame visible for action inference. The full model uses $\rho_{\mathrm{img}}=\beta\rho$ in reverse mode. Both variants use the same action and reasoning masks and supervise the same regions.

\noindent \textbf{Decoding strategies.}~All variants in Tab.~4 use the same trained checkpoint. The default denoising budget is $D=128$ denoising steps per environment step. The following variants change how this budget is used, when actions are committed, or which reciprocal checks are applied.

\begin{itemize}
    \item \textbf{Serial.} Denoising steps 1--31 generate the reasoning and action while the entire next frame remains masked. The reasoning and action are then fixed, and steps 32--128 generate next frame. Both checks are disabled.

    \item \textbf{Joint.} Reasoning, action, and next-frame tokens are generated using the base schedules in Sec.~\ref{sec:app-recd}, including the initial action commitment rule. Both checks are disabled. This is the \ours{} w/o checks configuration.

    \item \textbf{Joint with $D=132$.} Full RecD uses 128 denoising steps and two check positions. Each position adds one model evaluation for the action check and one for the imagine check, giving $128+2\times2=132$ model evaluations per environment step. This variant allocates all 132 evaluations to base denoising and disables both checks. Warm-up ratio remains $\alpha=1/4$, so image generation begins at step 33.

    \item \textbf{Late-$k$.} For $k\in\{64,96\}$, the action region and the NEXT STEP and PREDICTION slots remain masked until denoising step $k$. Image generation begins at step 32. Before action commitment, the model can use the context prefix and the revealed SCENE and PROGRESS tokens, together with the visual tokens revealed so far. At step $k$, the model commits the action as a whole using the available visual evidence, without applying $\tau_c$. The two reasoning slots are then revealed on a linear schedule over the remaining $D-k$ steps. Both checks are disabled.

    \item \textbf{Action-only and imagine-only.} Each variant enables one type of check at both scheduled positions. In the action-only variant, a rejected action is regenerated at the next denoising step. In the imagine-only variant, the committed action and the current reasoning remain available when the model scores the visual tokens.

    \item \textbf{NoVis.} Both checks are enabled. During the action check, we temporarily hide all revealed next-frame tokens along with the action, NEXT STEP, and PREDICTION. The model therefore computes $q_a$ using the context prefix and the remaining visible reasoning. If the action is rejected, we also hide the next frame when regenerating the action in the next denoising step. This additional visual mask applies only to action checking and rejected-action regeneration. The imagine check uses its usual scoring and restoration rules, and all thresholds retain default values.

    \item \textbf{Full.} At each scheduled position, RecD applies the action check followed immediately by the imagine check. A rejected action is still masked when the imagine check runs and is regenerated in the next denoising step.
\end{itemize}

\noindent \textbf{Warm-up ratio and denoising budget.}~For the warm-up ablation in Fig.~\ref{fig:abl_schedule}, we vary $\alpha\in\{0,1/8,1/4,3/8\}$ while holding $D=128$ and keeping the checks after steps 64 and 96. When $\alpha=0$, image generation begins in the first denoising step. The action is committed when its confidence reaches $\tau_c$ or, at the latest, immediately before the first scheduled check. For the denoising-budget ablation, we vary $D\in\{64,128,256\}$ while keeping $\alpha=1/4$ and placing the checks after $\lfloor D/2\rfloor$ and $\lfloor3D/4\rfloor$.

\noindent \textbf{Check positions.}~We evaluate five pairs of relative check positions: $\{3/8,1/2\}$, $\{3/8,5/8\}$, $\{1/2,3/4\}$, $\{5/8,7/8\}$, and $\{3/4,7/8\}$. For a relative position $s$, the checks run after denoising step $\lfloor sD\rfloor$. Every schedule uses the same thresholds and applies the action check followed by the imagine check at each position.

\noindent \textbf{Number of check positions.}~For one check position, we report the mean of the results for $\{1/2\}$ and $\{3/4\}$. For two positions, we use the default pair $\{1/2,3/4\}$. For three positions, we evaluate three schedules, each formed by adding one of $\{3/8,5/8,7/8\}$ to the default pair, and average their results. Schedules receive equal weight within each count. Each position includes both types of check, and the denoising budget and thresholds are fixed across schedules.

\noindent \textbf{Visual context size.}~We evaluate $K\in\{1,2,4\}$ at a fixed training checkpoint. All variants start with the same available observation history: $H=1$ for gaming and $H=4$ for navigation. At environment step $t$, each variant receives only the most recent $\min(K,H+t)$ frames. Thus, for navigation with $K=1$, the model receives only the latest of the four available initial frames.

\noindent \textbf{Reasoning supervision.}~We train three variants separately with matched total optimization updates. The first omits the reasoning region, the second includes only the four slot headers, and the third uses the full reasoning annotations. Reasoning supervision and slot-specific masking apply only to the tokens present in each variant. The action and next-frame regions follow the masking and supervision rules of each training mode.

\subsection{Full Results of Decoding Schedule Ablations}
\label{sec:app-additional-results}

Tab.~\ref{tab:abl_schedule} provides the complete numerical results for the decoding schedule ablations summarized in Fig.~\ref{fig:abl_schedule}, including SPL and consistency in addition to SR and final-frame DreamSim. Each metric is reported as the unweighted mean of the gaming and navigation scores.

\begin{table}[t]
    \centering
    \caption{\textbf{Decoding ablations} of warm-up ratio, denoising steps, and check schedules. Bold values indicate default settings. Final DS denotes final-frame DreamSim.}
    \label{tab:abl_schedule}
    \small
    \setlength{\tabcolsep}{4pt}
    \resizebox{\columnwidth}{!}{
    \begin{tabular}{p{4.6cm}|p{2.8cm}<{\centering}IIp{1.65cm}<{\centering}p{1.65cm}<{\centering}p{2.0cm}<{\centering}p{1.8cm}<{\centering}}
        \hline\thickhline
         \rowcolor{gray!20}Setting & Value
        & SR$\uparrow$ & SPL$\uparrow$
        & Final DS$\downarrow$ & Cons.$\uparrow$ \\
        \hline\hline
        
        & $0$
        & 74.9 & 71.8 & 0.148 & 0.838 \\
        & $1/8$
        & 77.8 & 74.5 & 0.135 & 0.879 \\
        \rowcolor{recjointbg!80}\cellcolor{white}& $\mathbf{1/4}$
        & \textbf{79.1} & \textbf{75.7} & \textbf{0.130} & \textbf{0.895} \\
        \multirow{-4}{*}{Warm-up ratio $\alpha$}& $3/8$
        & 77.2 & 73.9 & 0.141 & 0.881 \\
        \hline
        
        & 64
        & 75.6 & 72.7 & 0.154 & 0.856 \\
        \rowcolor{recjointbg!80}\cellcolor{white}& \textbf{128}
        & \textbf{79.1} & \textbf{75.7} & \textbf{0.130} & \textbf{0.895} \\
        \multirow{-3}{*}{Denoising steps $D$}& 256
        & 79.5 & 75.9 & 0.126 & 0.899 \\
        \hline
        
        & $\{3/8,1/2\}$
        & 76.1 & 71.9 & 0.139 & 0.864 \\
        & $\{3/8,5/8\}$
        & 77.6 & 73.8 & 0.134 & 0.878 \\
        \rowcolor{recjointbg!80}\cellcolor{white}& $\{\mathbf{1/2},\mathbf{3/4}\}$
        & \textbf{79.1} & \textbf{75.7} & \textbf{0.130} & \textbf{0.895} \\
        & $\{5/8,7/8\}$
        & 78.0 & 74.8 & 0.131 & 0.889 \\
        \multirow{-5}{*}{Check positions}& $\{3/4,7/8\}$
        & 76.4 & 73.3 & 0.136 & 0.872 \\
        \hline
        
        & 1
        & 76.8 & 73.6 & 0.137 & 0.871 \\
        \rowcolor{recjointbg!80}\cellcolor{white}& \textbf{2}
        & \textbf{79.1} & \textbf{75.7} & \textbf{0.130} & \textbf{0.895} \\
        \multirow{-3}{*}{Number of check positions}& 3
        & 79.6 & 76.0 & 0.128 & 0.898 \\
        \hline
    \end{tabular}}
     \vspace{-0.1cm}
\end{table}

\begin{figure*}[t]
    \centering
    \includegraphics[width=\linewidth]{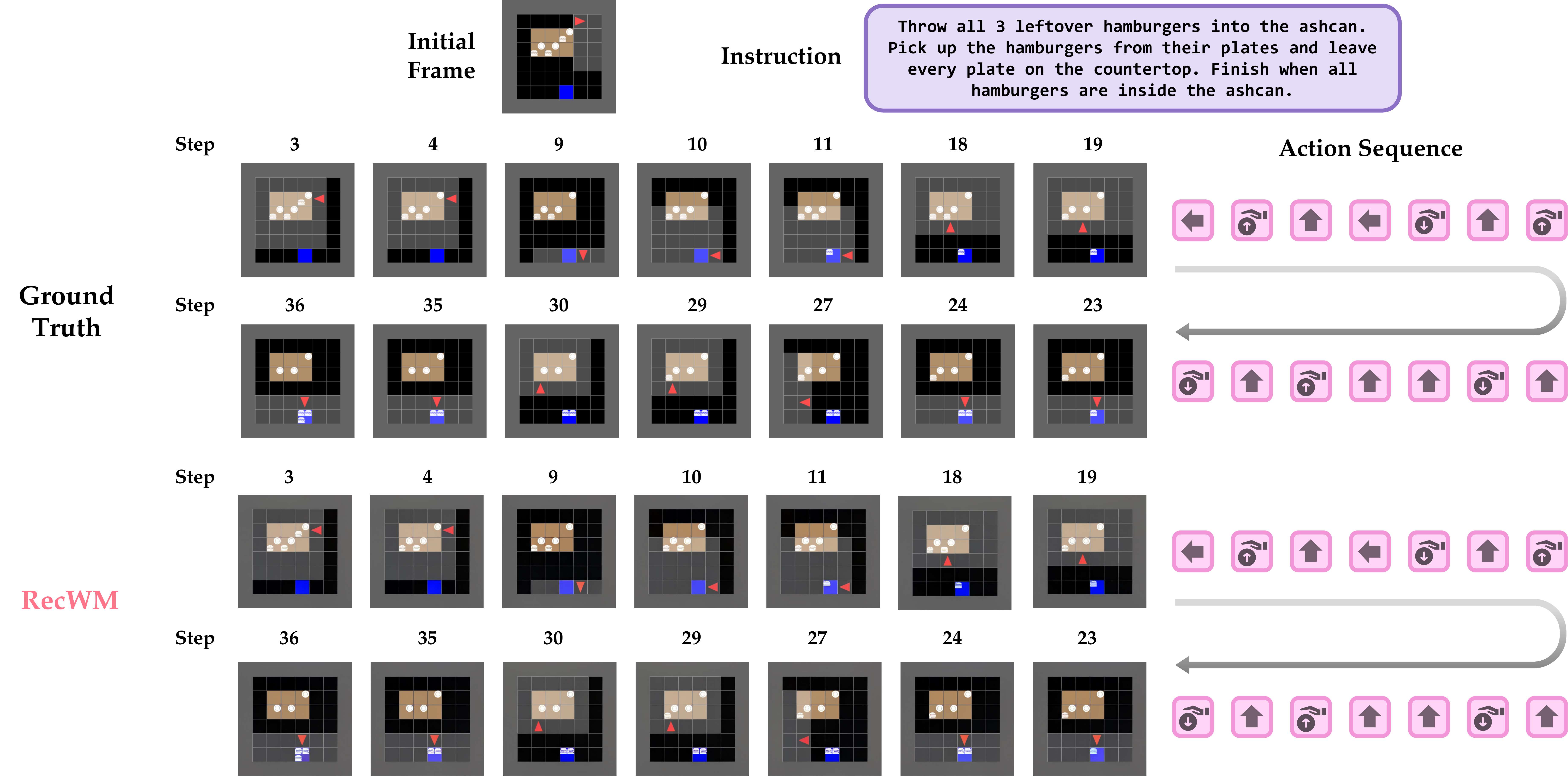}
    \caption{\textbf{Additional qualitative results in egocentric-grid gaming.} On the MiniBehavior hamburger task, \ours{} matches the ground-truth actions and observations at the selected steps. In each rollout, the second row reads right to left.}
    \label{fig:ego_res}
\end{figure*}

\subsection{Reciprocal Decoding Analysis}
\label{sec:app-check-analysis}

Fig.~\ref{fig:check_activation} of the main paper measures how often each check intervenes during full RecD rollouts. Tab.~\ref{tab:check_effects} measures the effect of a triggered intervention by completing the same environment step from a shared intermediate decoding state, with and without the selected checks.

\noindent \textbf{Action-check activation.}~At each check position in a full RecD rollout, we record whether the action is rejected and, upon rejection, whether the action regenerated in the next denoising step differs from the original. The rejection rate uses all samples evaluated at that check position as its denominator. The action change rate uses only samples with a rejected action. For navigation, we compare the parsed displacement and yaw values so that a change in wording alone does not count as an action change. Statistics at the second check position include the effects of the first checks and the intervening denoising steps.

\noindent \textbf{Imagine-check activation.}~We report two quantities with different denominators. The trigger rate is the fraction of samples in which the imagine check re-masks at least one previously revealed next-frame token. The token re-masking fraction is computed within each sample by dividing the number of re-masked visual tokens by the number of visual tokens revealed immediately before the check. We then average these per-sample fractions, including a value of zero when all revealed tokens are restored. Both statistics are reported over all samples and separately for samples whose preceding action check retained or rejected the action. Each group uses its own denominator. 

\noindent \textbf{Paired interventions.}~For each comparison in Tab.~\ref{tab:check_effects}, we copy the decoding state immediately before the selected checks into two branches. The intervention branch applies the action check alone, the imagine check alone, or both checks in sequence. The control branch skips the checks at that position. Both branches use the same base denoising settings and random seed, disable all subsequent checks, and finish the current environment step. We evaluate the completed action and next observation from each branch.

At $\lfloor D/2\rfloor$, both branches skip the later checks at $\lfloor3D/4\rfloor$. At $\lfloor3D/4\rfloor$, the shared starting state already incorporates the default checks at $\lfloor D/2\rfloor$ and the subsequent denoising steps. In the imagine-only intervention, the committed action remains visible during visual scoring. In the combined intervention, the imagine check uses the state left by the action check. A rejected action remains masked until the next denoising step.

\noindent \textbf{Triggered subsets.}~We retain a pair for analysis when the intervention branch triggers re-masking. For the action-only intervention, the trigger is action rejection. For the imagine-only intervention, the trigger is the re-masking of at least one previously revealed visual token. For the combined intervention, either event qualifies. An action rejection counts as a trigger even if regeneration later produces the same action. Each intervention defines its own subset of paired samples.

\noindent \textbf{Fix and Harm.}~We score each branch using the binary action--observation consistency criterion defined in Sec.~\ref{sec:app-evaluation}. For triggered pair $n$, let $c_n^0$ be the judgment for the control branch and $c_n^1$ the judgment for the intervention branch, where 1 denotes a consistent step and 0 an inconsistent or uncertain step. Let $N$ be the number of triggered pairs for the selected intervention and check position. We compute
\begin{equation}
\begin{aligned}
    \mathrm{Fix}  &= \frac{100}{N}\sum_{n=1}^{N}\mathbb{I}[c_n^0=0,\ c_n^1=1], \\
    \mathrm{Harm} &= \frac{100}{N}\sum_{n=1}^{N}\mathbb{I}[c_n^0=1,\ c_n^1=0].
\end{aligned}
\end{equation}
Both percentages use all $N$ triggered pairs as their denominator, including pairs whose judgments agree across branches. Their difference, $\mathrm{Fix}-\mathrm{Harm}$, is the change in consistency on that triggered subset in percentage.

\subsection{More Qualitative Results}
\label{sec:app-visualizations}

Fig.~\ref{fig:ego_res} shows a \ours{} rollout on the egocentric MiniBehavior hamburger task alongside the ground truth. At the 14 selected steps, the generated actions match the reference actions, and the generated observations show the same agent pose, field of view, and hamburger placement as the reference frames. In particular, the highlighted field of view changes with each turn while the countertop and ashcan stay in place.

\clearpage\clearpage
\bibliographystyle{assets/plainnat}
\bibliography{main}

@article{hafner2023mastering,
  title={Mastering diverse control tasks through world models},
  author={Hafner, Danijar and Pasukonis, Jurgis and Ba, Jimmy and Lillicrap, Timothy},
  journal={Nature},
  volume={640},
  number={8059},
  pages={647--653},
  year={2025},
  publisher={Nature Publishing Group UK, London},
  doi = {10.1038/s41586-025-08744-2},
  url = {https://arxiv.org/abs/2301.04104},
}

@inproceedings{hafner2022masteringataridiscreteworld,
title={Mastering Atari with Discrete World Models},
author={Danijar Hafner and Timothy P Lillicrap and Mohammad Norouzi and Jimmy Ba},
booktitle={International Conference on Learning Representations},
year={2021},
url={https://arxiv.org/abs/2010.02193}
}

@inproceedings{wu2025janus,
  title={{Janus}: Decoupling Visual Encoding for Unified Multimodal Understanding and Generation},
  author={Wu, Chengyue and Chen, Xiaokang and Wu, Zhiyu and Ma, Yiyang and Liu, Xingchao and Pan, Zizheng and Liu, Wen and Xie, Zhenda and Yu, Xingkai and Ruan, Chong and Luo, Ping},
  booktitle={IEEE/CVF Conference on Computer Vision and Pattern Recognition},
  pages={12966--12977},
  year={2025},
  organization={IEEE},
  url = {https://arxiv.org/abs/2410.13848},
}

@article{liu2026lumina,
  title={{Lumina-mGPT}: Flexible Photorealistic Autoregressive Text-to-Image Generation},
  author={Liu, Dongyang and Xin, Yi and Zhao, Shitian and Zhuo, Le and Lin, Weifeng and Li, Xinyue and Qin, Qi and Zhai, Guangtao and Liu, Xiaohong and Li, Hongsheng and Qiao, Yu and Gao, Peng},
  journal={International Journal of Computer Vision},
  volume={134},
  number={4},
  pages={141},
  year={2026},
  publisher={Springer},
  doi = {10.1007/s11263-026-02795-2},
  url = {https://arxiv.org/abs/2408.02657},
}

@inproceedings{wu2025omnigen2,
  title={{OmniGen2}: Towards Instruction-Aligned Multimodal Generation},
  author={Wu, Chenyuan and Wang, Jiahao and Zheng, Pengfei and Yan, Ruiran and Xiao, Shitao and Luo, Xin and Wang, Yueze and Li, Wanli and Jiang, Xiyan and Liu, Yexin and Zhou, Junjie and Xia, Ziyi and Liu, Ze and Li, Chaofan and Deng, Haoge and Luo, Kun and Zhang, Bo and Zhang, Jiajun and Liu, Dong and Lian, Defu and Wang, Xinlong and Wang, Zhongyuan and Huang, Tiejun and Liu, Zheng},
  booktitle={IEEE/CVF Conference on Computer Vision and Pattern Recognition},
  pages={21964--21975},
  year={2026},
  url = {https://arxiv.org/abs/2506.18871},
}

@article{cao2025hunyuanimage,
  title={HunyuanImage 3.0 Technical Report},
  author={Team, Tencent Hunyuan Foundation Model},
  journal={arXiv preprint arXiv:2509.23951},
  year={2025},
  url = {https://arxiv.org/abs/2509.23951},
}

@article{deng2025emerging,
  title={Emerging Properties in Unified Multimodal Pretraining},
  author={Deng, Chaorui and Zhu, Deyao and Li, Kunchang and Gou, Chenhui and Li, Feng and Wang, Zeyu and Zhong, Shu and Yu, Weihao and Nie, Xiaonan and Song, Ziang and others},
  journal={arXiv preprint arXiv:2505.14683},
  year={2025},
  url = {https://arxiv.org/abs/2505.14683},
}

@inproceedings{lou2023discrete,
  title={Discrete Diffusion Modeling by Estimating the Ratios of the Data Distribution},
  author={Aaron Lou and Chenlin Meng and Stefano Ermon},
  booktitle={International Conference on Machine Learning},
  pages={32819--32848},
  year={2024},
  organization={PMLR},
  volume = {235},
  series = {Proceedings of Machine Learning Research},
  url = {https://arxiv.org/abs/2310.16834},
}

@inproceedings{liu2026palm,
  title={{PALM}: Progress-Aware Policy Learning via Affordance Reasoning for Long-Horizon Robotic Manipulation},
  author={Liu, Yuanzhe and Zhu, Jingyuan and Mo, Yuchen and Li, Gen and Cao, Xu and Jin, Jin and Shen, Yifan and Li, Zhengyuan and Yu, Tianjiao and Yuan, Wenzhen and Ding, Fangqiang and Lourentzou, Ismini},
  booktitle={IEEE/CVF Conference on Computer Vision and Pattern Recognition},
  pages={28096--28110},
  year={2026},
  url = {https://arxiv.org/abs/2601.07060},
}

@inproceedings{sahoo2024simple,
  title={Simple and Effective Masked Diffusion Language Models},
  author={Sahoo, Subham S. and Arriola, Marianne and Schiff, Yair and Gokaslan, Aaron and Marroquin, Edgar and Chiu, Justin T. and Rush, Alexander and Kuleshov, Volodymyr},
  volume={37},
  pages={130136--130184},
  year={2024},
  booktitle = {Advances in Neural Information Processing Systems},
  doi = {10.52202/079017-4135},
  url = {https://arxiv.org/abs/2406.07524},
}

@inproceedings{sridhar2024nomad,
    title={{NoMaD}: Goal Masked Diffusion Policies for Navigation and Exploration},
    author={Sridhar, Ajay and Shah, Dhruv and Glossop, Catherine and Levine, Sergey},
    booktitle={IEEE International Conference on Robotics and Automation},
    pages={63--70},
    year={2024},
  url = {https://arxiv.org/abs/2310.07896},
  doi = {10.1109/ICRA57147.2024.10610665},
}

@inproceedings{hafner2019dream,
title={Dream to Control: Learning Behaviors by Latent Imagination},
author={Danijar Hafner and Timothy Lillicrap and Jimmy Ba and Mohammad Norouzi},
booktitle={International Conference on Learning Representations},
year={2020},
url={https://arxiv.org/abs/1912.01603}
}

@article{hafner2025training,
  title={Training agents inside of scalable world models},
  author={Hafner, Danijar and Yan, Wilson and Lillicrap, Timothy},
  journal={arXiv preprint arXiv:2509.24527},
  year={2025},
  url = {https://arxiv.org/abs/2509.24527},
}

@inproceedings{dong2026towards,
  title={Towards Unified World Models for Visual Navigation via Memory-Augmented Planning and Foresight},
  author={Dong, Yifei and Wu, Fengyi and Chen, Guangyu and Kong, Lingdong and Hu, Qiyu and Zhou, Yuxuan and Zhu, Xu and Sun, Jingdong and He, Jun-Yan and Dai, Qi and Hauptmann, Alexander G. and Cheng, Zhi-Qi},
  booktitle={European Conference on Computer Vision},
  pages={276--295},
  year={2026},
  organization={Springer},
  url = {https://arxiv.org/abs/2510.08713},
}

@inproceedings{ha2018world,
  title={Recurrent World Models Facilitate Policy Evolution},
  author={Ha, David and Schmidhuber, J{\"u}rgen},
  volume={31},
  year={2018},
  booktitle = {Advances in Neural Information Processing Systems},
  url = {https://arxiv.org/abs/1809.01999},
}

@article{xin2025lumina,
  title={{Lumina-DiMOO}: An Omni Diffusion Large Language Model for Multi-Modal Generation and Understanding},
  author={Xin, Yi and Qin, Qi and Luo, Siqi and Zhu, Kaiwen and Yan, Juncheng and Tai, Yan and Lei, Jiayi and Cao, Yuewen and Wang, Keqi and Wang, Yibin and others},
  journal={arXiv preprint arXiv:2510.06308},
  year={2025},
  url = {https://arxiv.org/abs/2510.06308},
}

@inproceedings{tian2025mmada,
 author = {Tian, Ye and Yang, Ling and Yang, Jiongfan and Wang, Anran and Tian, Yu and Zheng, Jiani and Wang, Haochen and Teng, Zhiyang and Wang, Zhuochen and Wang, Yinjie and Tong, Yunhai and Wang, Mengdi and Li, Xiangtai},
 booktitle = {International Conference on Learning Representations},
 editor = {C. Vondrick and B. Hariharan and C. Raffel and L. Pinto and D. Yang and A. Faust},
 pages = {36089--36123},
 title = {{MMaDA-Parallel}: Multimodal Large Diffusion Language Models for Thinking-Aware Editing and Generation},
 url = {https://arxiv.org/abs/2511.09611},
 volume = {2026},
 year = {2026}
}

@inproceedings{xiao2025worldmem,
    title={{WorldMem}: Long-term Consistent World Simulation with Memory},
    author={Xiao, Zeqi and LAN, Yushi and Zhou, Yifan and Ouyang, Wenqi and Yang, Shuai and Zeng, Yanhong and Pan, Xingang},
    year={2025},
  booktitle = {Advances in Neural Information Processing Systems},
  pages = {49632--49652},
  doi = {10.52202/085713-1659},
  volume = {38},
  url = {https://arxiv.org/abs/2504.12369},
}

@inproceedings{alonso2024diffusion,
    title={Diffusion for world modeling: Visual details matter in {Atari}},
    author={Alonso, Eloi and Jelley, Adam and Micheli, Vincent and Kanervisto, Anssi and Storkey, Amos and Pearce, Tim and Fleuret, Fran{\c{c}}ois},
    volume={37},
    pages={58757--58791},
    year={2024},
  booktitle = {Advances in Neural Information Processing Systems},
  doi = {10.52202/079017-1873},
  url = {https://arxiv.org/abs/2405.12399},
}

@inproceedings{valevski2024diffusion,
  title={Diffusion models are real-time game engines},
  author={Valevski, Dani and Leviathan, Yaniv and Arar, Moab and Fruchter, Shlomi},
  booktitle={International Conference on Learning Representations},
  volume={2025},
  pages={73754--73776},
  year={2025},
  url = {https://arxiv.org/abs/2408.14837},
}

@inproceedings{yu2025gamefactory,
    title={{GameFactory}: Creating new games with generative interactive videos},
    author={Yu, Jiwen and Qin, Yiran and Wang, Xintao and Wan, Pengfei and Zhang, Di and Liu, Xihui},
    year={2025},
  booktitle = {IEEE/CVF International Conference on Computer Vision},
  pages = {11590--11599},
  url = {https://arxiv.org/abs/2501.08325},
}

@inproceedings{gumbsch2023learning,
  title={Learning Hierarchical World Models with Adaptive Temporal Abstractions from Discrete Latent Dynamics},
  author={Gumbsch, Christian and Sajid, Noor and Martius, Georg and Butz, Martin V.},
  booktitle={International Conference on Learning Representations},
  year={2024},
  pages = {4756--4792},
  url = {https://proceedings.iclr.cc/paper_files/paper/2024/hash/13b45b44e26c353c64cba9529bf4724f-Abstract-Conference.html},
}

@article{russell2025gaia,
    title={{GAIA-2}: A controllable multi-view generative world model for autonomous driving},
    author={Russell, Lloyd and Hu, Anthony and Bertoni, Lorenzo and Fedoseev, George and Shotton, Jamie and Arani, Elahe and Corrado, Gianluca},
    journal={arXiv preprint arXiv:2503.20523},
    year={2025},
  url = {https://arxiv.org/abs/2503.20523},
}

@inproceedings{koh2021pathdreamer,
    title={{PathDreamer}: A world model for indoor navigation},
    author={Koh, Jing Yu and Lee, Honglak and Yang, Yinfei and Baldridge, Jason and Anderson, Peter},
    booktitle={IEEE/CVF International Conference on Computer Vision},
    pages={14738--14748},
    year={2021},
  url = {https://arxiv.org/abs/2105.08756},
}

@inproceedings{yao2025navmorph,
    title={{NavMorph}: A Self-Evolving World Model for Vision-and-Language Navigation in Continuous Environments},
    author={Yao, Xuan and Gao, Junyu and Xu, Changsheng},
    year={2025},
  booktitle = {IEEE/CVF International Conference on Computer Vision},
  pages = {5536--5546},
  url = {https://arxiv.org/abs/2506.23468},
}

@inproceedings{bar2025navigation,
    title={Navigation World Models},
    author={Bar, Amir and Zhou, Gaoyue and Tran, Danny and Darrell, Trevor and LeCun, Yann},
    booktitle={IEEE/CVF Conference on Computer Vision and Pattern Recognition},
    pages={15791--15801},
    year={2025},
  url = {https://arxiv.org/abs/2412.03572},
}

@inproceedings{bardes2024revisiting,
    title={Revisiting feature prediction for learning visual representations from video},
    author={Bardes, Adrien and Garrido, Quentin and Ponce, Jean and Chen, Xinlei and Rabbat, Michael and LeCun, Yann and Assran, Mahmoud and Ballas, Nicolas},
    year={2025},
  url = {https://arxiv.org/abs/2404.08471},
  booktitle = {International Conference on Learning Representations},
}

@article{assran2025v,
    title={{V-JEPA} 2: Self-Supervised Video Models Enable Understanding, Prediction and Planning},
    author={Assran, Mido and Bardes, Adrien and Fan, David and Garrido, Quentin and Howes, Russell and Komeili, Mojtaba and Muckley, Matthew and Rizvi, Ammar and Roberts, Claire and Sinha, Koustuv and others},
    journal={arXiv preprint arXiv:2506.09985},
    year={2025},
  url = {https://arxiv.org/abs/2506.09985},
}

@inproceedings{bruce2024genie,
    title={Genie: Generative Interactive Environments},
    author={Bruce, Jake and Dennis, Michael D and Edwards, Ashley and Parker-Holder, Jack and Shi, Yuge and Hughes, Edward and Lai, Matthew and Mavalankar, Aditi and Steigerwald, Richie and Apps, Chris and others},
    booktitle={International Conference on Machine Learning},
    pages={4603--4623},
    year={2024},
  volume = {235},
  series = {Proceedings of Machine Learning Research},
  url = {https://arxiv.org/abs/2402.15391},
}

@article{baldassarre2025back,
    title={Back to the Features: {DINO} as a Foundation for Video World Models},
    author={Baldassarre, Federico and Szafraniec, Marc and Terver, Basile and Khalidov, Vasil and Massa, Francisco and LeCun, Yann and Labatut, Patrick and Seitzer, Maximilian and Bojanowski, Piotr},
    journal={arXiv preprint arXiv:2507.19468},
    year={2025},
  url = {https://arxiv.org/abs/2507.19468},
}

@article{agarwal2025cosmos,
    title={Cosmos world foundation model platform for physical {AI}},
    author={Agarwal, Niket and Ali, Arslan and Bala, Maciej and Balaji, Yogesh and Barker, Erik and Cai, Tiffany and Chattopadhyay, Prithvijit and Chen, Yongxin and Cui, Yin and Ding, Yifan and others},
    journal={arXiv preprint arXiv:2501.03575},
    year={2025},
  url = {https://arxiv.org/abs/2501.03575},
}

@inproceedings{xu2025U4D,
    title = {{U4D}: Uncertainty-Aware {4D} World Modeling from {LiDAR} Sequences},
    author = {Xu, Xiang and Liang, Alan and Liu, Youquan and Li, Linfeng and Kong, Lingdong and Liu, Ziwei and Liu, Qingshan},
    year = {2026},
  booktitle = {IEEE/CVF Conference on Computer Vision and Pattern Recognition},
  pages = {10027--10039},
  url = {https://arxiv.org/abs/2512.02982},
}

@inproceedings{li2025imagine,
  title={Imagine While Reasoning in Space: Multimodal Visualization-of-Thought},
  author={Chengzu Li and Wenshan Wu and Huanyu Zhang and Yan Xia and Shaoguang Mao and Li Dong and Ivan Vuli{\'c} and Furu Wei},
  booktitle={International Conference on Machine Learning},
  pages={36340--36364},
  year={2025},
  organization={PMLR},
  volume = {267},
  series = {Proceedings of Machine Learning Research},
  url = {https://arxiv.org/abs/2501.07542},
}

@article{cen2025worldvla,
  title={{WorldVLA}: Towards Autoregressive Action World Model},
  author={Cen, Jun and Yu, Chaohui and Yuan, Hangjie and Jiang, Yuming and Huang, Siteng and Guo, Jiayan and Li, Xin and Song, Yibing and Luo, Hao and Wang, Fan and others},
  journal={arXiv preprint arXiv:2506.21539},
  year={2025},
  url = {https://arxiv.org/abs/2506.21539},
}

@article{li2025unified,
  title={Unified video action model},
  author={Li, Shuang and Gao, Yihuai and Sadigh, Dorsa and Song, Shuran},
  journal={arXiv preprint arXiv:2503.00200},
  year={2025},
  url = {https://arxiv.org/abs/2503.00200},
}

@article{survey_3d_4d_world_models,
    title = {{3D} and {4D} World Modeling: A Survey},
    author = {Lingdong Kong and Wesley Yang and Jianbiao Mei and Youquan Liu and Ao Liang and Dekai Zhu and Dongyue Lu and Wei Yin and Xiaotao Hu and Mingkai Jia and Junyuan Deng and Kaiwen Zhang and Yang Wu and Tianyi Yan and Shenyuan Gao and Song Wang and Linfeng Li and Liang Pan and Yong Liu and Jianke Zhu and Wei Tsang Ooi and Steven C. H. Hoi and Ziwei Liu},
    journal = {arXiv preprint arXiv:2509.07996},
    year = {2025},
  url = {https://arxiv.org/abs/2509.07996},
}

@misc{google2024genie2,
    title = {Genie 2: A Large-Scale Foundation World Model},
    author = {Jack Parker-Holder and others},
    year = {2024},
    howpublished = {Google DeepMind technical blog},
  url = {https://deepmind.google/discover/blog/genie-2-a-large-scale-foundation-world-model/},
}

@misc{google2025genie3,
    title = {Genie 3: A New Frontier for World Models},
    author = {Philip J. Ball and Jakob Bauer and Frank Belletti and others},
    year = {2025},
    howpublished = {Google DeepMind technical blog},
  url = {https://deepmind.google/discover/blog/genie-3-a-new-frontier-for-world-models/},
}

@misc{brooks2024video,
  title={Video generation models as world simulators},
  author={Tim Brooks and Bill Peebles and Connor Holmes and Will DePue and Yufei Guo and Li Jing and David Schnurr and Joe Taylor and Troy Luhman and Eric Luhman and Clarence Ng and Ricky Wang and Aditya Ramesh},
  year={2024},
  url={https://openai.com/index/video-generation-models-as-world-simulators/},
  howpublished = {OpenAI technical report},
}

@inproceedings{zhang2025epona,
    title = {Epona: Autoregressive Diffusion World Model for Autonomous Driving},
    author = {Zhang, Kaiwen and Tang, Zhenyu and Hu, Xiaotao and Pan, Xingang and Guo, Xiaoyang and Liu, Yuan and Huang, Jingwei and Yuan, Li and Zhang, Qian and Long, Xiao-Xiao and Cao, Xun and Yin, Wei},
    booktitle = {IEEE/CVF International Conference on Computer Vision},
    pages = {27220--27230},
    year = 2025,
  url = {https://arxiv.org/abs/2506.24113},
}

@article{ivanitskiy2023configurable,
  title={A configurable library for generating and manipulating maze datasets},
  author={Ivanitskiy, Michael Igorevich and Shah, Rusheb and Spies, Alex F and R{\"a}uker, Tilman and Valentine, Dan and Rager, Can and Quirke, Lucia and Mathwin, Chris and Corlouer, Guillaume and Behn, Cecilia Diniz and others},
  journal={arXiv preprint arXiv:2309.10498},
  year={2023},
  url = {https://arxiv.org/abs/2309.10498},
}

@article{jin2023mini,
  title={{Mini-BEHAVIOR}: A Procedurally Generated Benchmark for Long-Horizon Decision-Making in Embodied {AI}},
  author={Jin, Emily and Hu, Jiaheng and Huang, Zhuoyi and Zhang, Ruohan and Wu, Jiajun and Li, Fei-Fei and Mart{\'\i}n-Mart{\'\i}n, Roberto},
  journal={arXiv preprint arXiv:2310.01824},
  year={2023},
  url = {https://arxiv.org/abs/2310.01824},
}

@article{brockman2016openai,
  title={{OpenAI Gym}},
  author={Brockman, Greg and Cheung, Vicki and Pettersson, Ludwig and Schneider, Jonas and Schulman, John and Tang, Jie and Zaremba, Wojciech},
  journal={arXiv preprint arXiv:1606.01540},
  year={2016},
  url = {https://arxiv.org/abs/1606.01540},
}

@inproceedings{yangmmada,
 author = {Yang, Ling and Tian, Ye and Li, Bowen and Zhang, Xinchen and Shen, Ke and Tong, Yunhai and Wang, Mengdi},
 booktitle = {Advances in Neural Information Processing Systems},
 doi = {10.52202/085713-4636},
 editor = {D. Belgrave and C. Zhang and H. Lin and R. Pascanu and P. Koniusz and M. Ghassemi and N. Chen},
 pages = {138867--138907},
 publisher = {Curran Associates, Inc.},
 title = {{MMaDA}: Multimodal Large Diffusion Language Models},
 url = {https://arxiv.org/abs/2505.15809},
 volume = {38},
 year = {2025}
}

@inproceedings{wu2024vsp,
  title={{VSP}: Diagnosing the Dual Challenges of Perception and Reasoning in Spatial Planning Tasks for {MLLMs}},
  author={Wu, Qiucheng and Zhao, Handong and Saxon, Michael and Bui, Trung and Wang, William Yang and Zhang, Yang and Chang, Shiyu},
  booktitle={IEEE/CVF International Conference on Computer Vision},
  pages={2270--2280},
  year={2025},
  organization={IEEE},
  url = {https://arxiv.org/abs/2407.01863},
}

@inproceedings{you2025llada,
  title={{LLaDA-V}: Large Language Diffusion Models with Visual Instruction Tuning},
  author={You, Zebin and Nie, Shen and Zhang, Xiaolu and Zhou, Jun and Lu, Zhiwu and Wen, Ji-Rong and Li, Chongxuan},
  year={2026},
  booktitle = {IEEE/CVF Conference on Computer Vision and Pattern Recognition},
  pages = {10093--10105},
  url = {https://arxiv.org/abs/2505.16933},
}

@article{team2024chameleon,
  title={Chameleon: Mixed-modal early-fusion foundation models},
  author={Team, Chameleon},
  journal={arXiv preprint arXiv:2405.09818},
  year={2024},
  url = {https://arxiv.org/abs/2405.09818},
}

@article{chern2024anole,
  title={{Anole}: An Open, Autoregressive, Native Large Multimodal Model for Interleaved Image-Text Generation},
  author={Chern, Ethan and Su, Jiadi and Ma, Yan and Liu, Pengfei},
  journal={arXiv preprint arXiv:2407.06135},
  year={2024},
  url = {https://arxiv.org/abs/2407.06135},
}

@inproceedings{xieshow,
  title={Show-o: One single transformer to unify multimodal understanding and generation},
  author={Xie, Jinheng and Mao, Weijia and Bai, Zechen and Zhang, David Junhao and Wang, Weihao and Lin, Kevin Qinghong and Gu, Yuchao and Chen, Zhijie and Yang, Zhenheng and Shou, Mike Zheng},
  booktitle={International Conference on Learning Representations},
  volume={2025},
  pages={28240--28264},
  year={2025},
  url = {https://arxiv.org/abs/2408.12528},
}

@article{patil2024amused,
  title={{aMUSEd}: An Open {MUSE} Reproduction},
  author={Patil, Suraj and Berman, William and Rombach, Robin and von Platen, Patrick},
  journal={arXiv preprint arXiv:2401.01808},
  year={2024},
  url = {https://arxiv.org/abs/2401.01808},
}

@inproceedings{chang2022maskgit,
  title={{MaskGIT}: Masked Generative Image Transformer},
  author={Chang, Huiwen and Zhang, Han and Jiang, Lu and Liu, Ce and Freeman, William T.},
  booktitle={IEEE/CVF Conference on Computer Vision and Pattern Recognition},
  pages={11315--11325},
  year={2022},
  url = {https://arxiv.org/abs/2202.04200},
}

@inproceedings{hafner2019learning,
  title={Learning Latent Dynamics for Planning from Pixels},
  author={Danijar Hafner and Timothy Lillicrap and Ian Fischer and Ruben Villegas and David Ha and Honglak Lee and James Davidson},
  booktitle={International Conference on Machine Learning},
  pages={2555--2565},
  year={2019},
  organization={PMLR},
  volume = {97},
  series = {Proceedings of Machine Learning Research},
  url = {https://arxiv.org/abs/1811.04551},
}

@inproceedings{chen2024diffusion,
  title={Diffusion Forcing: Next-token Prediction Meets Full-Sequence Diffusion},
  author={Chen, Boyuan and Mart{\'i} Mons{\'o}, Diego and Du, Yilun and Simchowitz, Max and Tedrake, Russ and Sitzmann, Vincent},
  volume={37},
  pages={24081--24125},
  year={2024},
  booktitle = {Advances in Neural Information Processing Systems},
  doi = {10.52202/079017-0759},
  url = {https://arxiv.org/abs/2407.01392},
}

@article{li2026toward,
  title={Toward Cognitive Supersensing in Multimodal Large Language Model},
  author={Li, Boyi and Shen, Yifan and Liu, Yuanzhe and Xu, Yifan and Liu, Jiateng and Li, Xinzhuo and Li, Zhengyuan and Zhu, Jingyuan and Zhong, Yunhan and Lan, Fangzhou and others},
  journal={arXiv preprint arXiv:2602.01541},
  year={2026},
  url = {https://arxiv.org/abs/2602.01541},
}

@inproceedings{janner2022planning,
  title={Planning with Diffusion for Flexible Behavior Synthesis},
  author={Michael Janner and Yilun Du and Joshua Tenenbaum and Sergey Levine},
  booktitle={International Conference on Machine Learning},
  pages={9902--9915},
  year={2022},
  organization={PMLR},
  volume = {162},
  series = {Proceedings of Machine Learning Research},
  url = {https://arxiv.org/abs/2205.09991},
}

@inproceedings{song2025history,
  title={History-Guided Video Diffusion},
  author={Kiwhan Song and Boyuan Chen and Max Simchowitz and Yilun Du and Russ Tedrake and Vincent Sitzmann},
  booktitle={International Conference on Machine Learning},
  pages={56242--56280},
  year={2025},
  organization={PMLR},
  volume = {267},
  series = {Proceedings of Machine Learning Research},
  url = {https://arxiv.org/abs/2502.06764},
}

@inproceedings{hirose2018gonet,
    title={{GONet}: A Semi-Supervised Deep Learning Approach For Traversability Estimation},
    author={Hirose, Noriaki and Sadeghian, Amir and V{\'a}zquez, Marynel and Goebel, Patrick and Savarese, Silvio},
    booktitle={IEEE/RSJ International Conference on Intelligent Robots and Systems},
    pages={3044--3051},
    year={2018},
  url = {https://arxiv.org/abs/1803.03254},
}

@inproceedings{shah2021rapid,
  title={Rapid Exploration for Open-World Navigation with Latent Goal Models},
  author={Dhruv Shah and Benjamin Eysenbach and Nicholas Rhinehart and Sergey Levine},
  booktitle={Conference on Robot Learning},
  pages={674--684},
  year={2022},
  organization={PMLR},
  volume = {164},
  series = {Proceedings of Machine Learning Research},
  url = {https://arxiv.org/abs/2104.05859},
}

@article{hirose2023sacson,
    title={{SACSoN}: Scalable Autonomous Control for Social Navigation},
    author={Hirose, Noriaki and Shah, Dhruv and Sridhar, Ajay and Levine, Sergey},
    journal={IEEE Robotics and Automation Letters},
    volume={9},
    number={1},
    pages={49--56},
    year={2024},
  url = {https://arxiv.org/abs/2306.01874},
  doi = {10.1109/LRA.2023.3329626},
}

@inproceedings{liu2025grl,
  title={GRL-Prompt: Towards Prompts Optimization via Graph-Empowered Reinforcement Learning Using LLMs’ Feedback},
  author={Liu, Yuze and Liu, Tingjie and Zhang, Tiehua and Xia, Youhua and Wang, Jinze and Shen, Zhishu and Jin, Jiong and Ding, Zhijun and Yu, Fei Richard},
  booktitle={Pacific-Asia Conference on Knowledge Discovery and Data Mining},
  pages={426--438},
  year={2025},
  organization={Springer}
}

@article{karnan2022socially,
    title={Socially Compliant Navigation Dataset ({SCAND}): A Large-Scale Dataset of Demonstrations for Social Navigation},
    author={Karnan, Haresh and Nair, Anirudh and Xiao, Xuesu and Warnell, Garrett and Pirk, S{\"o}ren and Toshev, Alexander and Hart, Justin and Biswas, Joydeep and Stone, Peter},
    journal={IEEE Robotics and Automation Letters},
    volume={7},
    number={4},
    pages={11807--11814},
    year={2022},
  url = {https://arxiv.org/abs/2203.15041},
  doi = {10.1109/LRA.2022.3184025},
}

@article{fu2025learning,
  title={Learning human-perceived fakeness in ai-generated videos via multimodal llms},
  author={Fu, Xingyu and Liu, Siyi and Xu, Yinuo and Lu, Pan and Hu, Guangqiuse and Yang, Tianbo and Anantasagar, Taran and Shen, Christopher and Mao, Yikai and Liu, Yuanzhe and others},
  journal={arXiv preprint arXiv:2509.22646},
  year={2025}
}

@misc{qwen36_27b,
    title = {{Qwen3.6-27B}: Flagship-Level Coding in a 27B Dense Model},
    url = {https://qwen.ai/blog?id=qwen3.6-27b},
    author = {{Qwen Team}},
    month = {April},
    year = {2026}
}

@inproceedings{chi2026driver,
  title={{Driver-WM}: A Driver-Centric Traffic-Conditioned Latent World Model for In-Cabin Dynamics Rollout},
  author={Chi, Haozhuang and Qiu, Daosheng and Su, Hao and Liu, Haochen and Li, Zirui and Zhang, Haoruo and Lv, Chen},
  booktitle={European Conference on Computer Vision},
  pages={402--421},
  year={2026},
  organization={Springer},
  url = {https://arxiv.org/abs/2605.05092},
  doi = {10.1007/978-3-032-37083-9_23},
}

@article{chi2026pv,
  title={PV-WM: A Heterogeneous Micro-Macro World Model for Articulated Pedestrian-Vehicle Co-Rollout},
  author={Chi, Haozhuang and Liang, Jingsong and Song, Ziying and Yang, Lei and Li, Shihao and Zhang, Haoruo and Lv, Chen},
  journal={arXiv preprint arXiv:2609.07328},
  year={2026},
  url = {https://arxiv.org/abs/2609.07328},
}

@inproceedings{dong2026lcvn,
  title={{Language-Conditioned World Modeling for Visual Navigation}},
  author={Dong, Yifei and Wu, Fengyi and Dai, Yilong and Kong, Lingdong and Chen, Guangyu and Sha, Yetong and Hu, Qiyu and Liu, Feng and Huang, Siyu and Dai, Qi and Cheng, Zhi-Qi},
  booktitle={Advances in Neural Information Processing Systems},
  year={2026},
  url={https://arxiv.org/abs/2603.26741}
}

@inproceedings{hu2022lora,
title={Lo{RA}: Low-Rank Adaptation of Large Language Models},
author={Edward J Hu and Yelong Shen and Phillip Wallis and Zeyuan Allen-Zhu and Yuanzhi Li and Shean Wang and Lu Wang and Weizhu Chen},
booktitle={International Conference on Learning Representations},
year={2022},
url={https://openreview.net/forum?id=nZeVKeeFYf9}
}

@article{wang2025we,
  title={Do we really need sft? prompt-as-policy over knowledge graphs for cold-start next poi recommendation},
  author={Wang, Jinze and Zhang, Lu and Cui, Yiyang and Zhang, Tiehua and Shen, Zhishu and Liu, Yuze and Ma, Xingjun and Jin, Jiong},
  journal={arXiv preprint arXiv:2510.08012},
  year={2025}
}

@article{loshchilov2017decoupled,
    title={Decoupled weight decay regularization},
    author={Loshchilov, Ilya and Hutter, Frank},
    journal={arXiv preprint arXiv:1711.05101},
    year={2017}
}

@misc{openai2026gpt56,
  author       = {{OpenAI}},
  title        = {{GPT-5.6}: Frontier Intelligence That Scales with Your Ambition},
  year         = {2026},
  month        = jul,
  url = {https://openai.com/index/gpt-5-6/},
}

@inproceedings{heusel2017fid,
  title={{GANs} Trained by a Two Time-Scale Update Rule Converge to a Local {Nash} Equilibrium},
  author={Heusel, Martin and Ramsauer, Hubert and Unterthiner, Thomas and Nessler, Bernhard and Hochreiter, Sepp},
  booktitle={Advances in Neural Information Processing Systems},
  volume={30},
  year={2017},
  url={https://arxiv.org/abs/1706.08500}
}

@inproceedings{fu2023dreamsim,
  title={{DreamSim}: Learning New Dimensions of Human Visual Similarity Using Synthetic Data},
  author={Fu, Stephanie and Tamir, Netanel and Sundaram, Shobhita and Chai, Lucy and Zhang, Richard and Dekel, Tali and Isola, Phillip},
  booktitle={Advances in Neural Information Processing Systems},
  volume={36},
  year={2023},
  url={https://arxiv.org/abs/2306.09344}
}

@article{yang2026quovadis,
  title = {Quo Vadis, World Modeling?},
  author = {Yang, Yu and Yang, Xuemeng and Wen, Licheng and Kong, Lingdong and Hu, Xiaobin and Lu, Dongyue and Chow, Wei and Huang, Xiyan and Feng, Yuxiang and Liao, Yue and Mei, Jianbiao and Fu, Daocheng and Wu, Rong and Cai, Pinlong and Yi, Ran and Tai, Ying and Zhang, Jiangning and Shi, Botian and Liu, Yong and Yan, Shuicheng},
  journal = {arXiv preprint arXiv:2608.02713},
  year = {2026},
  url = {https://arxiv.org/abs/2608.02713}
}

@article{chu2026agentic,
  title = {Agentic World Modeling: Foundations, Capabilities, Laws, and Beyond},
  author = {Chu, Meng and Zhang, Xuan Billy and Lin, Kevin Qinghong and Kong, Lingdong and Zhang, Jize and Tu, Teng and Ma, Weijian and Huang, Ziqi and Yang, Senqiao and Huang, Wei and Jin, Yeying and Rao, Zhefan and Ye, Jinhui and Lin, Xinyu and Zhang, Xichen and Hu, Qisheng and Yang, Shuai and Shen, Leyang and Chow, Wei and Dong, Yifei and Wu, Fengyi and Long, Quanyu and Xia, Bin and Yu, Shaozuo and Zhu, Mingkang and Zhang, Wenhu and Huang, Jiehui and Gui, Haokun and Li, Runyi and Tang, Chenyu and Huang, Dong and Chen, Xuhang and Liu, Rui and Li, Chengzu and Du, Shiyi and Huang, Xu and Che, Haoxuan and Chen, Long and Chen, Qifeng and Wang, Wenya and Zhang, Wenxuan and Qi, Xiaojuan and Deng, Yang and Li, Yanwei and Shou, Mike Zheng and Cheng, Zhi-Qi and Ng, See-Kiong and Liu, Ziwei and Torr, Philip and Jia, Jiaya},
  journal = {arXiv preprint arXiv:2604.22748},
  year = {2026},
  url = {https://arxiv.org/abs/2604.22748}
}

@article{liao2025envisioner,
  title = {Genie Envisioner: A Unified World Foundation Platform for Robotic Manipulation},
  author = {Liao, Yue and Zhou, Pengfei and Huang, Siyuan and Yang, Donglin and Chen, Shengcong and Jiang, Yuxin and Hu, Yue and Cai, Jingbin and Liu, Si and Luo, Jianlan and Chen, Liliang and Yan, Shuicheng and Yao, Maoqing and Ren, Guanghui},
  journal = {arXiv preprint arXiv:2508.05635},
  year = {2025},
  url = {https://arxiv.org/abs/2508.05635}
}

@inproceedings{hu2026navthinker,
  title = {{NavThinker}: Action-Conditioned World Models for Coupled Prediction and Planning in Social Navigation},
  author = {Hu, Tianshuai and Gong, Zeying and Kong, Lingdong and Mei, XiaoDong and Ding, Yiyi and Zeng, Qi and Liang, Ao and Li, Rong and Zhong, Yangyi and Liang, Junwei},
  booktitle = {IEEE/RSJ International Conference on Intelligent Robots and Systems},
  year = {2026},
  url = {https://arxiv.org/abs/2603.15359}
}

@inproceedings{li2026lingbotva,
  title = {Causal World Modeling for Robot Control},
  author = {Lin Li and Qihang Zhang and Yiming Luo and Shuai Yang and Ruilin Wang and Luyao Zhang and Mingrui Yu and Zelin Gao and Nan Xue and Boyu Zhou and Xing Zhu and Mingyu Ding and Yujun Shen and Yinghao Xu},
  booktitle = {Robotics: Science and Systems},
  year = {2026},
  doi = {10.15607/RSS.2026.XXII.016},
  url = {https://arxiv.org/abs/2601.21998}
}

@article{wang2024emu3,
  title = {{Emu3}: Next-Token Prediction is All You Need},
  author = {Wang, Xinlong and Zhang, Xiaosong and Luo, Zhengxiong and Sun, Quan and Cui, Yufeng and Wang, Jinsheng and Zhang, Fan and Wang, Yueze and Li, Zhen and Yu, Qiying and Zhao, Yingli and Ao, Yulong and Min, Xuebin and Li, Tao and Wu, Boya and Zhao, Bo and Zhang, Bowen and Wang, Liangdong and Liu, Guang and He, Zheqi and Yang, Xi and Liu, Jingjing and Lin, Yonghua and Huang, Tiejun and Wang, Zhongyuan},
  journal = {arXiv preprint arXiv:2409.18869},
  year = {2024},
  url = {https://arxiv.org/abs/2409.18869}
}

@inproceedings{dong2026havln,
  title={{HA-VLN 2.0: An Open Benchmark and Leaderboard for Human-Aware Navigation in Discrete and Continuous Environments with Dynamic Multi-Human Interactions}},
  author={Dong, Yifei and Wu, Fengyi and He, Qi and Kong, Lingdong and Li, Heng and Li, Minghan and Cheng, Zebang and Zhou, Yuxuan and Sun, Jingdong and Dai, Qi and others},
  booktitle={IEEE/RSJ International Conference on Intelligent Robots and Systems},
  year={2026},
  url={https://arxiv.org/abs/2503.14229}
}

@article{chen2025januspro,
  title = {{Janus-Pro}: Unified Multimodal Understanding and Generation with Data and Model Scaling},
  author = {Chen, Xiaokang and Wu, Zhiyu and Liu, Xingchao and Pan, Zizheng and Liu, Wen and Xie, Zhenda and Yu, Xingkai and Ruan, Chong},
  journal = {arXiv preprint arXiv:2501.17811},
  year = {2025},
  url = {https://arxiv.org/abs/2501.17811}
}

@inproceedings{wu2025vilau,
  title = {{VILA-U}: a Unified Foundation Model Integrating Visual Understanding and Generation},
  author = {Yecheng Wu and Zhuoyang Zhang and Junyu Chen and Haotian Tang and Dacheng Li and Yunhao Fang and Ligeng Zhu and Enze Xie and Hongxu Yin and Li Yi and Song Han and Yao Lu},
  booktitle = {International Conference on Learning Representations},
  year = {2025},
  url = {https://arxiv.org/abs/2409.04429}
}

@inproceedings{zhou2025transfusion,
  title = {{Transfusion}: Predict the Next Token and Diffuse Images with One Multi-Modal Model},
  author = {Zhou, Chunting and Yu, Lili and Babu, Arun and Tirumala, Kushal and Yasunaga, Michihiro and Shamis, Leonid and Kahn, Jacob and Ma, Xuezhe and Zettlemoyer, Luke and Levy, Omer},
  booktitle = {International Conference on Learning Representations},
  year = {2025},
  pages = {6446--6469},
  url = {https://arxiv.org/abs/2408.11039}
}

@article{wu2026tiresias,
  title={Tiresias: Towards Language-Conditioned World Models},
  author={Wu, Fengyi and Dong, Yifei and Sha, Yetong and Kong, Lingdong and Hu, Qiyu and Xiong, Bohan and Dai, Qi and Hauptmann, Alexander G and Cheng, Zhi-Qi},
  journal={arXiv preprint arXiv:2611.10392},
  year={2026}
}

@inproceedings{wu2026govig,
  title={GoViG: Goal-Conditioned Visual Navigation Instruction Generation via Multimodal Reasoning},
  author={Wu, Fengyi and Dong, Yifei and Dai, Yilong and Chen, Guangyu and Wu, Qifeng and Huang, Huiting and Wang, Hang and Dai, Qi and Hauptmann, Alexander G and Cheng, Zhi-Qi},
  booktitle={Findings of the Association for Computational Linguistics: ACL 2026},
  pages={14899--14920},
  year={2026}
}

@inproceedings{xie2025showo2,
  title = {{Show-o2}: Improved Native Unified Multimodal Models},
  author = {Xie, Jinheng and Yang, Zhenheng and Shou, Mike Zheng},
  booktitle = {Advances in Neural Information Processing Systems},
  year = {2025},
  pages = {47490--47518},
  doi = {10.52202/085713-1584},
  volume = {38},
  url = {https://arxiv.org/abs/2506.15564}
}

@inproceedings{austin2021d3pm,
  title = {Structured Denoising Diffusion Models in Discrete State-Spaces},
  author = {Austin, Jacob and Johnson, Daniel D. and Ho, Jonathan and Tarlow, Daniel and van den Berg, Rianne},
  booktitle = {Advances in Neural Information Processing Systems},
  year = {2021},
  pages = {17981--17993},
  volume = {34},
  url = {https://arxiv.org/abs/2107.03006}
}

@inproceedings{nie2025llada,
  title = {Large Language Diffusion Models},
  author = {Nie, Shen and Zhu, Fengqi and You, Zebin and Zhang, Xiaolu and Ou, Jingyang and Hu, Jun and Zhou, Jun and Lin, Yankai and Wen, Ji-Rong and LI, Chongxuan},
  booktitle = {Advances in Neural Information Processing Systems},
  year = {2025},
  pages = {50608--50646},
  doi = {10.52202/085713-1689},
  volume = {38},
  url = {https://arxiv.org/abs/2502.09992}
}

@inproceedings{chang2023muse,
  title = {{Muse}: Text-To-Image Generation via Masked Generative Transformers},
  author = {Huiwen Chang and Han Zhang and Jarred Barber and Aaron Maschinot and Jose Lezama and Lu Jiang and Ming-Hsuan Yang and Kevin Patrick Murphy and William T. Freeman and Michael Rubinstein and Yuanzhen Li and Dilip Krishnan},
  booktitle = {International Conference on Machine Learning},
  year = {2023},
  pages = {4055--4075},
  volume = {202},
  series = {Proceedings of Machine Learning Research},
  url = {https://arxiv.org/abs/2301.00704}
}

@article{anderson2018evaluation,
  title = {On Evaluation of Embodied Navigation Agents},
  author = {Anderson, Peter and Chang, Angel and Chaplot, Devendra Singh and Dosovitskiy, Alexey and Gupta, Saurabh and Koltun, Vladlen and Kosecka, Jana and Malik, Jitendra and Mottaghi, Roozbeh and Savva, Manolis and Zamir, Amir R.},
  journal = {arXiv preprint arXiv:1807.06757},
  year = {2018},
  url = {https://arxiv.org/abs/1807.06757}
}

@inproceedings{zheng2023llmjudge,
  title = {Judging {LLM}-as-a-Judge with {MT-Bench} and Chatbot Arena},
  author = {Zheng, Lianmin and Chiang, Wei-Lin and Sheng, Ying and Zhuang, Siyuan and Wu, Zhanghao and Zhuang, Yonghao and Lin, Zi and Li, Zhuohan and Li, Dacheng and Xing, Eric and Zhang, Hao and Gonzalez, Joseph and Stoica, Ion},
  booktitle = {Advances in Neural Information Processing Systems},
  year = {2023},
  pages = {46595--46623},
  doi = {10.52202/075280-2020},
  volume = {36},
  note = {Datasets and Benchmarks Track},
  url = {https://arxiv.org/abs/2306.05685}
}

\end{document}